\documentclass[acmtog, authorversion]{acmart} %,anonymous,review
\acmSubmissionID{535}

\usepackage{booktabs} % For formal tables
\usepackage{graphicx}
\usepackage{makecell}
\usepackage{subcaption}
\usepackage{etoolbox,lipsum}
\usepackage{soul}
\usepackage{hyperref}

\usepackage[ruled]{algorithm2e} % For algorithms

\SetAlFnt{\small}
\SetAlCapFnt{\small}
\SetAlCapNameFnt{\small}
\SetAlCapHSkip{0pt}

\copyrightyear{2024}
\acmYear{2024}
\setcopyright{rightsretained}
\acmConference[SA Conference Papers '24]{SIGGRAPH Asia 2024 Conference Papers}{December 3--6, 2024}{Tokyo, Japan}
\acmBooktitle{SIGGRAPH Asia 2024 Conference Papers (SA Conference Papers '24), December 3--6, 2024, Tokyo, Japan}
\acmDOI{10.1145/3680528.3687621}
\acmISBN{979-8-4007-1131-2/24/12}

\graphicspath{ {./images/} }

\begin{document}
% Title portion
\title{Text-Guided Texturing by Synchronized Multi-View Diffusion}

% DO NOT ENTER AUTHOR INFORMATION FOR ANONYMOUS TECHNICAL PAPER SUBMISSIONS TO SIGGRAPH 2019!
\author{Yuxin Liu}
\email{yxliu22@cse.cuhk.edu.hk}
\orcid{0009-0001-5658-7360}
\affiliation{%
 \institution{The Chinese University of Hong Kong}
 %\city{Hong Kong}
 \country{Hong Kong}}

\author{Minshan Xie}
\email{msxie@cse.cuhk.edu.hk}
\orcid{0000-0002-6288-1611}
\affiliation{%
 \institution{The Chinese University of Hong Kong}
 %\city{Hong Kong}
 \country{Hong Kong}
}

\author{Hanyuan Liu}
\email{hy.liu@cityu.edu.hk}
\affiliation{%
\institution{City University of Hong Kong}
\country{Hong Kong}}

\author{Tien-Tsin Wong}
\email{tt.wong@monash.edu}
\affiliation{%
 \institution{The Chinese University of Hong Kong}
 \country{Hong Kong}
}
\affiliation{%
 \institution{Monash University}
 \country{Australia}
}

\begin{teaserfigure}
    \centering
    \captionsetup{type=figure}    
    \vspace{-0.1cm}
    \includegraphics[width=.9\linewidth]{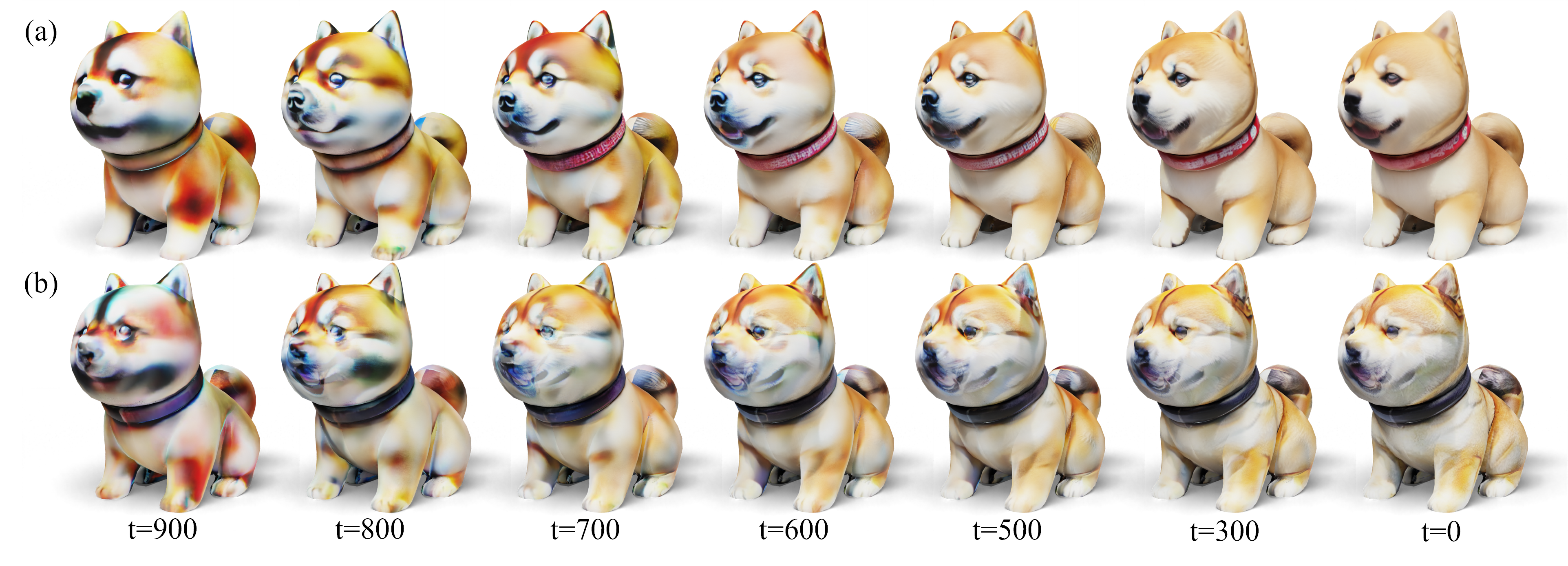}
    \vspace{-0.1cm}
  \caption{Visualization of RGB textures at different intermediate denoising steps. (a) With the proposed MVD, consensus on surface color layouts is reached from the early stage of the diffusion process. (b) Without MVD and in the absence of consensus, obvious seams, ghosting, and fragmentation artifacts are observed at the end as views are inconsistent with each other.}        
    \label{fig:consensus}
\end{teaserfigure}%

\begin{abstract}
This paper introduces a novel approach to synthesize texture to dress up a 3D object, given a text prompt. 
Based on the pre-trained text-to-image (T2I) diffusion model, existing methods usually employ a project-and-inpaint approach, in which a view of the given object is first generated and warped to another view for inpainting. But it tends to generate inconsistent texture due to the asynchronous diffusion of multiple views. We believe that such asynchronous diffusion and insufficient information sharing among views are the root causes of the inconsistent artifacts. 
In this paper, we propose a synchronized multi-view diffusion approach that allows the diffusion processes from different views to reach a consensus on the generated content early in the process, and hence ensures the texture consistency. To synchronize the diffusion, we share the denoised content among different views in each denoising step, specifically by blending the latent content in the texture domain from overlapping views. Our method demonstrates superior performance in generating consistent, seamless and highly detailed textures, comparing to state-of-the-art methods.
\end{abstract}






\maketitle

\section{Introduction}
\label{sec:intro}

Existing rendering systems predominantly utilize polygonal geometric primitives, like triangles, and apply texture mapping to enhance visual appeal. These textures are sourced from photographs, manual paintings, or procedural computations~\cite{ebert2003}. However, most of these texture acquisition methods require domain-specific expertise and manual efforts, which is far less convenient than generating textures from a text prompt. In this paper, we introduce a method for generating textures for a 3D object based on text descriptions.

There exists a few works~\cite{richardson2023texture, chen2023text2tex, cao2023texfusion} that try to achieve the above goal. One typical approach adopted is to generate a view using depth-conditioned text-to-image (T2I) models, project this view onto the object's surface, and render the partially textured object from a rotated camera position. Next, the missing texture regions are filled using inpainting from the current viewpoint with the T2I model. These project-and-inpaint methods often result in texture inconsistencies, as shown in Fig.~\ref{fig:illustrate}. Each view, diffused separately, lacks adequate consistency constraints, failing to ensure a seamless appearance from various viewpoints.

The primary issue, we believe, lies in the {\em asynchronous} diffusion among different views. Note that rendered geometry does not fully define the appearance of the target object, imposing only limited control over the colors, patterns and structures on the surface of the object, thus has difficulty in resisting drifts in texture content during a sequential inpainting process. As a consequence, visible seams will be left unsolved on the textured object due to error accumulation, as shown in Fig. \ref{fig:illustrate}. 
Our solution is a {\em synchronized multi-view diffusion} approach. This method achieves early-stage content {\em consensus}, which is essential for consistent structure and error correction across views.

\begin{figure*}[th!]
\vspace{-0.3cm}
  \includegraphics[width=1.\linewidth]{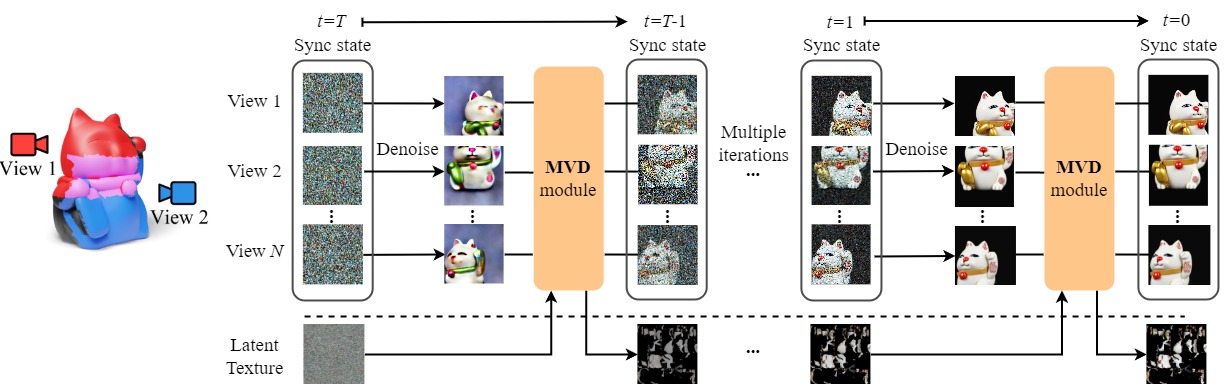}
  \caption{Left: An illustration of information exchange in the overlapped region (pink region) at intermediate steps of diffusion. Without information exchange, denoising results in different views of the same object could diverge into different directions, leading to seams when projecting to an output texture. Right: To address this issue, we propose a Multi-View Diffusion module that fuses intermediate steps of the denoising process, basing the next denoising step on a consensus of the current step. Here, we illustrate how MVD synchronizes and fuses view information from timesteps $T$ to 0.}
    \vspace{-0.3cm}
  \label{fig:framework}
\end{figure*}

In order to synchronize the diffusion among different views, it is necessary to allow the denoised content to be shared among different views in each denoising step. The overlapping regions among different views on the textured object (Fig.~\ref{fig:framework}, left) serve as the information exchange sites. During each denoising step, we share (blend in our case) the latent from different views in the UV texture domain, if they have an overlap. The texture consensus can be obtained during the early stage of denoising, as demonstrated in Fig~\ref{fig:consensus}a. Note that, all views share the same importance during this consensus process, i.e. no view is overriding the others during the denoising. 

With the proposed approach, we obtain plausible and highly detailed textures for arbitrarily given 3D objects. Please refer to the numerous results presented in this paper and the supplement. We have evaluated the results generated by our synchronous multi-view diffusion, via quantitative   and qualitative evaluations. Superior performance is achieved compared to state-of-the-art methods. In summary, our contributions are as follows.
\begin{itemize}
    \item We identify that the problem in existing methods stems from asynchronous diffusion and propose a novel synchronous multi-view diffusion approach to address this issue.

    \item We proposed a practical solution to generate consistent, seamless, plausible, and highly detailed textures given a text prompt. 
    
    \item We conduct extensive experiments on a variety of 3D meshes, demonstrating superior texturing performance compared with state-of-the-art methods.
\end{itemize}

\if 0
While neural-based 3D generation from text prompts becomes feasible, there remains a gap to practically apply neural-based 3D objects in existing rendering pipelines, which are based on polygonal primitives (specifically triangles). Instead, we present in this paper a neural-based 
\fi

\section{Related Work}
\label{sec:relatedwork}

\subsection{Texture Synthesis Methods} 
Early works on texture related topic mainly focused on generating 2D and 3D tileable texture maps from exemplars \cite{efros1999texture, ashikhmin2001synthesizing, kwatra2005texture, kopf2007solid}, and applying them to a given object in a seamless way \cite{praun2000lapped, ebert2003texturing}. Existing works also explored the correlation between surface structures and texture details, enabling geometry-aware synthesis and transfer of texture details \cite{wong1997geometry, lu2007context, mertens2006texture}. However, both rule-based methods and early machine learning models are of limited capacity, making it infeasible to  synthesize textures with rich semantics on complicated geometry at that time. 

In the recent few years, deep learning methods, especially convolutional neural networks exhibit their strong capabilities on image-related tasks. Texture synthesis methods in the recent years are based on popular generative models such as GANs \cite{goodfellow2014generative, karras2019style}, VAEs \cite{kingma2013auto, van2017neural}, and state-of-the-art diffusion model \cite{ho2020denoising, sohl2015deep}, to sample from a prior trained on rendered views \cite{raj2019learning, xian2018texturegan}, voxels \cite{zhou20213d, smith2017improved}, point clouds \cite{Zeng2022LIONLP, Nichol2022PointEAS, yu2023pointsto3d}, and implicit representations \cite{nam20223d, gupta20233dgen, lorraine2023att3d, Li2022DiffusionSDFTV} of a textured 3D object. However, generating highly-detailed texture for a given mesh is seldom discussed, even it is demanded.

% \noindent\textbf{Zero-shot Texturing Using Text-Image Prior.}
\subsection{Zero-shot Texturing Using Text-Image Prior}
Recently, priors trained on large scale text-image data have empowered researches on various image generation and modification task. Due to the lack of large-scale 3D dataset with high-quality annotation for training a prior natively in 3D, many previous 3D content generation works have alternatively chosen to use 2D priors, and achieved strong zero-shot capability when paired with carefully designed sampling pipelines. 
Following this idea, several works in texture generation \cite{sanghi2022clip, Khalid2022CLIPMeshGT, 9878592, jain2022zero} distill gradients from CLIP model \cite{radford2021learning}, which trained to correlate text and image using contrastive learning. These gradients are used to update rendered views of a 3D object iteratively, to make it conform to the given text. 

On the other hand, score distillation methods (SDS) \cite{poole2022dreamfusion, lin2023magic3d, metzer2023latent, chen2023fantasia3d, youwang2024paint, wang2024prolificdreamer} distill gradients from state-of-the-art image generation models, namely, diffusion models \cite{ho2020denoising, sohl2015deep}. For texture generation, SDS-based methods add noise to a rendered view of the current 3D object, and use the diffusion model to reconstruct the noiseless input. By this means, the resulting clean view is incorporated with prior knowledge from the diffusion model and can be used to update the object texture. However, these methods cannot balance both high-quality generation and short optimization time, compared to generating 2D images using diffusion ancestral sampling.

TEXTure \cite{richardson2023texture} and Text2Tex \cite{chen2023text2tex} on the other hand, approached this problem by progressive inpainting with a depth-conditioned diffusion model \cite{rombach2022high}. Their methods start by generating a view of the object using the diffusion model with the depth map rendered from the mesh as control, and then projecting the screen-space textures onto the object surface. In every iteration, they rotate the object by an angle, and inpaint the partially visible texture from this new view. These methods achieve better texture sharpness and faster running time, but may suffer from obvious seams and over-fragmentation, due to the asynchronous nature of their diffusion processes.
A recent work, TexFusion \cite{cao2023texfusion}, attempts to address the inconsistency problem by performing a complete round of project-and-inpaint in each denoising time-step, in an autoregressive manner. 
Although both their method and ours use a latent UV texture for projection, our method treats each view equally, fusing latent information from all views in a synchronized fashion to reach a consensus in terms of content structure and color distribution. We show that consensus can be quickly reached in the very early stage of denoising process with good convergence (Fig.~\ref{fig:consensus}a).
In addition to the above-mentioned zero-shot methods, Paint3D\cite{zeng2024paint3d} extends the progressive inpainting by a data-driven refinement stage in the UV texture domain. Paint3D  forms a 2D ``position map'' by mapping  the coordinate of each surface point to the texture domain according to the UV layout. This position map serves as a 2D representation of the input geometry, and is used as a conditional input to a 2D T2I diffusion model for  generating texture in UV domain. The UV-space T2I model has to be  fine-tuned from a pre-trained diffusion model, using the paired mesh and texture data. In contrast to Paint3D, our method does not require a high-quality UV layout, thus is more practical for meshes that have their UV layout unwarped automatically. Our zero-shot pipeline is also not limited to specific training data distribution, thus is not bounded by the quality and variety of the training data.

\section{Diffusion Model Preliminaries}
\label{sec:preliminary}

Denoising Diffusion Models are generative neural networks trained to reverse a Markov process which diffuses its input data to pure Gaussian noise. During training, given input data $x$, a fixed forward process gradually destroys the information in each intermediate time step $t$, eventually approaching pure Gaussian noise $x_T$ at $t=T$. Then a neural noise predictor $\epsilon_\theta(x_t, t)$ (e.g., a U-Net) is trained to estimate the true noise $\epsilon$ mixed into the input, given an arbitrary time step $t$ and the corresponding noisy data $x_t$. Inference can be performed by sampling from the Gaussian noise at time $T$, and iteratively removing part of the predicted noise to reach a fully denoised $x_0$.

In our work, we utilize Stable Diffusion model~\cite{rombach2022high}, which
is trained to denoise in low-resolution latent space $z=\mathcal{E}(x)$ encoded by a pre-trained VAE encoder $\mathcal{E}$, as it can significantly reduce the computational cost. Then, an image can be generated through the following steps:
\begin{enumerate}
    \item Sample a noise image  $z_T$ from the standard normal distribution. 
    
    \item For each intermediate diffusion time step $t$:
    \begin{enumerate}
        \item  Given noisy latent image $z_t$, the model predicts the noise $\epsilon_\theta(z_t, t)$ in the current latent image. A clean intermediate state $z_{0|t}$ can be obtained by removing the noise from $z_t$ (modifications on $z_{0|t}$ can be applied to affect the subsequent denoising process).
        \item  Compute the latent image $z_{t-1}$ as a linear combination of $z_t$ and $z_{0|t}$ using time-step-related coefficients. The obtained noisy latent image will be used as input at the next time step $t-1$.
    \end{enumerate}

    \item Decode the fully denoised $z_0$ with the VAE decoder $\mathcal{D}$ to obtain the output image $x=\mathcal{D}(z_0)$.
\end{enumerate}

In addition to text conditioning using built-in attention mechanisms, several external encoders and adapters \cite{zhang2023adding, mou2023t2i} have been designed to enable diffusion models to be conditioned on other modalities. 
ControlNet as one of these methods, allows diffusion models to generate images conditioned on screen-space depth or normal maps.

\section{Synchronized Multi-View Diffusion}
\label{sec:method}
Given the object geometry and a known camera, the ground truth depth map or normal map can be rendered to condition the above 2D image generation, enabling the generation of 2D views of the desired textured 3D object. 
Therefore, a naive object texturing framework can be designed as follows to perform zero-shot texture generation using a pretrained T2I diffusion model without texture domain knowledge. 
We first surround the target 3D object $m$ with multiple cameras $\{v_1, v_2, \cdots 
v_n\}$, each covers part of the object. Then, a T2I diffusion process is assigned to synthesize each of these 2D views $\{z^{(v_1)}_t, z^{(v_2)}_t, \cdots, z^{(v_n)}_t \}$, using the text prompt $y$ as guidance and conditional images (depth or normal map rendered from the corresponding viewpoints) as the condition. 
With sufficient views, we can obtain the complete texture map covering the whole object, by projecting the generated pixels from each view onto the object surface, which in turn can be mapped to the texture domain (UV space).

The key challenge lies in obtaining a uniform texture map from multiple generated views that are inconsistent to each other. 
Texture map with severe fragmentation or ghosting may be generated when directly combining each view generated through a separate diffusion process (Fig.~\ref{fig:consensus}b). One can use postprocessing smoothing to reduce the obvious seams, but with the trade-off of over-blurriness. 
Our major contribution lies on how to ensure the consistency among the generated views, and hence prevent obvious fragmentation and  seams, so that sharp and highly detailed textures can be obtained.

The root cause of the inconsistency among generated views is that the corresponding diffusion processes of views are performed in a more or less independent fashion. Medium and fine-scale structures that are not apparent in the rendered depth images will have ill-defined appearance in the generated view, leaving inconsistencies among views and leading to the above mentioned problems. In order to foster a consistent texture at different scales, we need to share the latent information among views since the beginning of the denoising process (from time step $T$). To achieve this, the latent values of views are shared among each other between every denoising step. We call this {\em synchronized multi-view diffusion} (MVD). 
As illustrated in Fig.~\ref{fig:framework}(right), we break down the multi-view texturing procedure into individual denoising time steps, and applies information sharing among the views to ensure convergence to a consistent object appearance, avoiding seams and fragments.

This sharing can be done through overlapping among views (Fig.~\ref{fig:framework}(left)) in the texture domain. The latent values from different views in the overlapping regions can be blended with appropriate weights, and the combined values are then used for the next round of denoising.% (Fig.~\ref{fig:sharing}).

Such latent value sharing allows the diffusion processes of all views to reach a consensus over the generated texture in terms of overall structure and color distribution in the early stage of the denoising process. Fig.~\ref{fig:consensus}a visualizes how fast the consensus could be reached in generating the texture, in comparing to the one without information sharing in Fig.~\ref{fig:consensus}b.

\subsection{Multi-View Diffusion in Texture Domain}
\label{subsec:uvmd}

\begin{figure}[t!]
  \centering
  \vspace{-0.3cm}
    \includegraphics[width=0.98\linewidth]{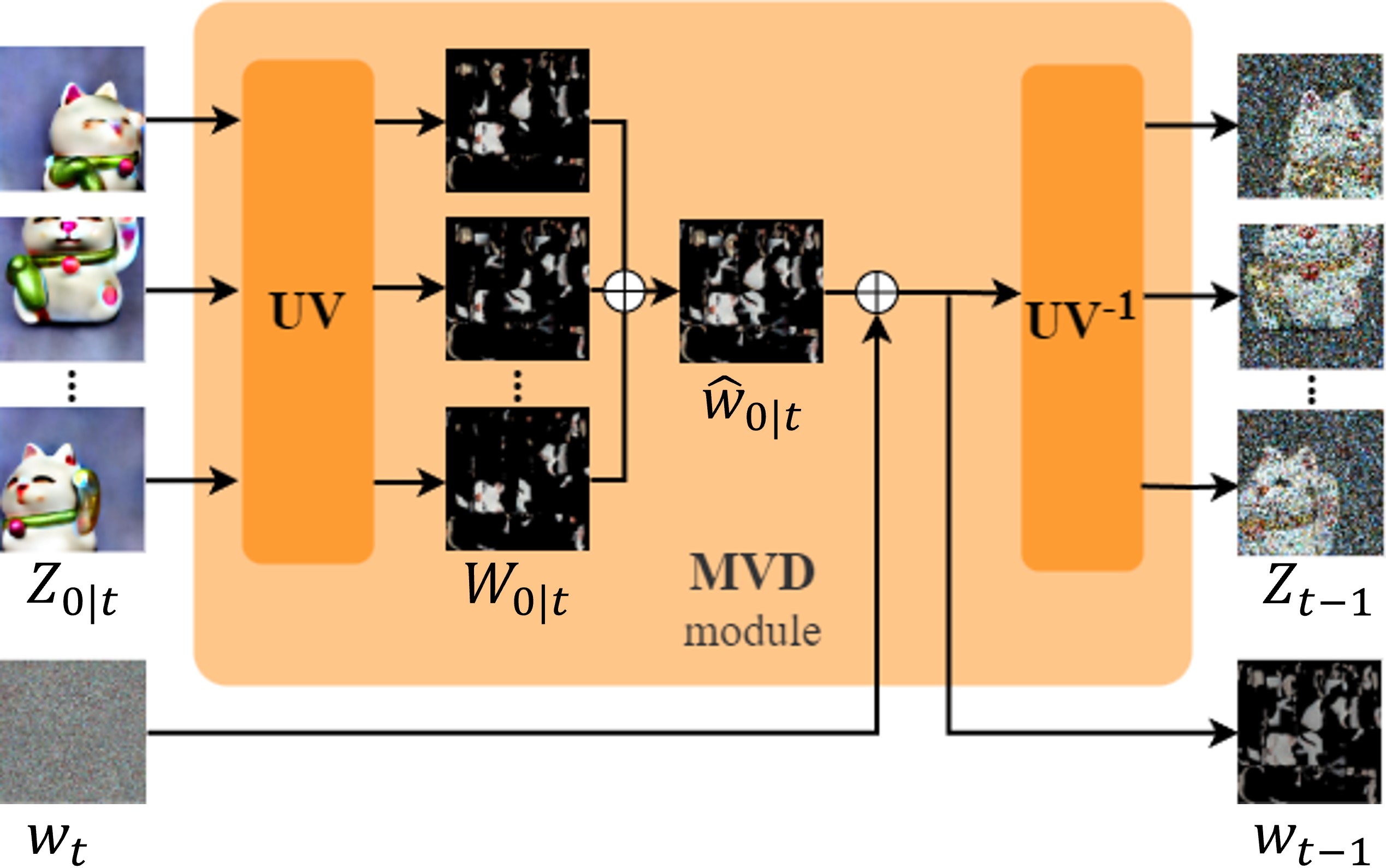}
  \vspace{-0.3cm}
  \caption{A zoom-in diagram of the MVD module. Here, denoised views are first projected to partial textures in the UV texture domain and aggregated into a complete, clean latent texture. Then, we can sample the latent texture of the next time step based on this clean texture, and project to screen space to obtain consistent views.}
  \label{fig:MVD}
\end{figure}

Our Multi-View Diffusion module  utilizes the overlapping region in UV texture space for synchronization. Instead of pairwise screen-space warping, we choose to use the UV mapping as a pre-established connection among all views. This mapping allows us to align predictions from multiple views to this canonical texture layout, and distribute the synchronized results to each individual view through rendering. The detailed procedures are as follows.

At initial time step $T$, we first initialize a latent texture $w_T$ with standard normal distribution. Then we render the source mesh from views $V=\{v_i\}_{i=1}^N$ using this map as the texture, to obtain 3D-consistent initial views $Z_T=\{z^{(v_i)}_T\}_{i=1}^N$ of the object. Background pixels are randomly sampled with the same noise distribution and then composited with the rendered foreground object using the object mask.

At each time step $t$, we can compute $Z_{t-1}$ of the next time step based on noiseless views $Z_{0|t}=\{z^{(v_i)}_{0|t}\}_{i=1}^N$ estimated by the model. This computation is done separately for each view by default, with no consistency constraints. To guarantee the consistency among the $Z_{t-1}$ views, we choose to project $Z_{0|t}$ to UV texture space, so that we can obtain a noisy latent texture in UV space and then distribute to all views through rendering, just as we did to obtain 3D-consistent noise in the initialization step. Here we obtain the partially covered textures $W_{0|t}= \textbf{UV}(Z_{0|t})$ from the noiseless views through a mapping operation from screen space to texture space, denoted as ${\rm \textbf{UV}}$. We will be using ${\rm \textbf{UV}^{-1}}$ to denote mapping from texture space back to screen space (i.e., rendering) later in this section. 

Inspired by \cite{richardson2023texture, chen2023text2tex}, we utilize the cosine similarities between per-pixel normal vectors and the viewing direction to determine the weight of each contributing pixel, as geometries facing away from the camera are less reliable. Hence, current-step clean texture can be obtained through aggregating these partial textures at each overlapping texel:

\begin{equation}
\hat{w}_{0|t}=\frac{\sum_{i=1}^N w_{0|t}^{(v_i)}  \odot {\bf UV}(\theta^{(v_i)})^\alpha}{\sum_{i=1}^N   {\bf UV}(\theta^{(v_i)})^\alpha + \gamma}
\label{eq:geoaware}
\end{equation}

\noindent where $\odot$ denotes element-wise (texel-wise) multiplication, $\alpha$ is an exponent to adjust the smoothness of blending and $\gamma$ is a small constant to avoid division-by-zero in the masked-out regions;
$\theta^{(v_i)}$ is the weight map for view $v_i$ and it is calculated as the cosine similarity between the viewing direction of $v_i$ and the normal vector, with each element calculated as,
\begin{equation}
    \theta^{(v_i)}(p)=\frac{\Vec{v}_i(p)\cdot\Vec{n}_m(p)}{\|\Vec{v}_i(p)\|\|\Vec{n}_m(p)\|},
\end{equation}
where $p$ donates the 3D surface point on object $m$, corresponding to the screen pixel of interest, $\Vec{v}_i(p)$ is the viewing direction from $p$ to view $v_i$ and $\Vec{n}_m(p)$ is the normal vector at $p$. The formulation of cosine similarity intuitively means the view facing directly to the surface point is more important.
To balance between strong consistency constraint and high-frequency detail preservation, we schedule the $\alpha$ exponent to increase linearly throughout the denoising process. In the early stage, consensus could be stably reached with a low exponent, as different views in $Z_{0|t}$ contribute approximately equal to each visible texel; while in the late stage, details could be effectively preserved with a higher exponent, as high-frequency details will not be smoothed out due to naive blending.

Based on the diffusion sampling method introduced in Sec. \ref{sec:preliminary}, the texture in next time step $w_{t-1}$ can be sampled using $\hat{w}_{0|t}$ and $w_t$, analogous to the sampling of $z_{t-1}$ based on $z_{0|t}$ and $z_t$ in diffusion image generation. 
Finally, this texture is mapped back to corresponding views $Z_{t-1}={\rm \textbf{UV}^{-1}}(w_{t-1})$. Fig.~\ref{fig:MVD} illustrates this MVD process.
These generated views should have consistent textures on corresponding regions. To assign background with valid latent noise, we encode a random solid color image and add proper noise based on the current time step.

\begin{figure}
    \centering
    \vspace{-0.3cm}
    \includegraphics[width=.95\linewidth]{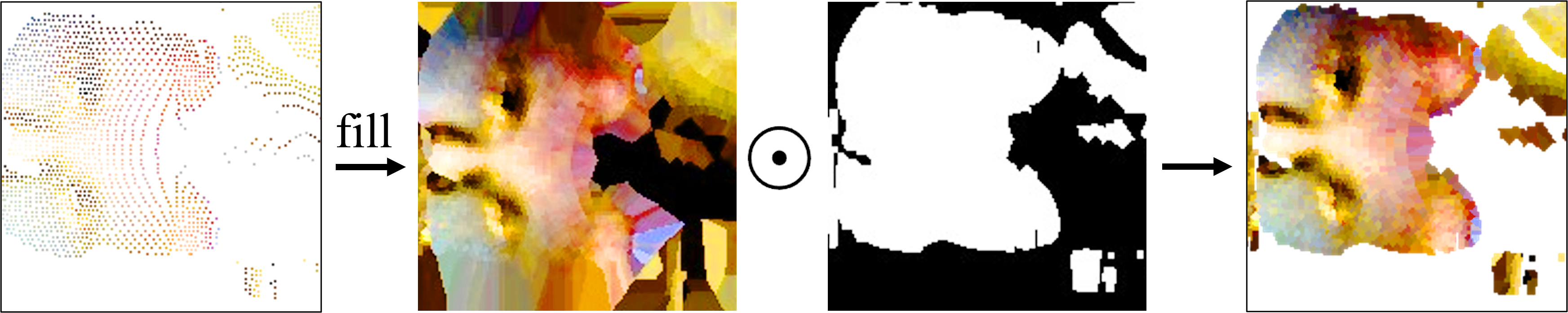}
    \vspace{-0.3cm}
    \caption{An illustration of how the forwardly projected pixels are disjoint in UV space, and how filling and masking are applied to obtain partial textures with large patches of valid texels.}
    \label{fig:emptyspace}
    \vspace{-0.5cm}
\end{figure}

Our method implements the screen-to-texture projection ${\rm \textbf{UV}}$ using back-propagation supported by a differentiable renderer. This implementation has the drawback that only texels visible on screen can receive non-zero gradients. Consequently, texels with valid latent values appear as disconnected dots instead of complete patches on the partial latent textures, as pixels are forwardly projected onto UV space. Information exchange will be compromised if multiple such textures are aggregated, due to the sparsity.
To tackle this issue, we simply apply a Voronoi-based filling~\cite{aurenhammer1991voronoi} to propagate the latent texels to fill up all empty regions in the UV space. Fig~\ref{fig:emptyspace} shows the latent texture filled with Voronoi diagram. The fully-filled latent texture are then masked according to a triangle visibility mask $M(v_i)$, ensuring the propagation does not exceed the boundary of regions visible to view $v_i$, as shown in Fig~\ref{fig:emptyspace}. These filled textures is now ready for aggregation since they are free of the sparsity problem mentioned above.

\subsection{Self-attention Reuse}
\label{subsec:selfattention}
Due to the fact that multiple views of the same object tend to develop coherent looks when they are concatenated into a single image during denoising \cite{tsalicoglou2023textmesh}, we perform concatenation in the batched self-attention function, allowing each view to attend to multiple relevant views to encourage consistency. Benefiting from the pre-trained self-attention weights, spatially distant views can now develop highly-related appearances even when overlapping in UV-space is small or unavailable. However, we observed certain degradation in details when pairwise attentions are computed among a large number of view, which may be due to the Softmax function used in attention computation \cite{vaswani2017attention}. Therefore, we propose two attention schemes to achieve the appearance harmonization in different stages of denoising.
\begin{equation}
\resizebox{0.42\textwidth}{!}{
    $\text{Attention}(v_i)=
    \begin{cases}
        \!\begin{aligned}
        & \beta\cdot\text{SA}(v_i, v_{\{i-1, i, M(i)\}})\\
        & + (1-\beta)\cdot\text{SA}(v_i, v_{\text{ref}})
        \end{aligned}        
        & \text{for } t>t_{\text{ref}}\\
        \\
        \text{SA}(v_i, v_{\{i-1, i, M(i)\}}) & \text{otherwise}\\
    \end{cases}$
}
\vspace{-0.2cm}
\end{equation}

\noindent where ${\rm SA}(v_m, v_n)$ denotes that we use the pre-trained self-attention layer to attend to view $v_n$ when calculating the attention results of view $v_m$ in the denoising U-Net. We use $t_{\rm ref}$ to denote the time step where we switch between two attention schemes. To enforce a global harmonization in the early stage of denoising, we apply two attention components before time step $t_{\rm ref}$: (1) attention to the view itself, its neighbor, and its mirrored view $v_{M(i)}$ with respect to the front view, which encourages harmonization between left and right view for bilaterally symmetric objects; and (2)  attention to a reference view $v_{\rm ref}$, which can be the default front view or a manually selected view. These two components are balanced by a weight factor $\beta$ (default to 1). In the remaining denoising steps, we disable the reference view attention to avoid content invisible from the reference view $v_{\rm ref}$, from being impaired by enforcing attention to the reference view.

\subsection{Finalizing the Texture}
\label{subsec:rgbtexture}
Although we can denoise a latent texture to a noiseless state to obtain the final texture, we choose not to do so for two reasons. 
First, a latent texture generated through projecting screen-space latent to a texture is not viable for directly decoding to the final RGB texture, as stretching and rotation in the latent texture are likely to cause mismatches with the trained distribution of the decoder. 
Second, we observed that the sharpness of the generated results could drop during the last few denoising steps  (when high-frequency details are  forming) when Multi-View Diffusion is enforced.

Therefore, we choose to conduct the final phase of denoising in screen space, with self-attention reuse enabled. The final RGB texture can then be extracted from fully denoised views $Z_0=\{z^{(v_i)}_0\}_{i=1}^N$ through decoding, projecting to texture space, and aggregation. Note that although latent texture guarantees 3D consistency of latents, the high frequency details in the decoded images might not be strictly consistent at pixel level. 
To retain the high-frequency details in the unified output, we follow a similar cosine-weighted aggregation method as the one used for latent texture aggregation to obtain the final RGB texture. In all our experiments, we set $\alpha=6$ and disable the Voronoi-based filling  during the texture project for sharpness preservation.

\begin{figure*}[t!]
    \captionsetup[subfigure]{justification=centering}
    \centering
    \begin{subfigure}[t]{0.135\linewidth}
        \includegraphics[width=\linewidth]{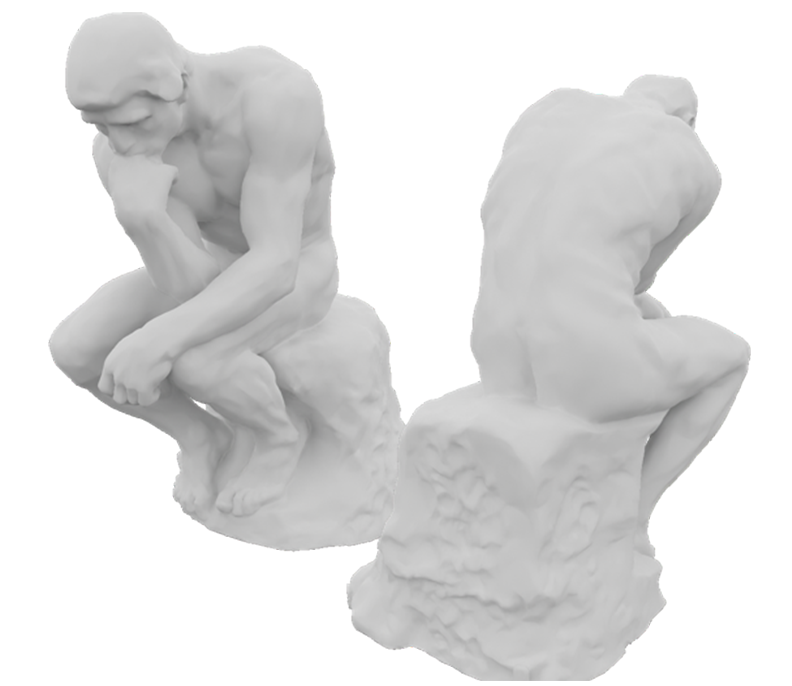}
        \includegraphics[width=\linewidth]{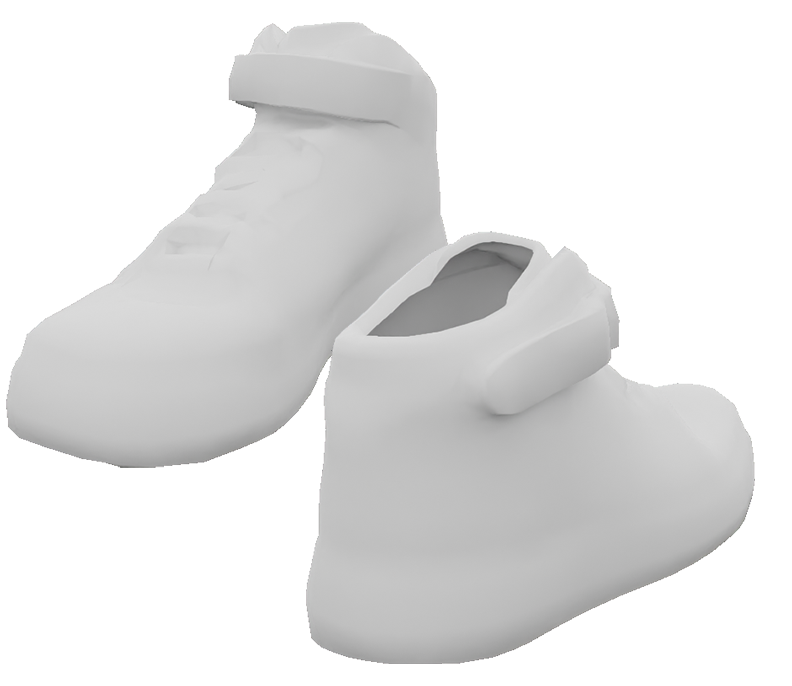}
        \includegraphics[width=\linewidth]{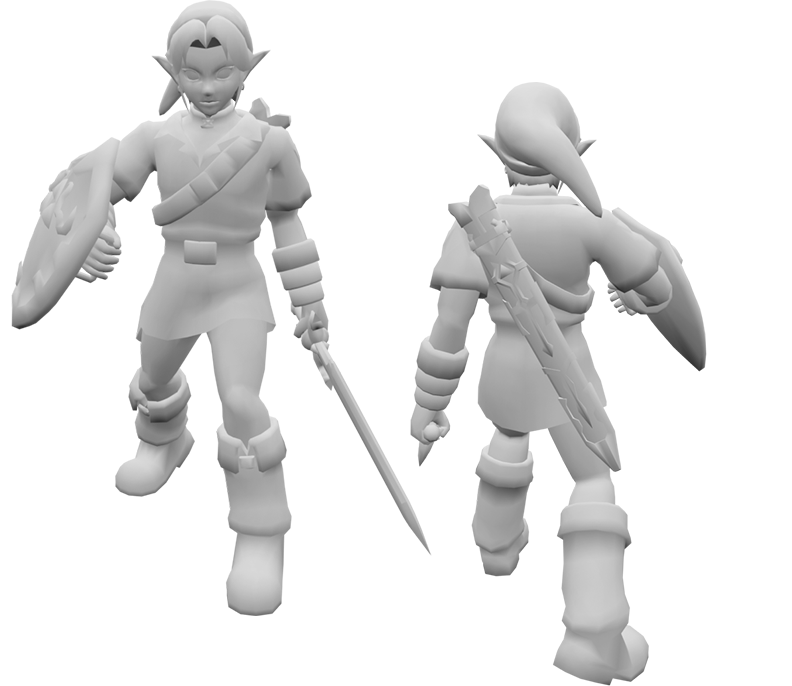}
        \includegraphics[width=\linewidth]{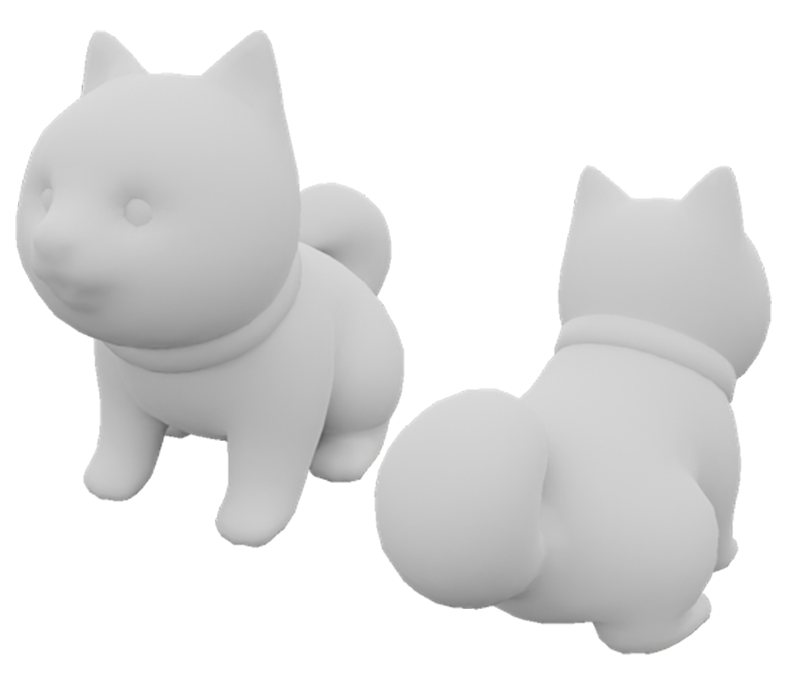}
        \includegraphics[width=\linewidth]{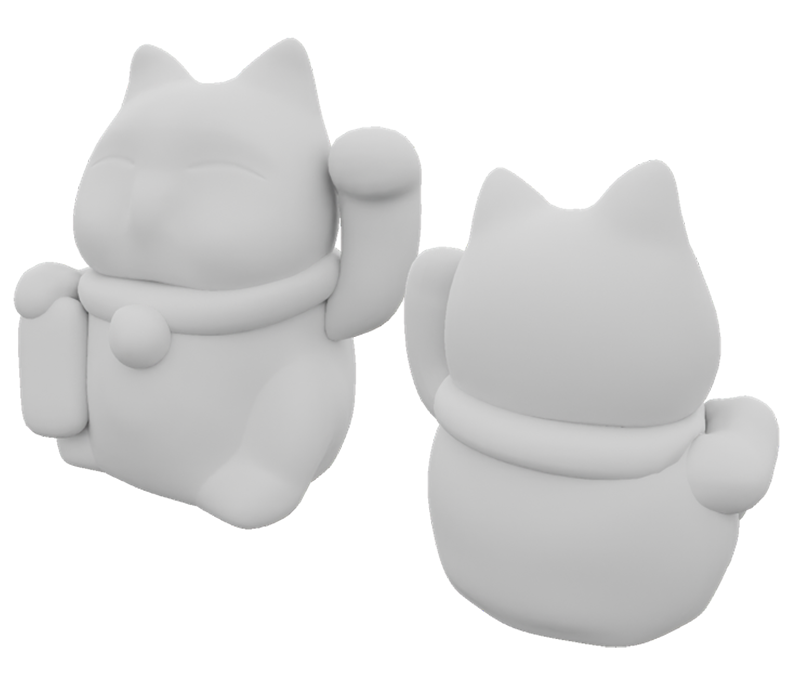}
        \caption{{\scriptsize Source mesh}}%
    \end{subfigure}
    \begin{subfigure}[t]{0.135\linewidth}
        \includegraphics[width=\linewidth]{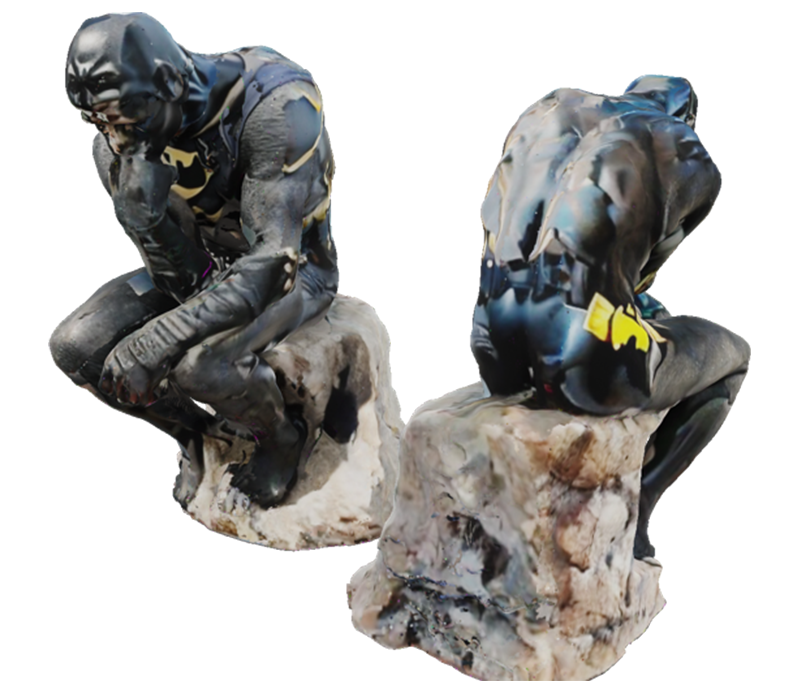}
        \includegraphics[width=\linewidth]{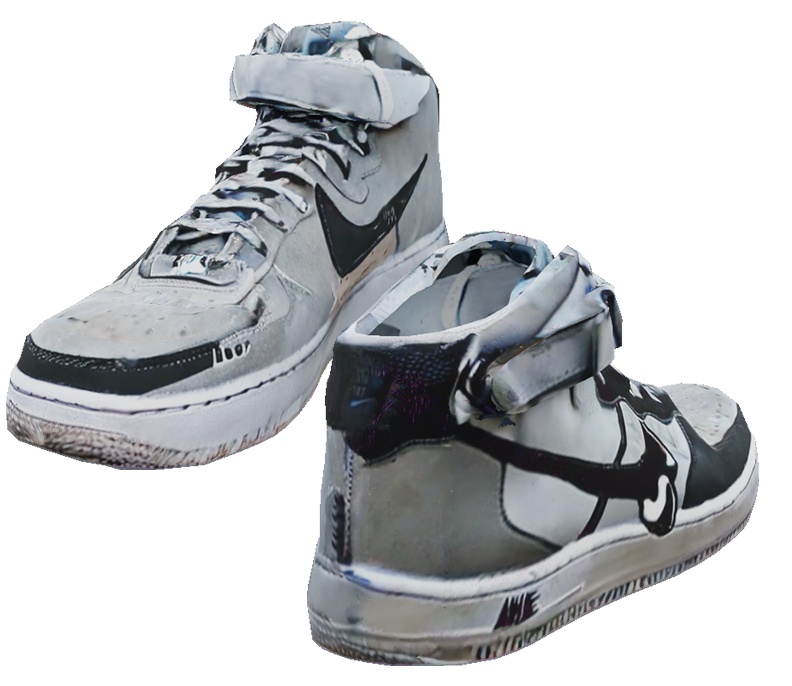}
        \includegraphics[width=\linewidth]{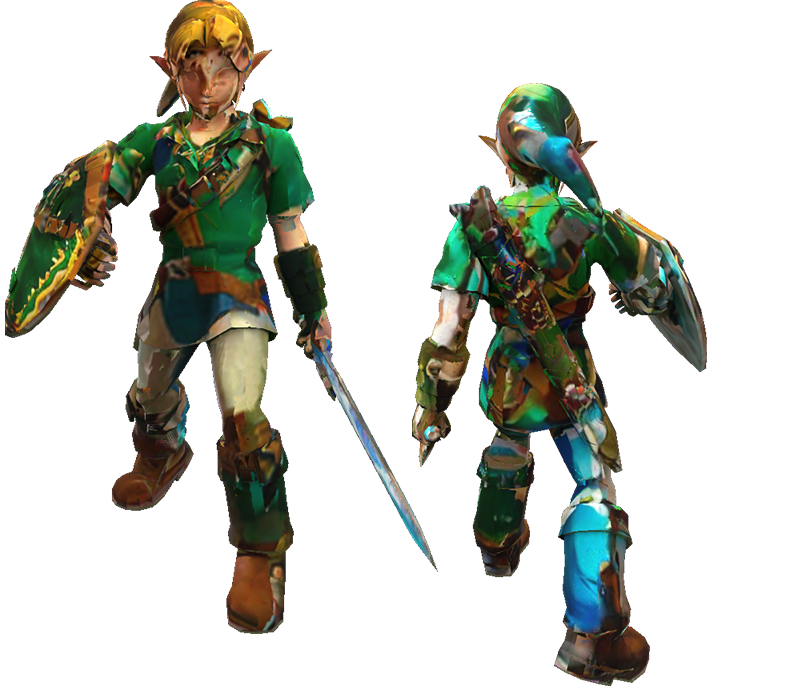}
        \includegraphics[width=\linewidth]{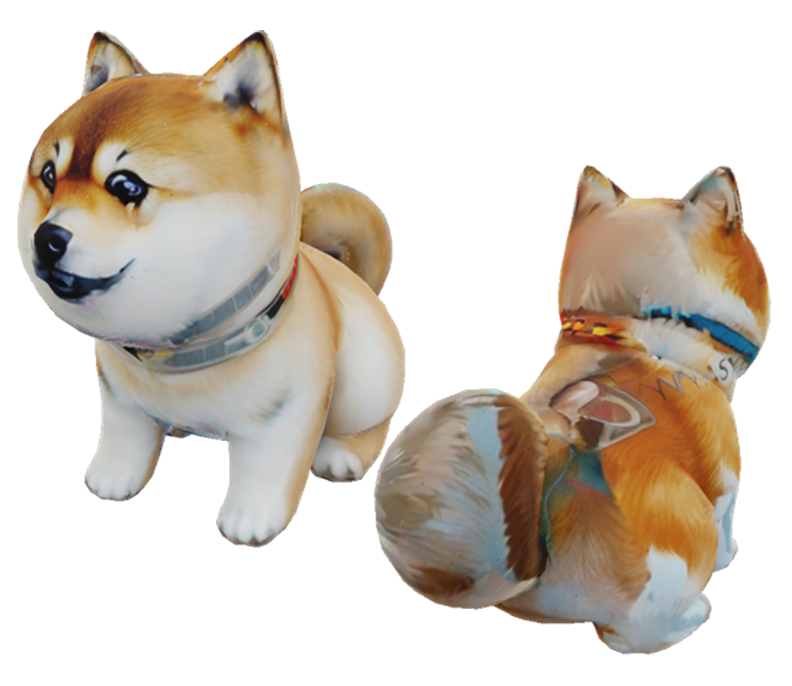}
        \includegraphics[width=\linewidth]{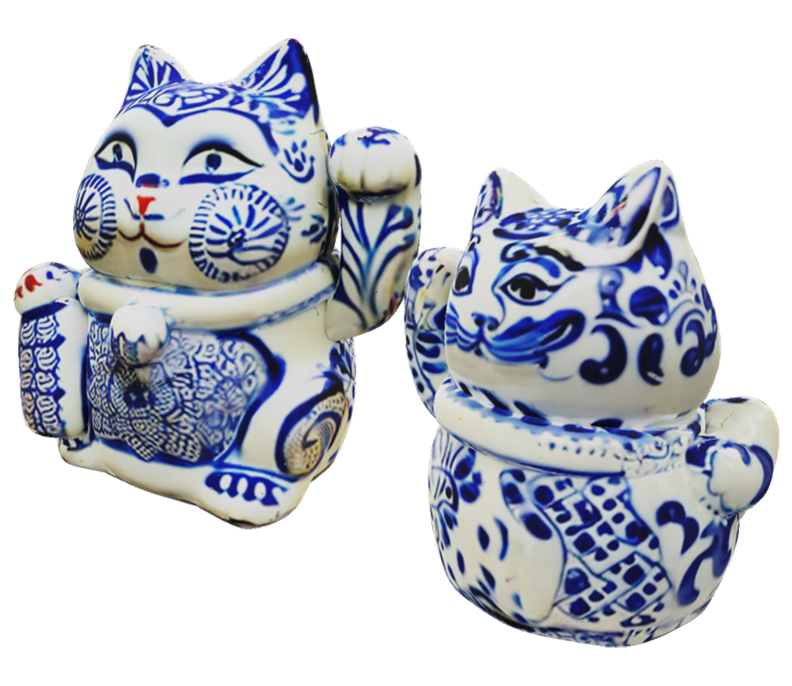}
        \caption{{\scriptsize TEXTure}}%
    \end{subfigure}
    \begin{subfigure}[t]{0.135\linewidth}
        \includegraphics[width=\linewidth]{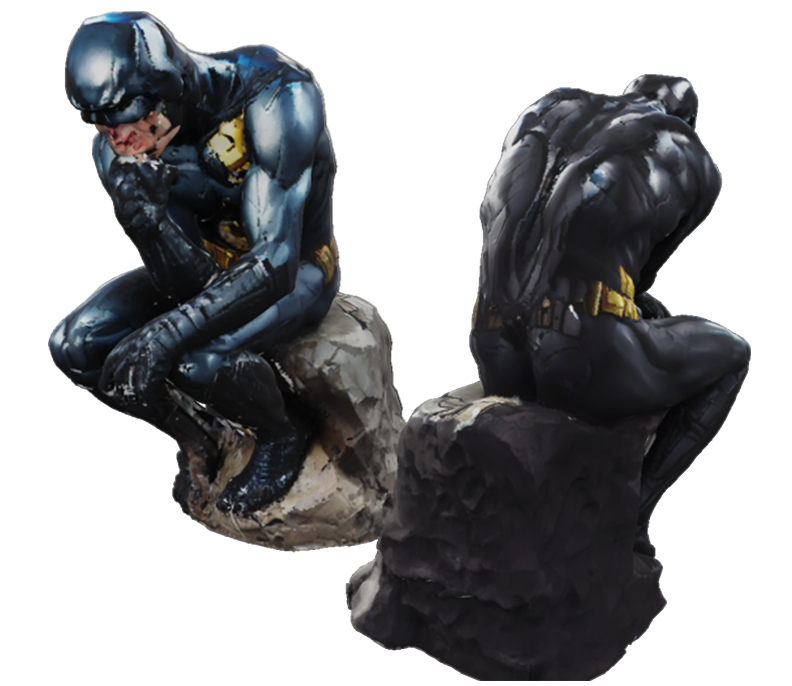}
        \includegraphics[width=\linewidth]{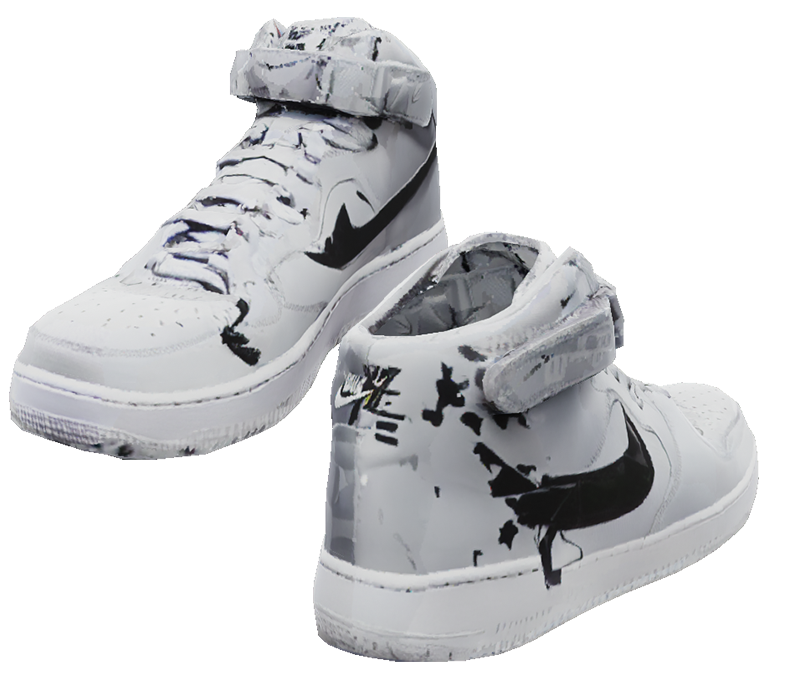}
        \includegraphics[width=\linewidth]{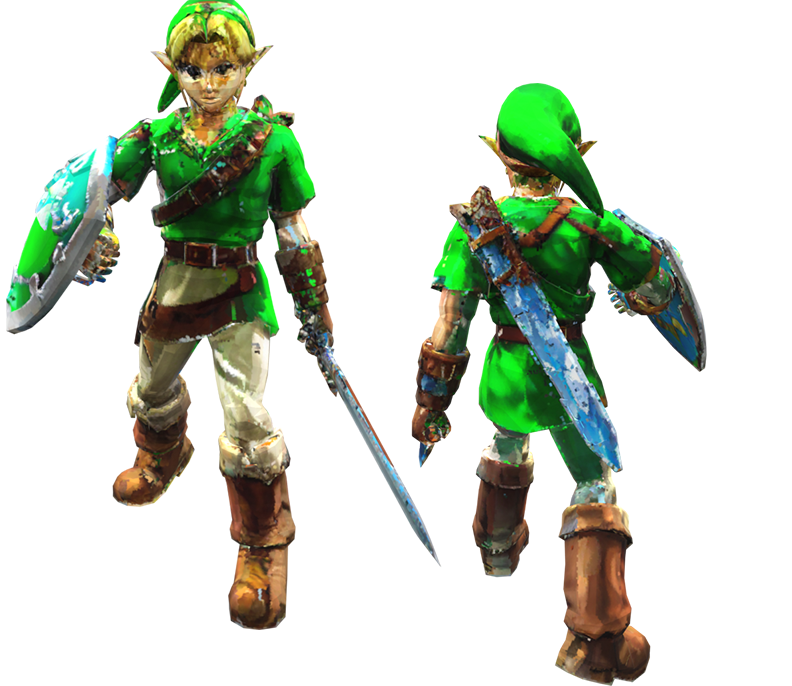}
        \includegraphics[width=\linewidth]{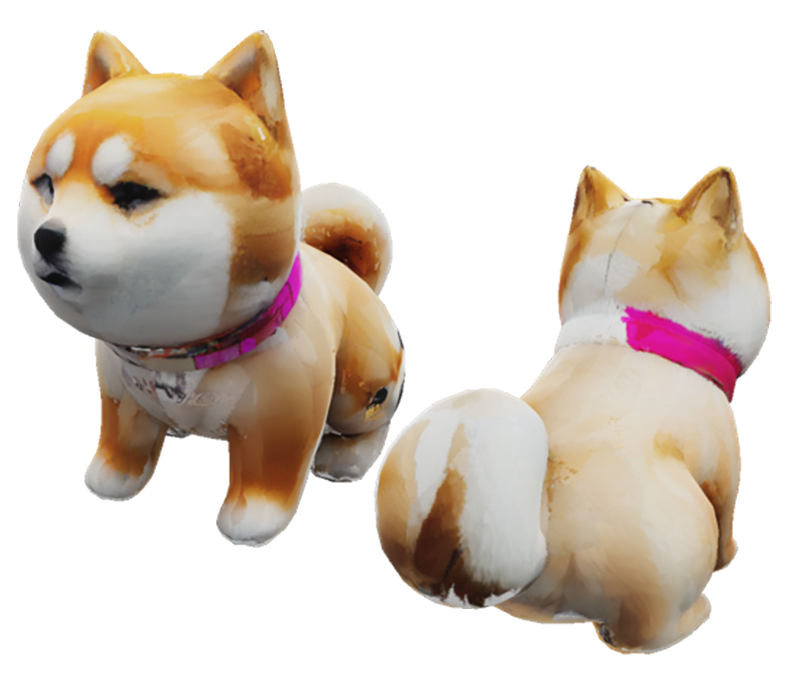}
        \includegraphics[width=\linewidth]{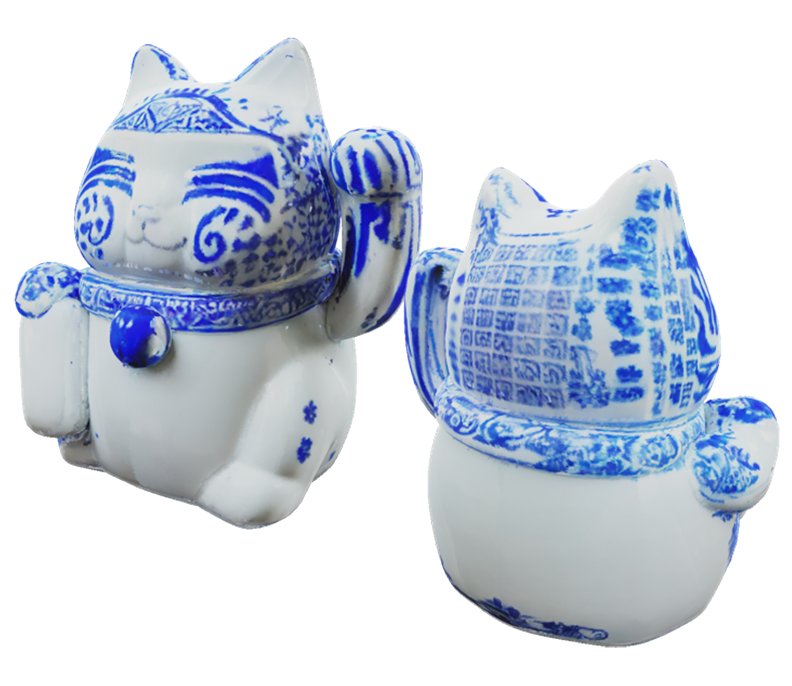}
        \caption{{\scriptsize Text2Tex}}%
    \end{subfigure}
    \begin{subfigure}[t]{0.135\linewidth}
        \includegraphics[width=\linewidth]{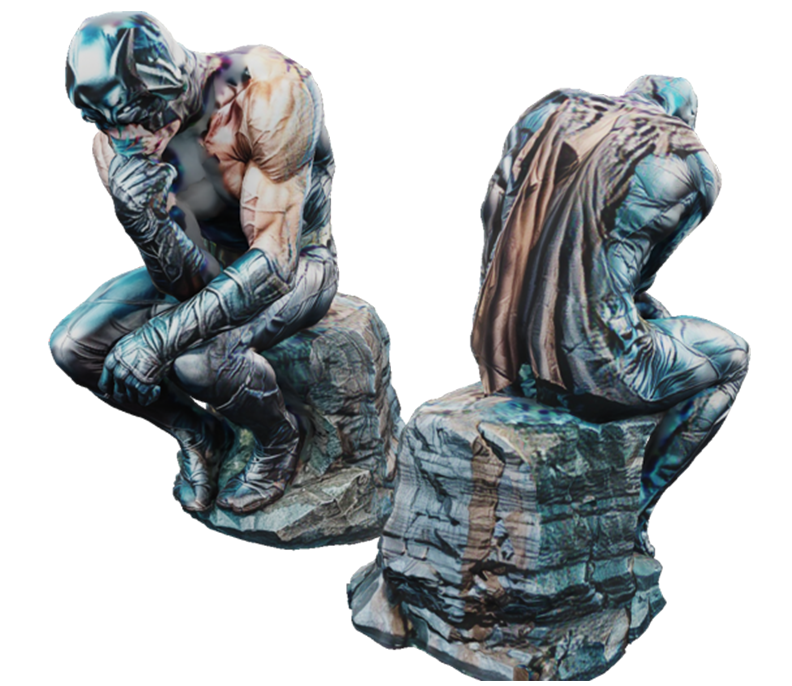}
        \includegraphics[width=\linewidth]{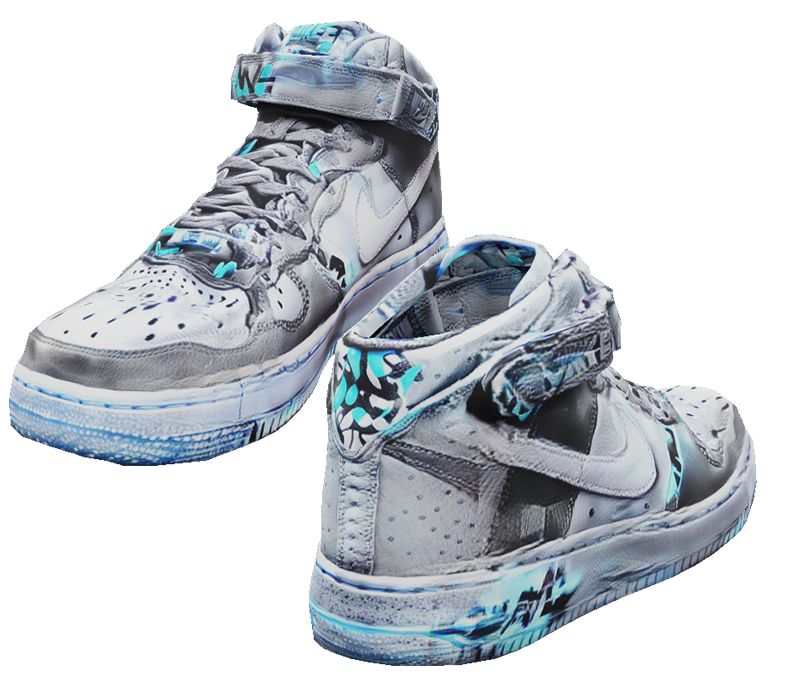}
        \includegraphics[width=\linewidth]{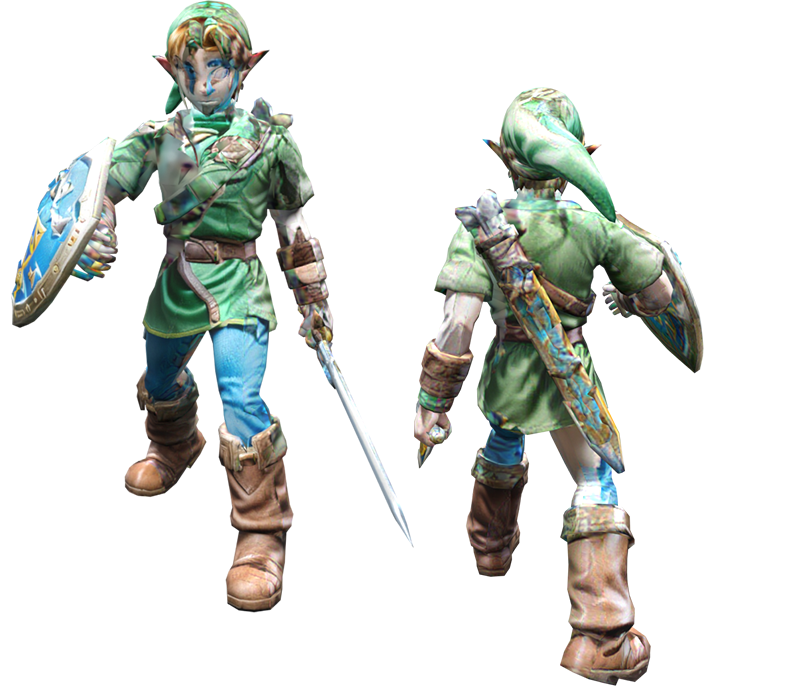}
        \includegraphics[width=\linewidth]{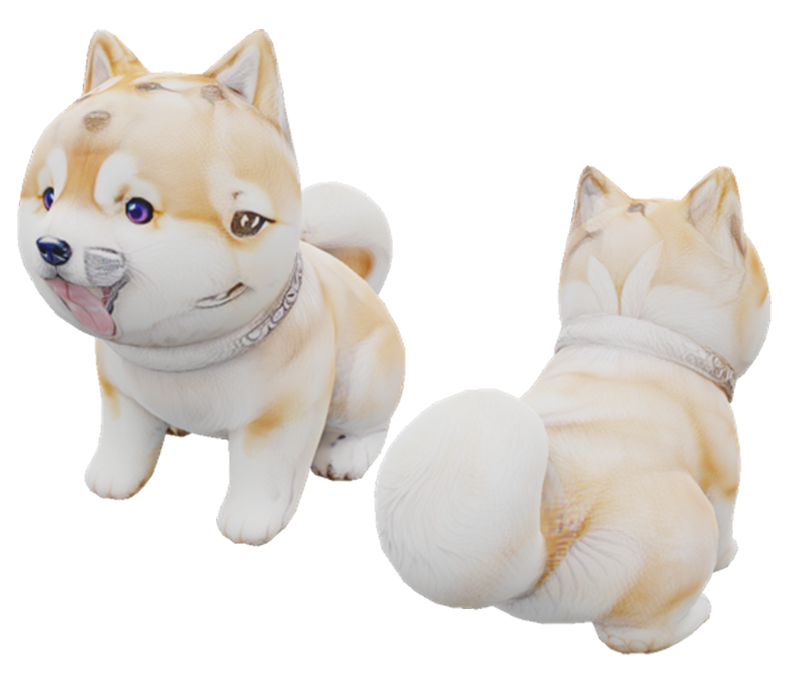}
        \includegraphics[width=\linewidth]{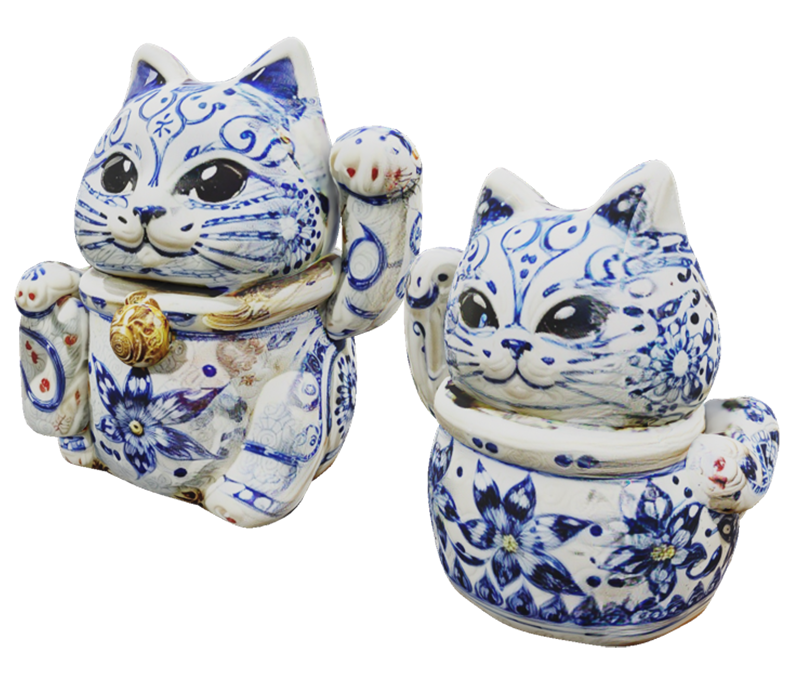}
        \caption{{\scriptsize Meshy}}%
    \end{subfigure}
    \begin{subfigure}[t]{0.135\linewidth}
        \includegraphics[width=\linewidth]{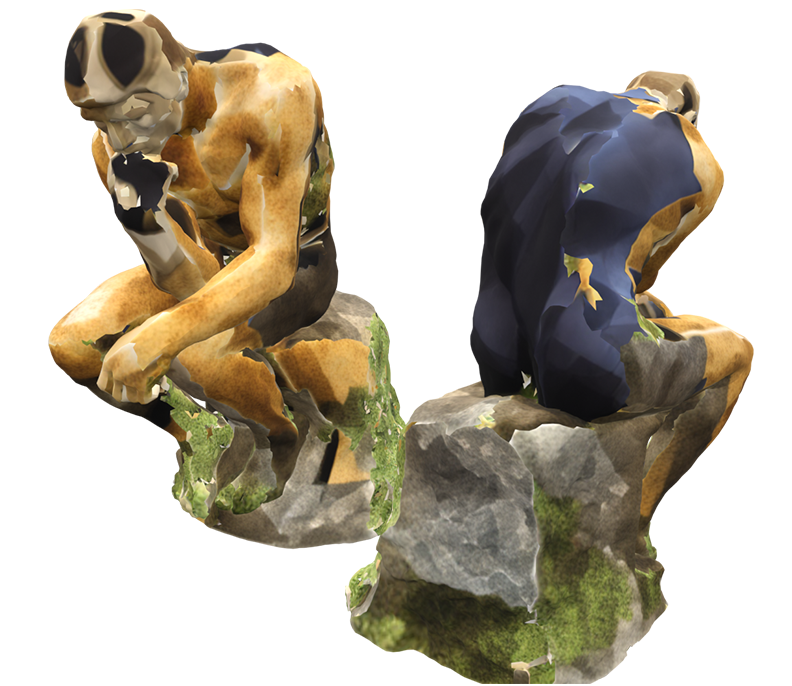}
        \includegraphics[width=\linewidth]{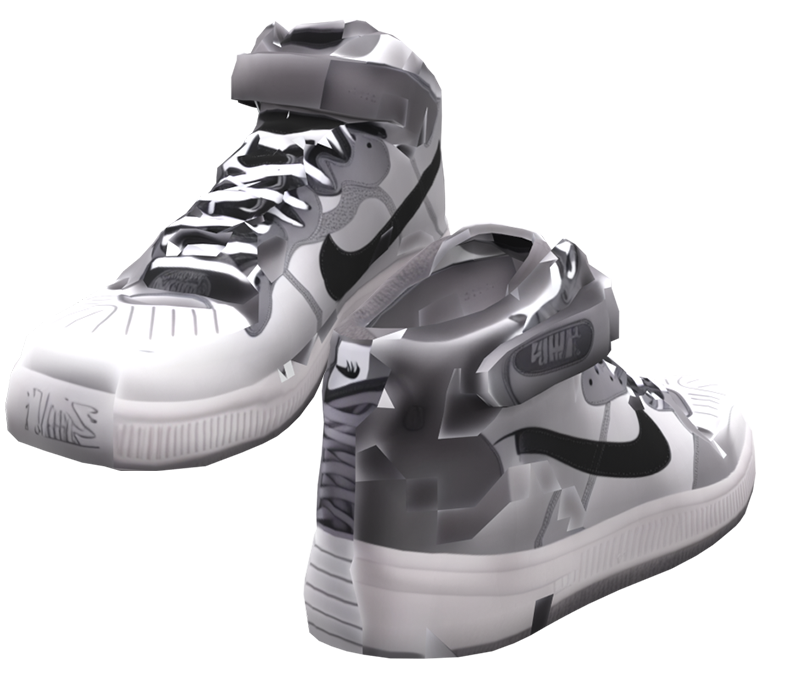}
        \includegraphics[width=\linewidth]{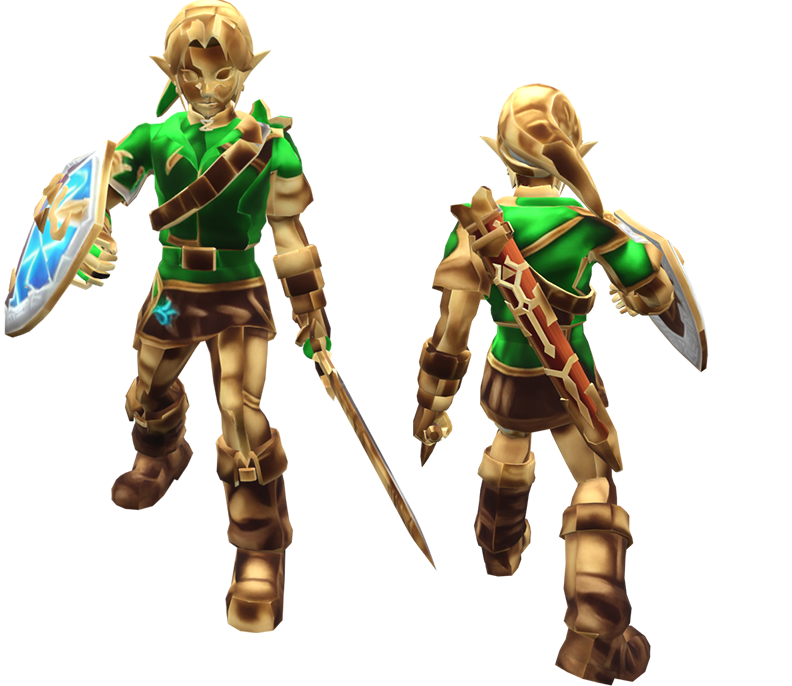}
        \includegraphics[width=\linewidth]{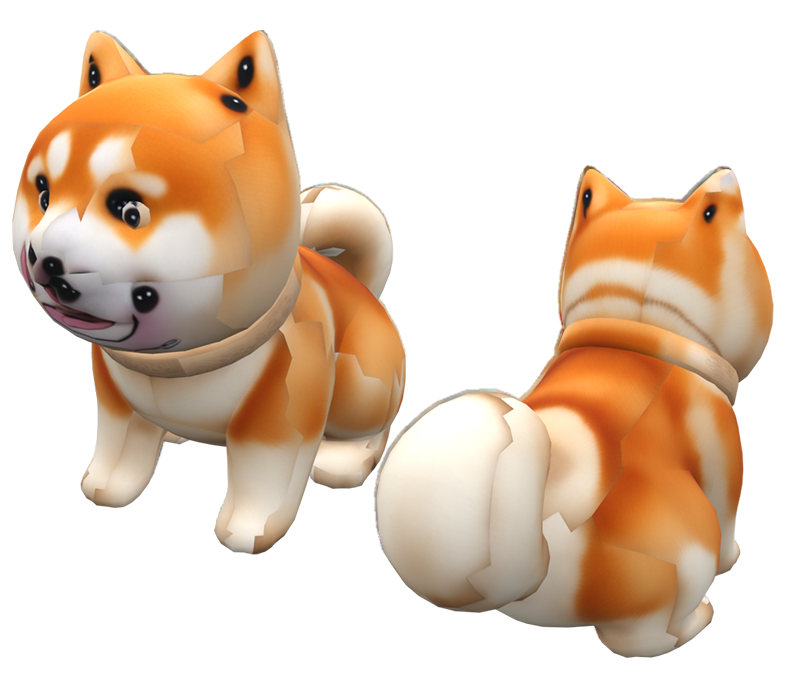}
        \includegraphics[width=\linewidth]{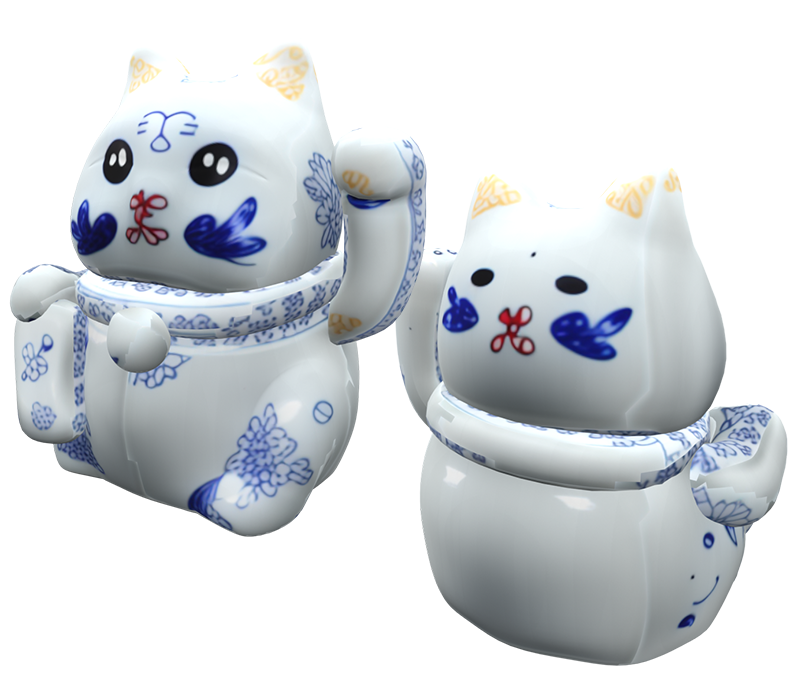}
        \caption{{\scriptsize Paint3D}}%
    \end{subfigure}
    \begin{subfigure}[t]{0.135\linewidth}
        \includegraphics[width=\linewidth]{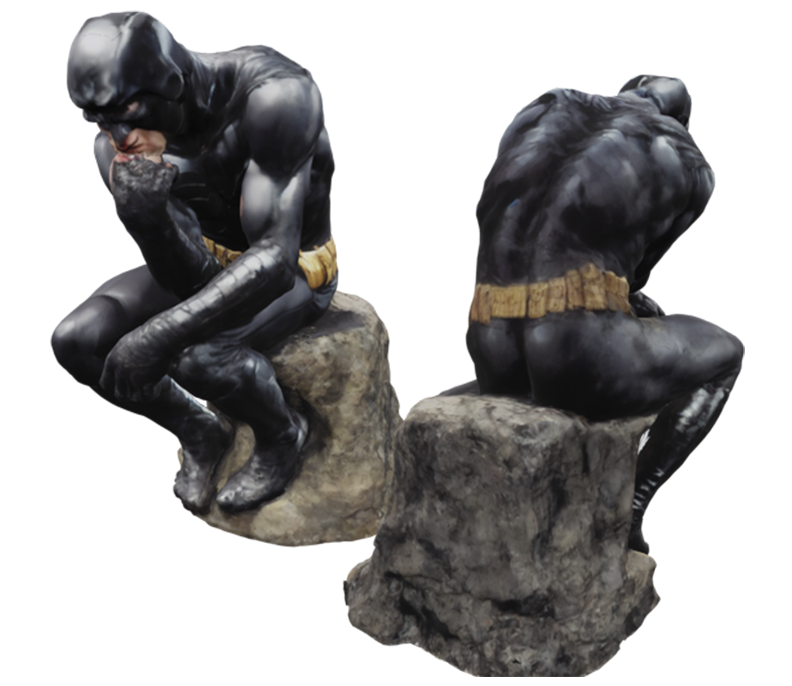}
        \includegraphics[width=\linewidth]{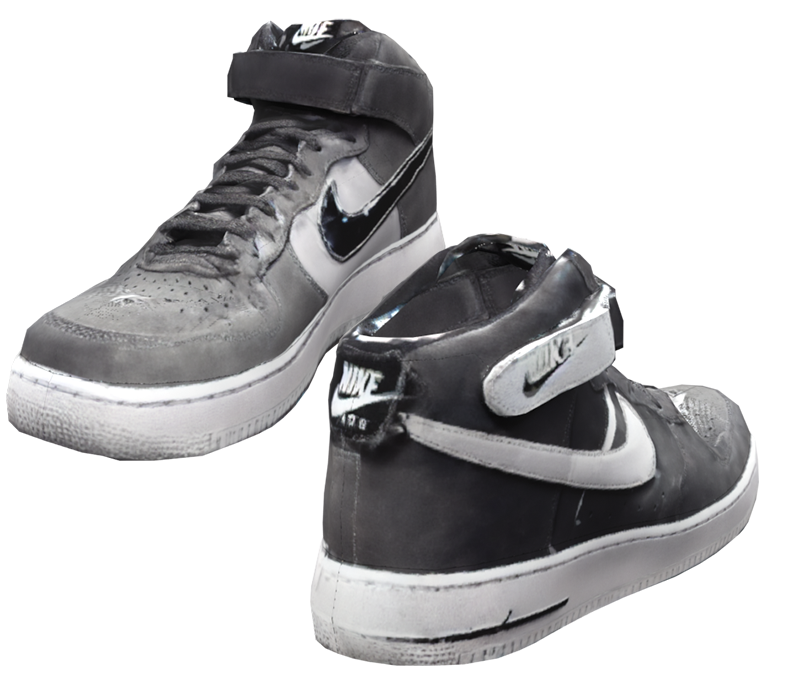}
        \includegraphics[width=\linewidth]{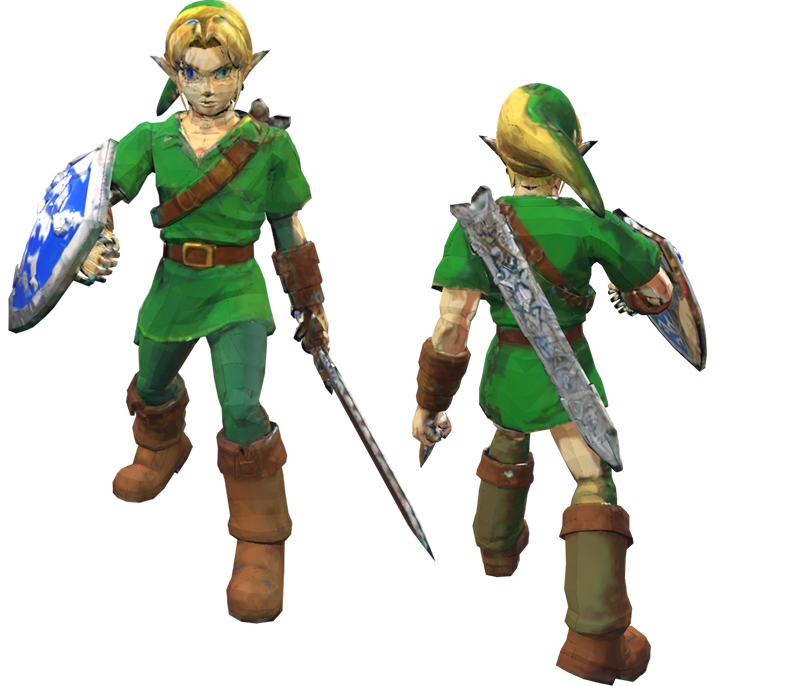}
        \includegraphics[width=\linewidth]{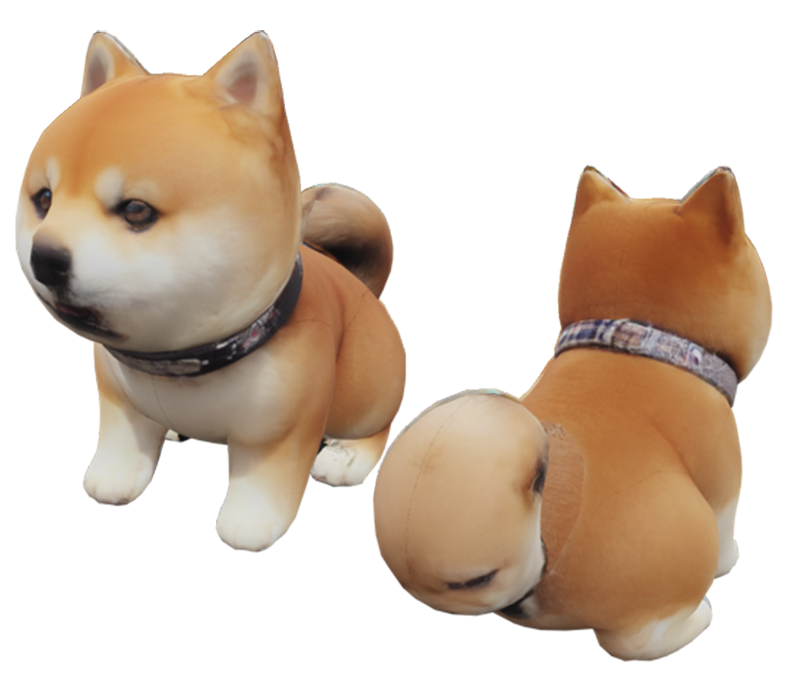}
        \includegraphics[width=\linewidth]{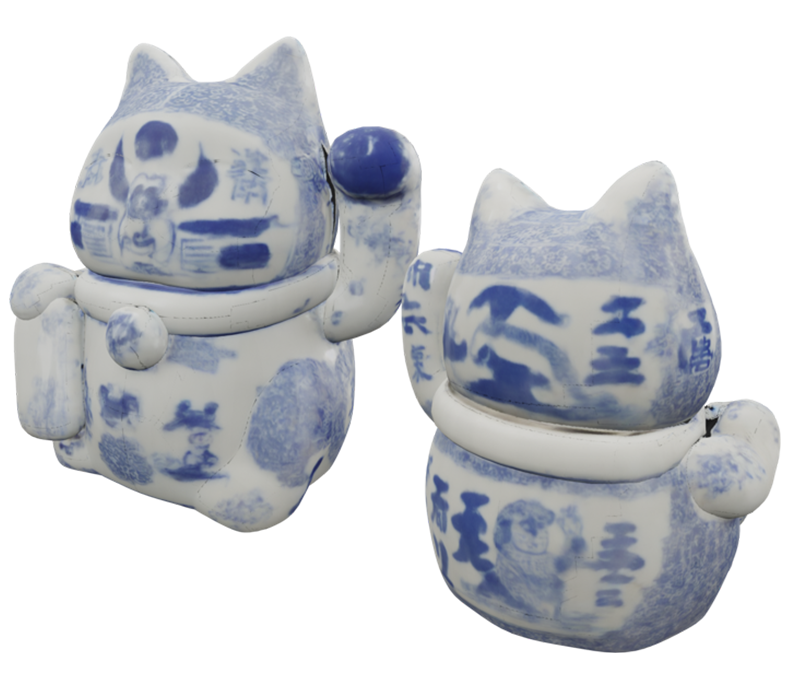}
        \caption{{\scriptsize TexFusion (sim)}}%
    \end{subfigure}
    \begin{subfigure}[t]{0.135\linewidth}
        \includegraphics[width=\linewidth]{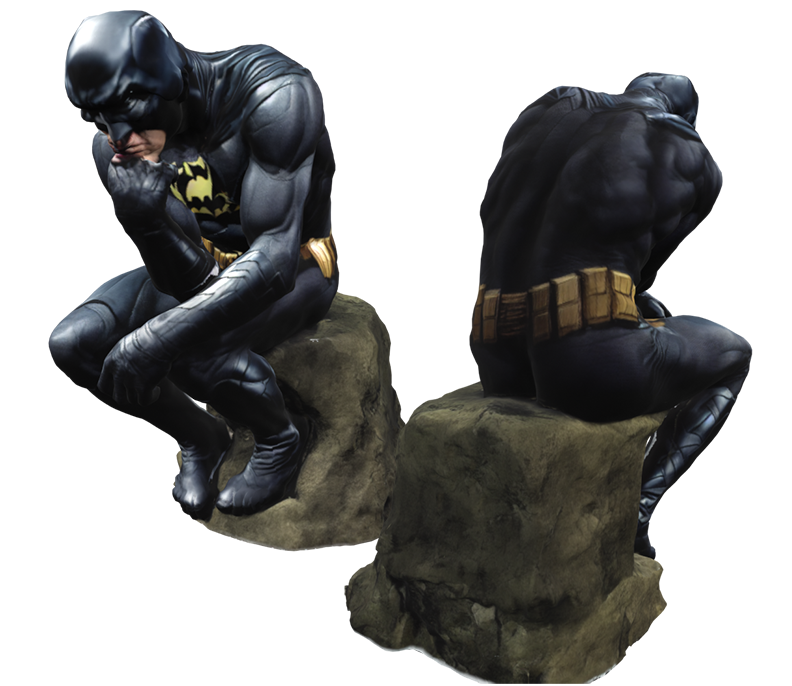}
        \includegraphics[width=\linewidth]{comparison/12_ours}
        \includegraphics[width=\linewidth]{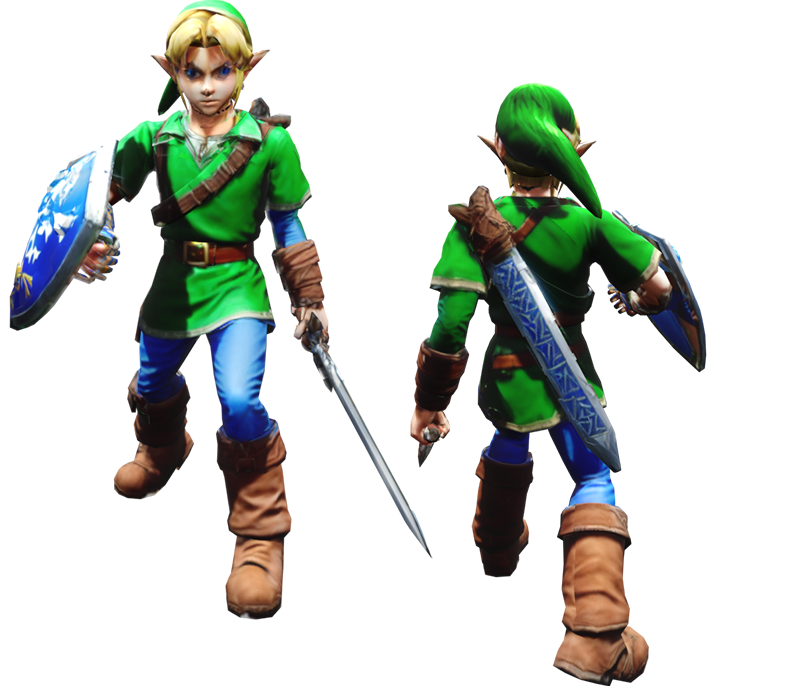}
        \includegraphics[width=\linewidth]{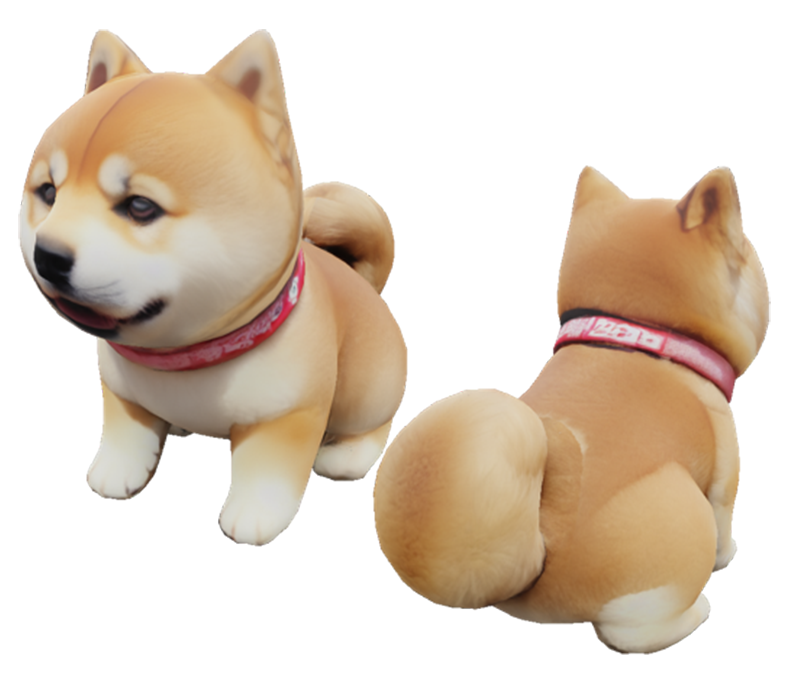}
        \includegraphics[width=\linewidth]{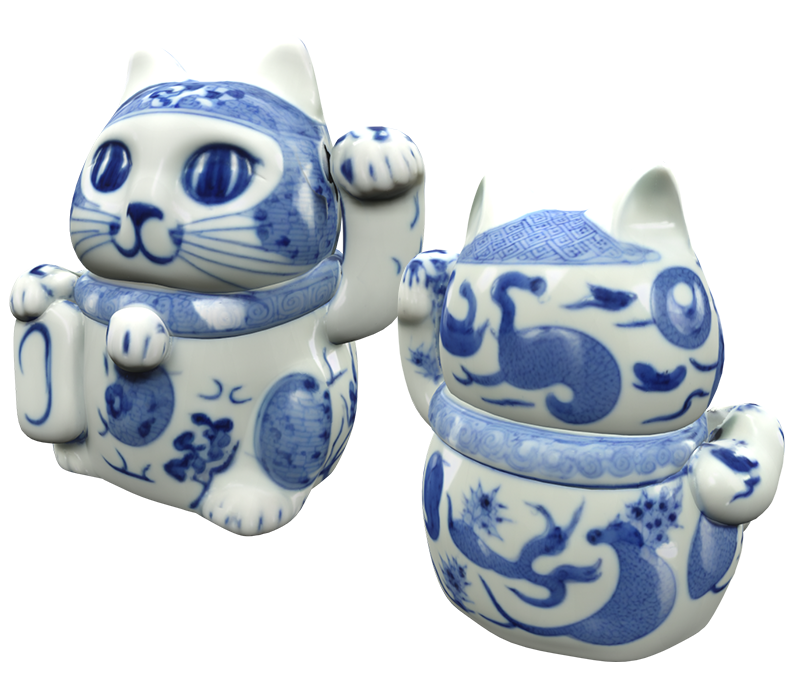}
        \caption{{\scriptsize Ours}}%
    \end{subfigure}
    
    \vspace{-0.2cm}
    % {\footnotesize \tt "A photo of a robot hand with mechanical joints."}
    \caption{Comparison of object texturing results. Text prompts from top to bottom: {{\tt\footnotesize "photo of Batman, sitting on a rock"}, {\tt\footnotesize "photo of a gray and black Nike airforce high top sneakers"}, {\tt\footnotesize "photo of link in the legend of zelda, photo-realistic, unreal 5"}, {\tt\footnotesize "A cute shiba inu dog"} and {\tt\footnotesize "blue and white pottery style lucky cat with intricate patterns"}. Readers are recommended to zoom in for better visualization and comparison. Results from LatentPaint are organized at the end of the paper due to space limit.}}
    \label{fig:illustrate}
    \vspace{-0.2cm}
\end{figure*}

\section{Results}
\label{sec:results}
\subsection{Implementation Details}
In all our experiments, we have 10 cameras (views) to cover the object. 
Eight orthographic cameras are placed along the equator surrounding the object with a 45$^\circ$ interval, and two  additional cameras at elevated locations pointing towards the top  of the object.
Each view has a latent resolution of 96$\times$96 to reduce the aliasing. The UV mappings are by default, automatically unwrapped using XAtlas \cite{xatlas2016} to avoid potential poor UV layout; the latent texture resolution is set by default to 512$\times$512 which is sufficient for most objects with a correct UV layout.  To encourage the model to generate views with the expected orientation, we append directional keyword (e.g., {\small\tt "front view"}) to the text prompt of each view automatically based on its camera position.
Our method takes around 60 to 150 seconds to denoise the above-mentioned views, depending on the number of denoising steps. 
We developed our diffusion pipeline based on Stable Diffusion v1-5 and ControlNet module v1-1 (normal and depth) of Huggingface Diffusers library, and our projection functions are implemented using Pytorch3D. 

\subsection{Comparison with Text-to-Texture Methods}
In this section, we compare our results with six state-of-the-art methods, including LatentPaint~\cite{metzer2023latent}, TEXTure~\cite{richardson2023texture}, Text2Tex~\cite{chen2023text2tex}, Meshy~\cite{meshy}, Paint3D~\cite{zeng2024paint3d}, and TexFusion~\cite{cao2023texfusion}. 

Since TexFusion \cite{cao2023texfusion} had not yet released its code and raw generation results when we prepared this paper, it is difficult to compare TexFusion with other methods under the same setting. Thus, we implemented a simulated version of TexFusion to facilitate comparison, which is used in the 
qualitative and quantitative evaluation. We need to make a disclaimer that the simulated result may \textbf{not} accurately reflect the exact performance of TexFusion. For each method mentioned above, we textured a given mesh with the same prompt, and rendered the mesh with texture produced by each method. Fig.~\ref{fig:illustrate} visually compares our results with these three methods. 

\begin{figure*}[hbtp]
    \centering
%       \captionsetup[subfigure]{justification=centering}
    \begin{subfigure}[t]{0.235\linewidth}
        \includegraphics[width=\linewidth]{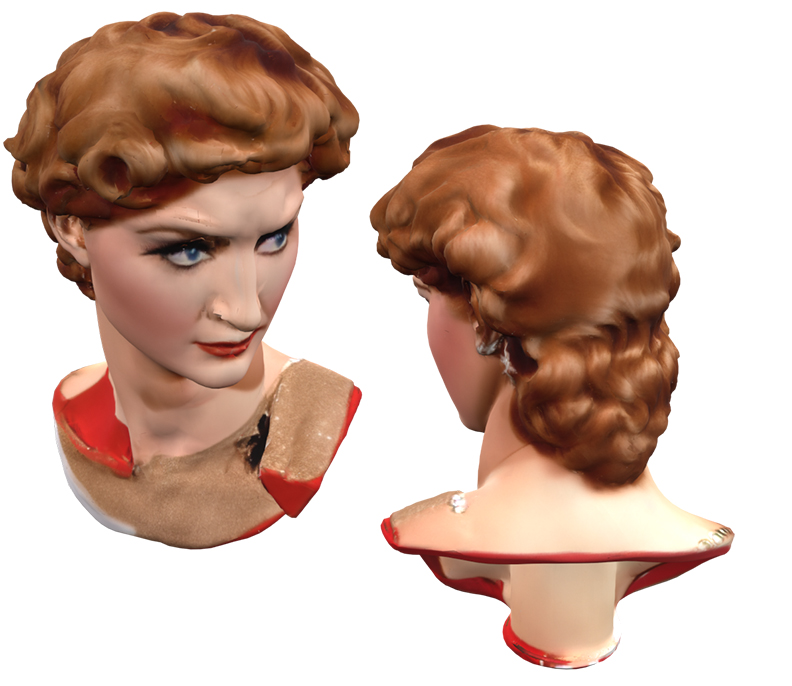}
        \caption*{{\scriptsize\tt "Publicity photo of a 60s movie, full color."}}%
    \end{subfigure}
    \begin{subfigure}[t]{0.235\linewidth}
        \includegraphics[width=\linewidth]{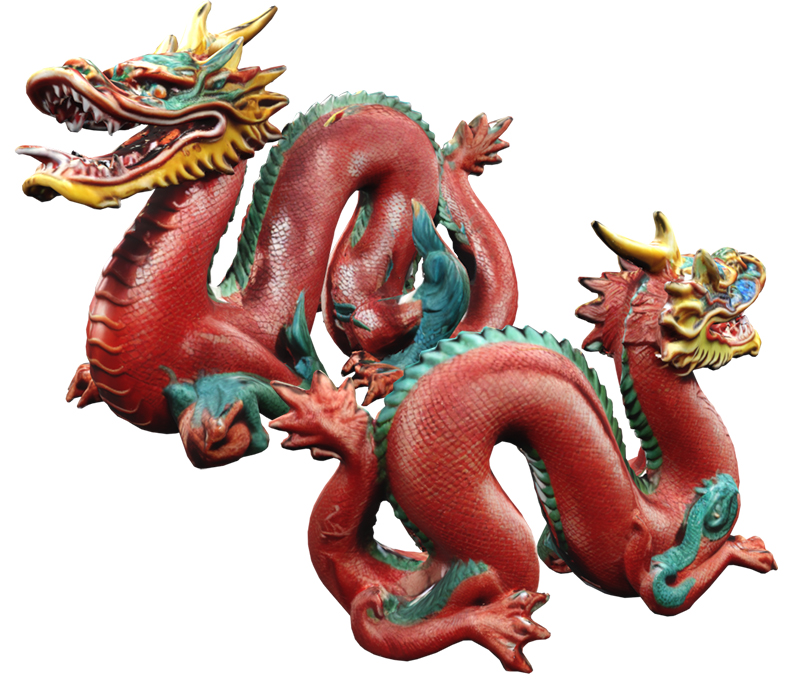}
        \caption*{{\scriptsize\tt "A photo of a Chinese dragon sculpture, glazed facing, vivid colors."}}%
    \end{subfigure}
    \begin{subfigure}[t]{0.235\linewidth}
        \includegraphics[width=\linewidth]{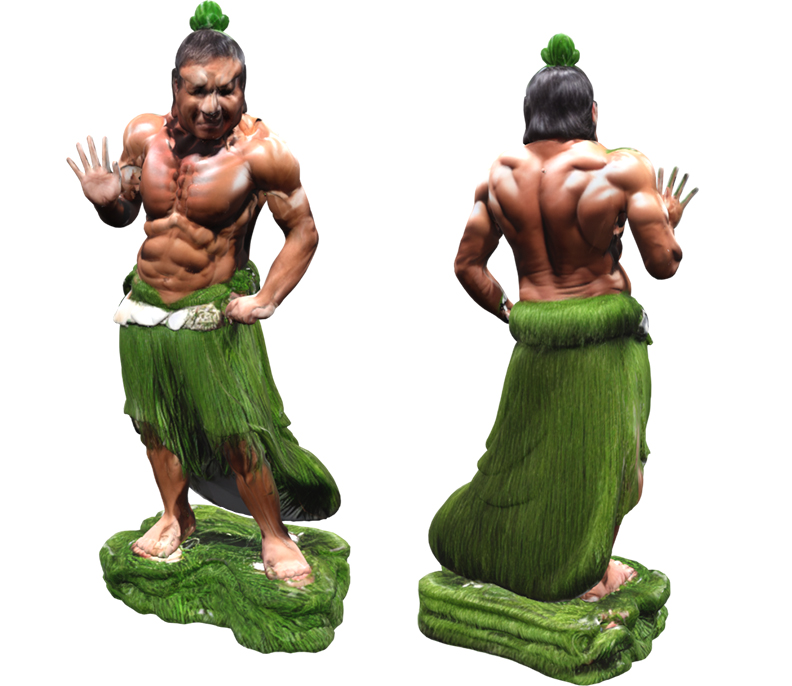}
        \caption*{{\scriptsize\tt "A muscular man wearing grass hula skirt."}}%
    \end{subfigure}
    \begin{subfigure}[t]{0.235\linewidth}
        \includegraphics[width=\linewidth]{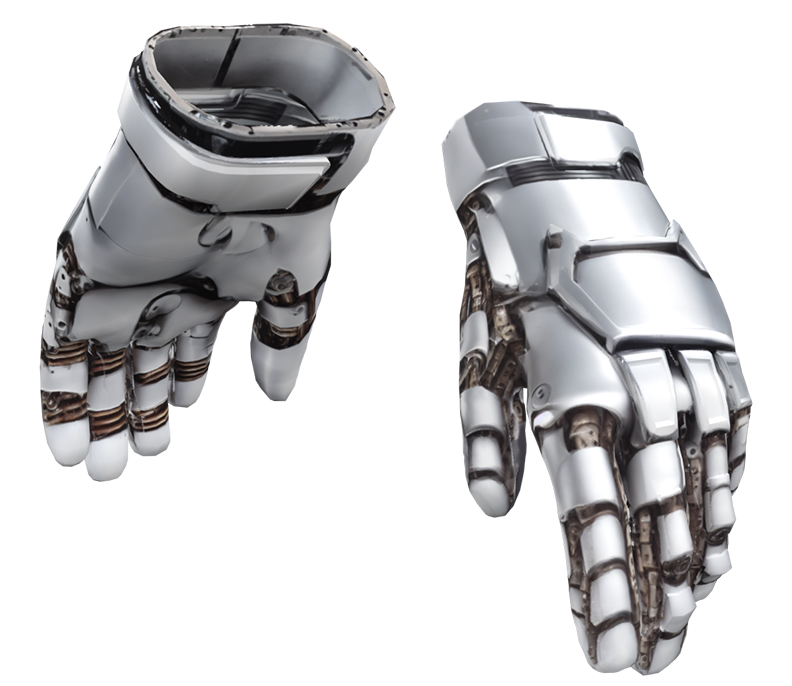}
        \caption*{{\scriptsize\tt "A photo of a robot hand with mechanical joints."}}%
    \end{subfigure}\\
    
    \begin{subfigure}[t]{0.235\linewidth}
        \includegraphics[width=\linewidth]{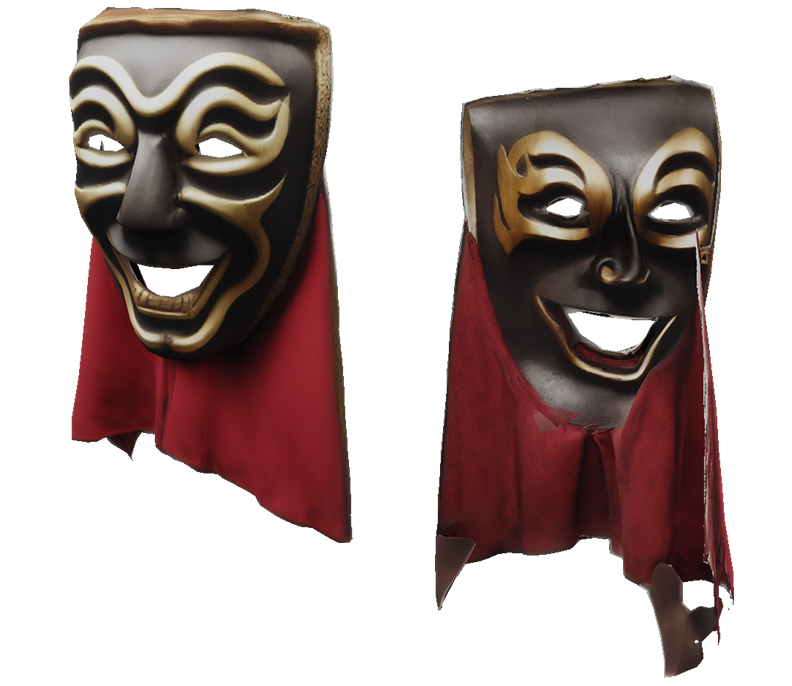}
        \caption*{{\scriptsize\tt "A Japanese demon mask."}}%
    \end{subfigure}
    \begin{subfigure}[t]{0.235\linewidth}
        \includegraphics[width=\linewidth]{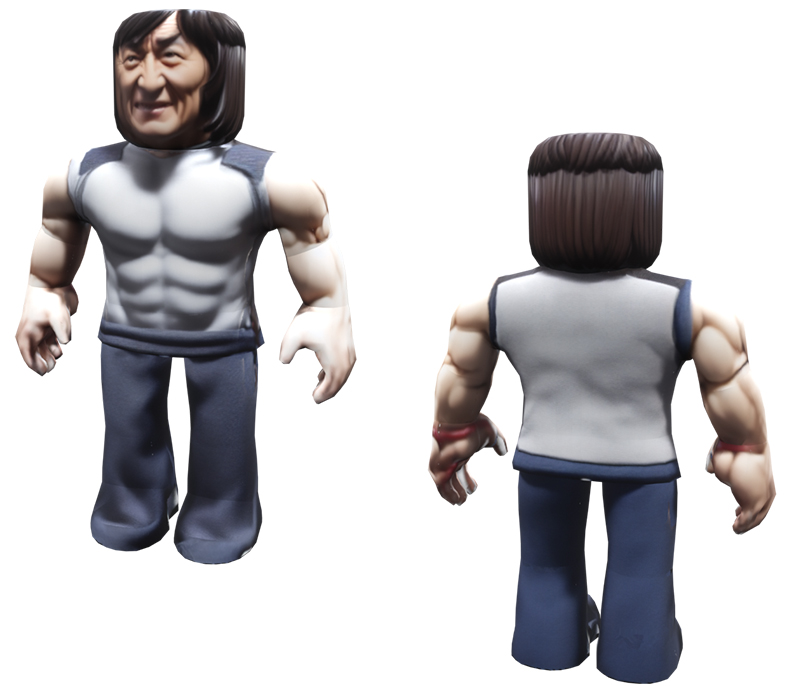}
        \caption*{{\scriptsize\tt "A Jackie Chan figure."}}%
    \end{subfigure}
    \begin{subfigure}[t]{0.235\linewidth}
        \includegraphics[width=\linewidth]{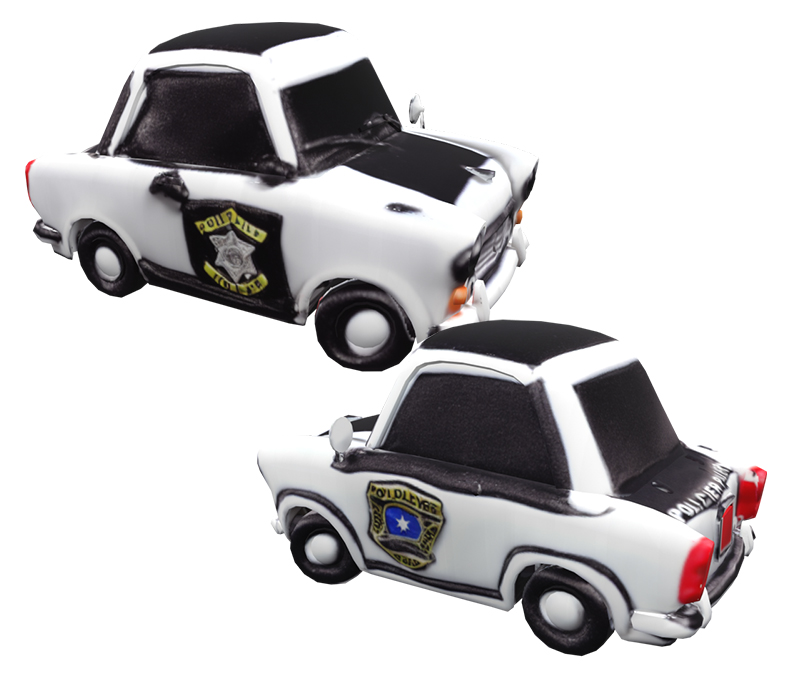}
        \caption*{{\scriptsize\tt "A photo of a polymer clay toy police car, black and white livery."}}%
    \end{subfigure}
    \begin{subfigure}[t]{0.235\linewidth}
        \includegraphics[width=\linewidth]{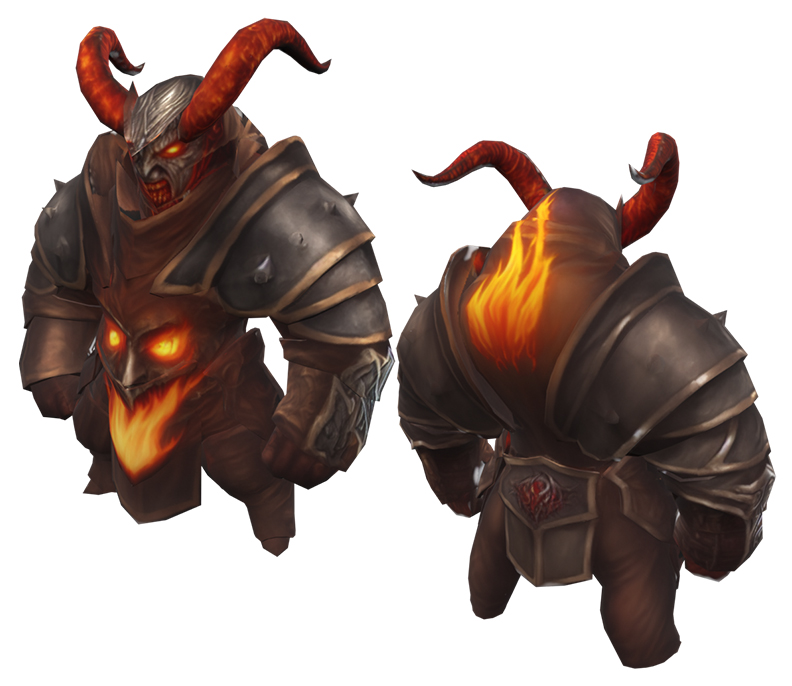}
        \caption*{{\scriptsize\tt "A photo of a demon knight, flame in eyes, warcraft style."}}%
    \end{subfigure}\\
    
    \begin{subfigure}[t]{0.235\linewidth}
        \includegraphics[width=\linewidth]{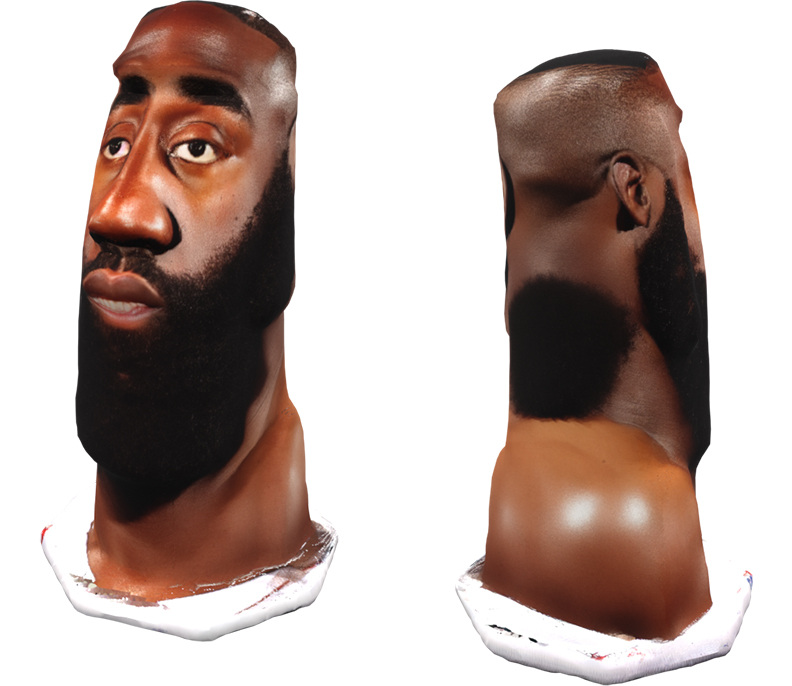}
        \caption*{{\scriptsize\tt "A photo of James Harden."}}%
    \end{subfigure}
    \begin{subfigure}[t]{0.235\linewidth}
        \includegraphics[width=\linewidth]{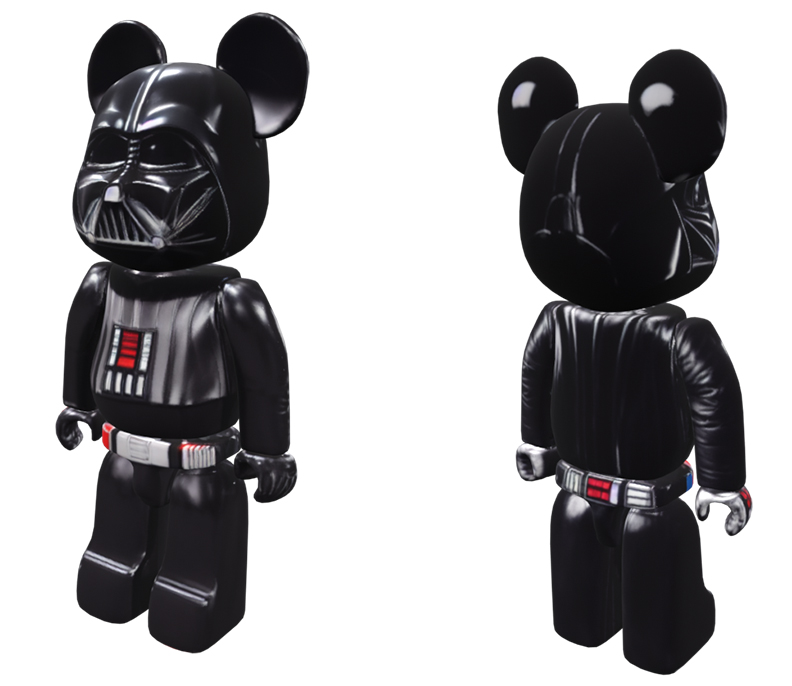}
        \caption*{{\scriptsize\tt "A photo of Darth Vader."}}%
    \end{subfigure}
    \begin{subfigure}[t]{0.235\linewidth}
        \includegraphics[width=\linewidth]{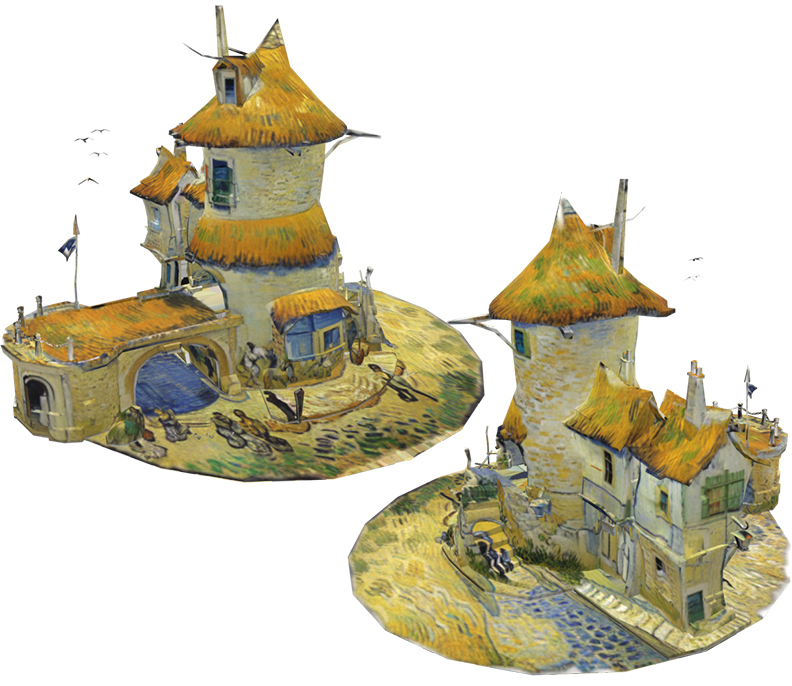}
        \caption*{{\scriptsize\tt "A beautiful oil paint of a stone building in Van Gogh style."}}%
    \end{subfigure}
     \begin{subfigure}[t]{0.235\linewidth}
        \includegraphics[width=\linewidth]{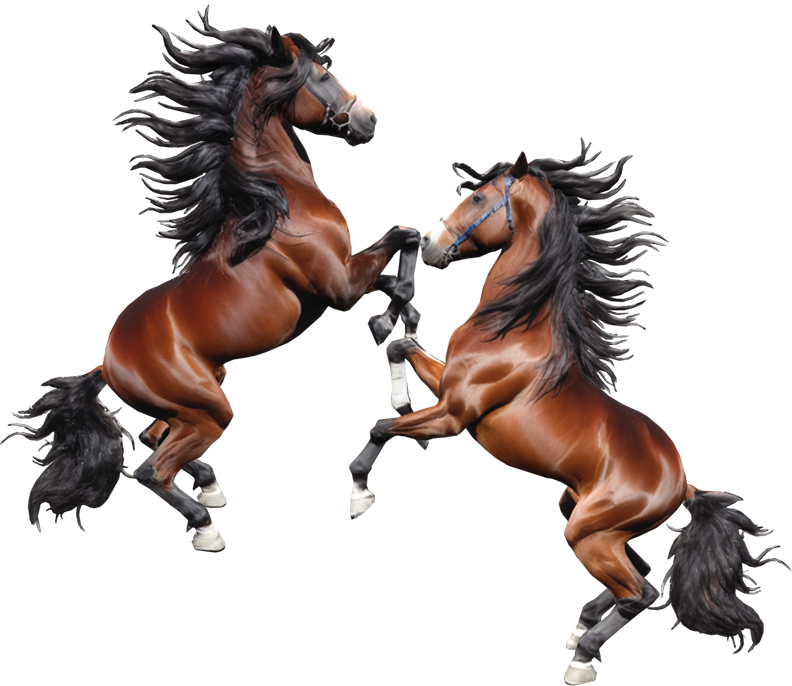}
        \caption*{{\scriptsize\tt "A photo of a horse."}}%
    \end{subfigure}\\

     \begin{subfigure}[t]{0.235\linewidth}
        \includegraphics[width=\linewidth]{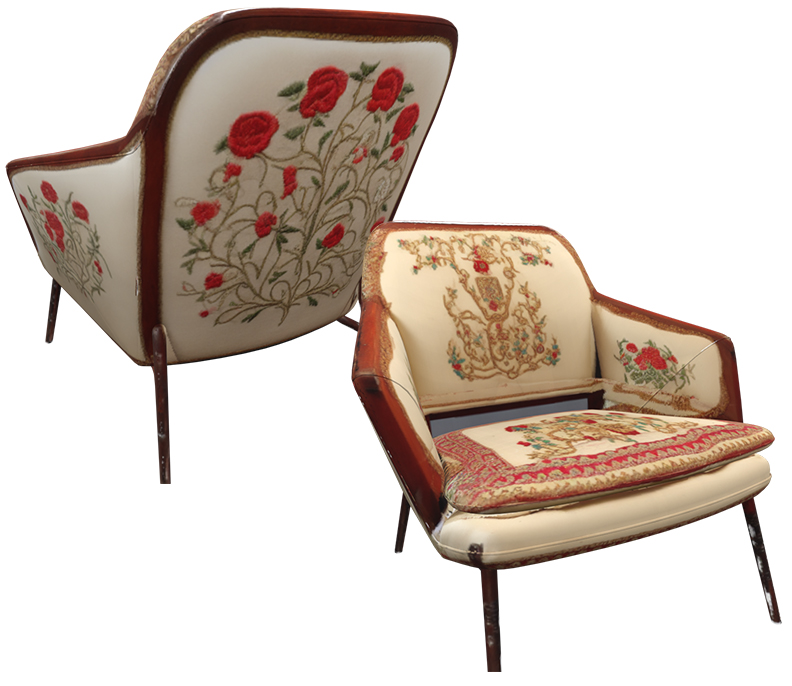}
        \caption*{{\scriptsize\tt "A photo of an beautiful embroidered seat with royal patterns"}}%
    \end{subfigure}
    \begin{subfigure}[t]{0.235\linewidth}
        \includegraphics[width=\linewidth]{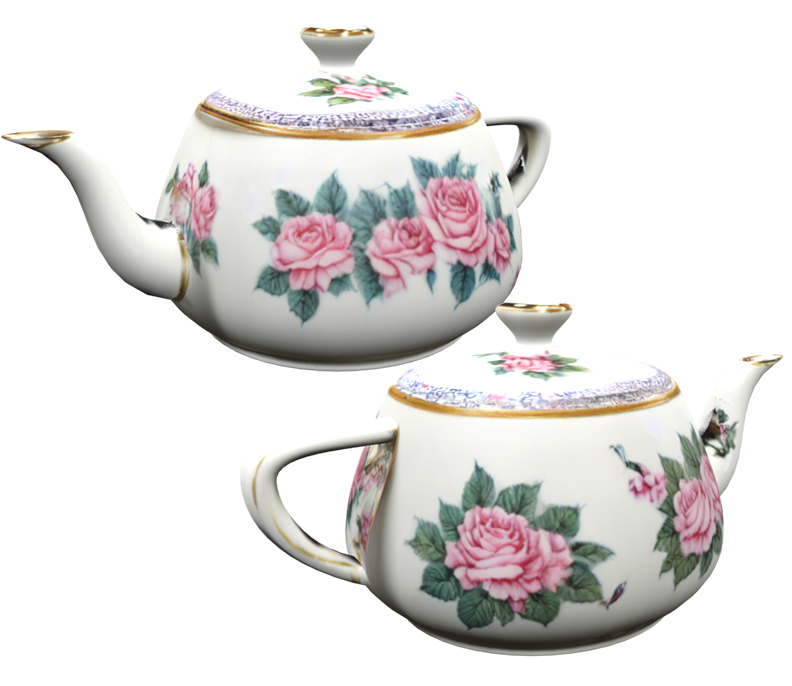}
        \caption*{{\scriptsize\tt "A photo of a beautiful chintz glided teapot."}}%
    \end{subfigure}
    \begin{subfigure}[t]{0.235\linewidth}
        \includegraphics[width=\linewidth]{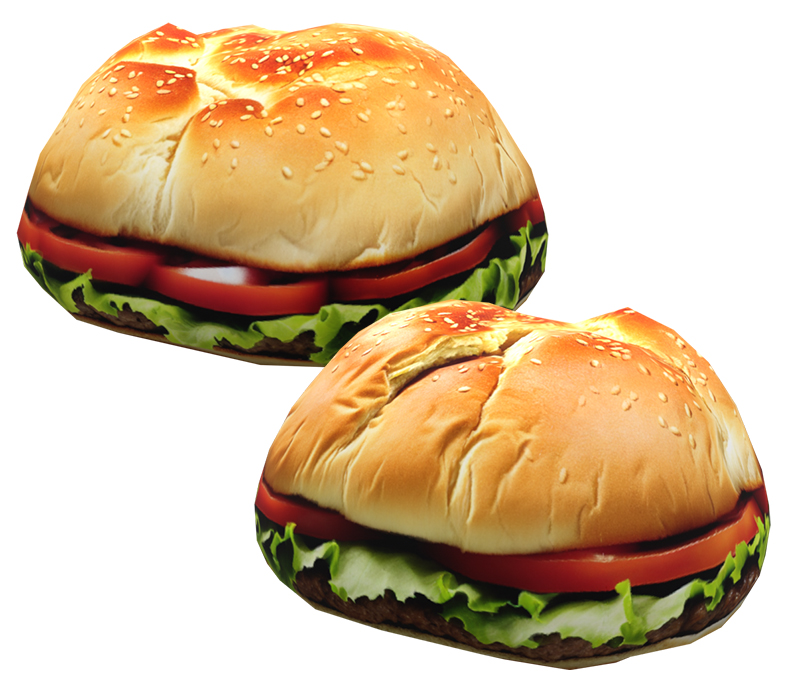}
        \caption*{{\scriptsize\tt "A photo of a hamburger."}}%
    \end{subfigure}
    \begin{subfigure}[t]{0.235\linewidth}
        \includegraphics[width=\linewidth]{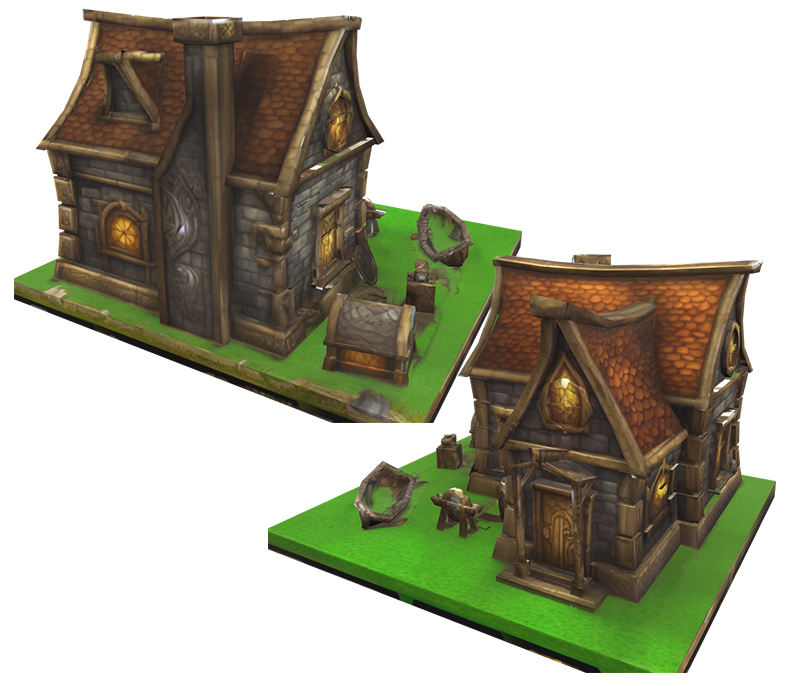}
        \caption*{{\scriptsize\tt "A photo of a lowpoly fantasy house from warcraft game."}}%
    \end{subfigure}
    \vspace{-.1in}
    \caption{Gallery of objects textured by our method. Corresponding text prompt is underneath each textured object.}
    \label{fig:gallery}
    % \makebox[height=\textheight]{}
    \vspace*{2cm}
\end{figure*}

More results of our method are shown in Fig.~\ref{fig:gallery} to illustrate the effectiveness and generalizability of our method. Readers are referred to the supplementary material for high resolution videos, including visual results in Fig.~\ref{fig:illustrate}, Fig.~\ref{fig:gallery} and several additional results not included in the main paper.

% This figures are 0.235 width before adjustment
% \newpage

\begin{figure*}[!ht]
    \centering
    \vspace{-0.3cm}
    \begin{subfigure}[t]{0.5\linewidth}
    \includegraphics[width=1.\linewidth]{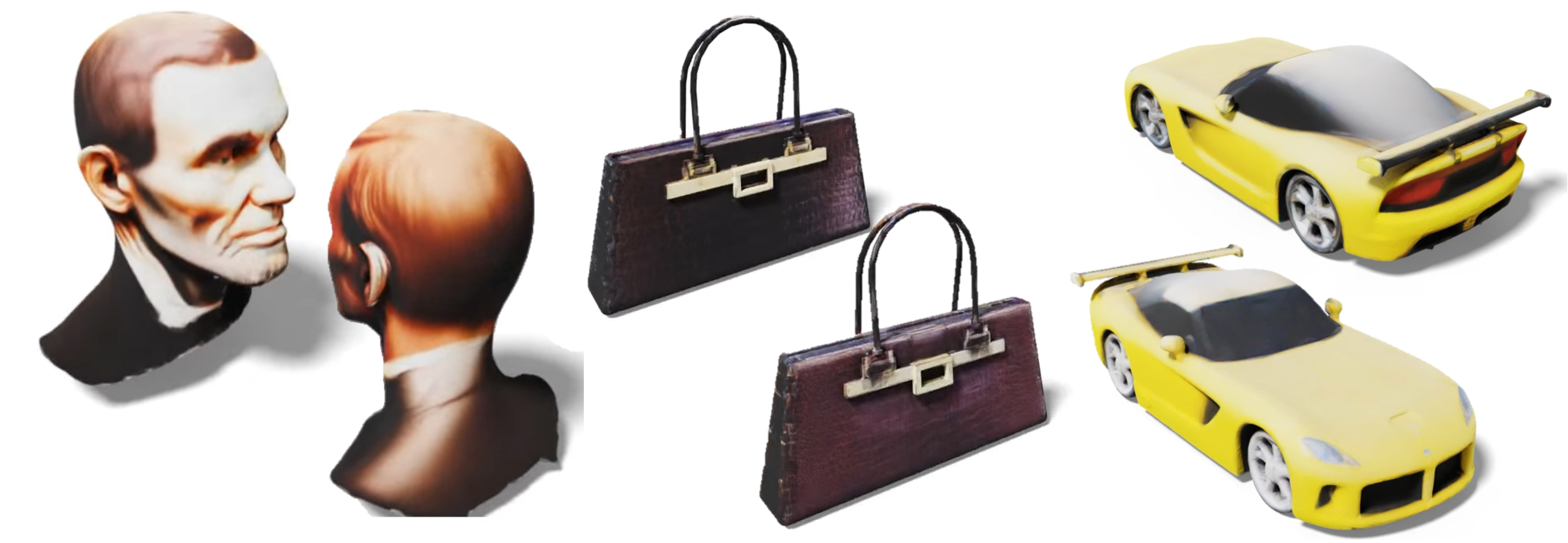}
    \caption{TexFusion}
    \end{subfigure}
    \begin{subfigure}[t]{0.48\linewidth}
    \includegraphics[width=1.\linewidth]{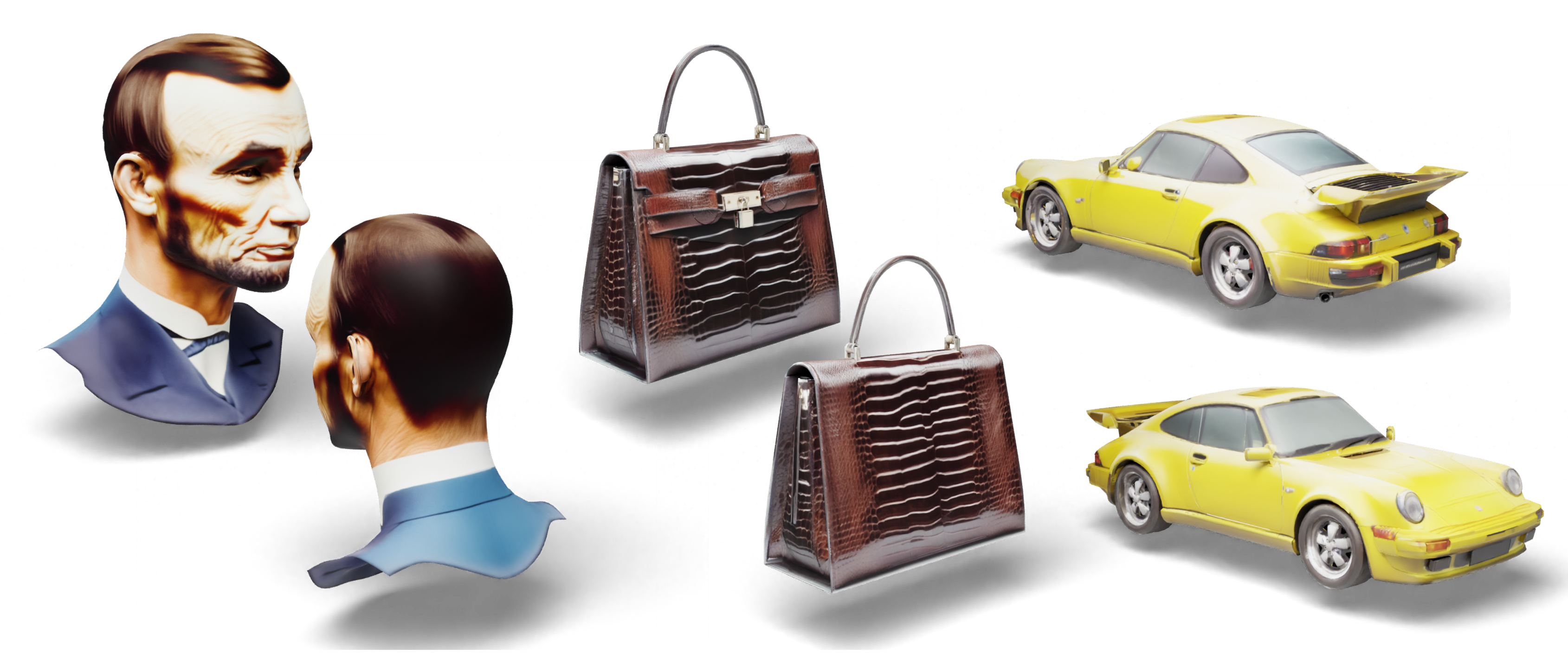}
    \caption{Ours}
    \end{subfigure}
    \vspace{-0.3cm}
    \caption{Comparison with visual results directly taken from TexFusion paper. Same prompts and similar meshes are used as input to our method. A similar lighting setup is used to render the images. Image credit: {\tt"Portrait photo of Abraham Lincoln, full color"},{\tt"Crocodile skin handbag"}, and {\tt"Beautiful yellow sports car"} by Tianshi Cao, Karsten Kreis, Sanja Fidler, Nicholas Sharp, and Kangxue Yin, 2023. Adapted from video frame of \textit {TexFusion: Synthesizing 3D Textures with Text-Guided Image Diffusion Models}. Accessed via \href{https://research.nvidia.com/labs/toronto-ai/texfusion/}{https://research.nvidia.com/labs/toronto-ai/texfusion/}}
    \label{fig:texfusion}
\end{figure*}

% \noindent\textbf{Comparison with TEXTure and Text2Tex.}
\subsubsection{Comparison with TEXTure and Text2Tex}
TEXTure \cite{richardson2023texture} and Text2Tex \cite{chen2023text2tex} are two progressive inpainting methods that performs texturing by iteratively warping painted texture to new views for inpainting. Despite that their inpainting reworks part of the existing textured region to reduce seams, their methods still suffer from over-fragmentation and obvious seams, especially at the back of the object, as shown in Fig. \ref{fig:illustrate}. On the other hand, by reusing self-attention to attend to other views, our methods significantly improves the visual quality and style consistency of the texture as viewed from different angles.

% \noindent\textbf{Comparison with Meshy.}
\subsubsection{Comparison with Meshy}
A commercial software named Meshy \cite{meshy} has shown better consistency in many test cases compared to TEXTure and Text2Tex. Their method produces highly contrasted content while tends to generate over-saturated colors and misinterpretation of prompts, with significant inconsistency or blank regions in some test cases in Fig.~\ref{fig:illustrate}. 

% \noindent\textbf{Comparison with Paint3D.}
\subsubsection{Comparison with Paint3D}
Paint3D \cite{zeng2024paint3d} is a data-driven approach that performs texture generation in the UV domain. While benefiting from 3D training data, the method may be vulnerable to complex geometry in which the UV layout  is  difficult to map, and results in obvious seam artifact (Fig.~\ref{fig:illustrate}). Besides, it also relies on a progressive inpainting method for texture initialization, which is prone to introduce errors into UV-space texture generation. 

% \noindent\textbf{Comparison with TexFusion.}
\subsubsection{Comparison with TexFusion}
Besides the visual results of simulated TexFusion in Fig.~\ref{fig:illustrate}, we also  side-by-side compare several test cases extracted from TexFusion paper (Fig~\ref{fig:texfusion}). Since we don't have access to their 3D objects, we can only use 3D objects that are as close as possible for this comparison. 
Comparing to TexFusion, we can observe that better view consistency and finer details in our results (also confirmed by Table~\ref{tab:userstudy}). We hypothesize that our method benefited from our true synchronized denoising (i.e., non auto-regressive), reinforced by self attention reuse which is not easily available in an auto-regressive pipeline. However, more in-depth  comparison and analysis can only be conducted when their code becomes available. 

\begin{table}[h]
    \centering
    \caption{Quantitative comparison and user study. }
    \vspace{-.1in}
    \resizebox{0.48\textwidth}{!}{
    \setlength\tabcolsep{3pt}
    \centering
    \begin{tabular}{c@{\hskip 0.05in} | c@{\hskip 0.15in} c@{\hskip 0.15in} c@{\hskip 0.05in} | c@{\hskip 0.15in} c@{\hskip 0.15in} c@{\hskip 0.15in} c@{\hskip 0.15in} c@{\hskip 0.05in}}
    
        \Xhline{2\arrayrulewidth}
         \makecell{Property} & \makecell{FID($\downarrow$)} &\makecell{CLIP\\Score($\uparrow$)} & \makecell{Consist.\\Score($\uparrow$)} & \makecell{Natural\\Color} & \makecell{More\\Details} & \makecell{Less\\Artifact} & \makecell{3D-Con-\\sistency} & \makecell{Align w.\\Text}\\ \hline
        \makecell{SD\\(reference)} & \makecell{-} & \makecell{26.99} &  \makecell{64.79} &
        \makecell{-} & \makecell{-} & \makecell{-} & \makecell{-} & \makecell{-} \tabularnewline[0.99em]
        % \hline
       % \vspace{.18in} \\
       \makecell{Latent-\\Paint} & \makecell{107.92} & \makecell{24.30}& \makecell{68.48} & 
       \makecell{3.77\%} & \makecell{2.63\%} & \makecell{0.21\%} & \makecell{1.49\%} & \makecell{3.86\%} \tabularnewline[0.99em]
       % \hline
       % \vspace{-.08in}\\
       \makecell{Meshy} & \makecell{82.24} & \makecell{26.09}& \makecell{68.92} &
       \makecell{10.62\%} & \makecell{\textbf{22.56\%}} & \makecell{6.91\%} & \makecell{6.94\%} & \makecell{6.45\%} \tabularnewline[0.99em]
       % \hline
       % \vspace{-.08in}\\
       \makecell{Text2Tex} &  \makecell{71.20} & \makecell{25.67} & \makecell{68.00} &
       \makecell{10.22\%} & \makecell{9.33\%} & \makecell{9.22\%} & \makecell{16.15\%} & \makecell{12.64\%} \tabularnewline[0.99em]
       % \hline
       % \vspace{-.08in}\\
       \makecell{TEXTure} & \makecell{74.36} & \makecell{26.33}  & \makecell{68.37} &
       \makecell{9.35\%} & \makecell{21.42\%} & \makecell{11.52\%} & \makecell{8.26\%} & \makecell{11.65\%} \tabularnewline[0.99em]
       % \hline
       % \vspace{-.08in}\\
       \makecell{Paint3D} & \makecell{72.01} & \makecell{24.91}  & \makecell{68.20} & 
       \makecell{8.86\%} & \makecell{9.84\%} & \makecell{8.46\%} & \makecell{12.84\%} & \makecell{9.23\%} \tabularnewline[0.99em]
       % \hline
       % \vspace{-.08in}\\
       \makecell{TexFusion\\(Simulated)} & \makecell{62.82} & \makecell{26.15}  & \makecell{68.43} & 
       \makecell{18.08\%} & \makecell{14.18\%} & \makecell{20.84\%} & \makecell{17.19\%} & \makecell{24.67\%} \tabularnewline[0.99em]
       % \hline
       % \vspace{-.08in}\\
       \makecell{Ours} & \makecell{\textbf{50.26}} & \makecell{\textbf{27.06}} & \makecell{\textbf{68.97}} &
       \textbf{\makecell{39.11\%}} & \makecell{20.03\%} & \textbf{\makecell{42.84\%}} & \textbf{\makecell{37.12\%}} & \textbf{\makecell{31.42\%}} 
       \tabularnewline[0.29em]
       \Xhline{2\arrayrulewidth}
    \end{tabular}
    }
    \label{tab:userstudy}
    % \vspace{-0.5cm}
    
\end{table}

% \noindent\textbf{Quantitative Evaluation.}
\subsubsection{Quantitative Evaluation}
% In view of the availability of source codes, we can only select TEXTure~\cite{richardson2023texture}, Text2Tex~\cite{chen2023text2tex} and Meshy~\cite{meshy} for quantitative comparison.
Here we compute the Fréchet Inception Distance (FID), CLIP score and 3D consistency score in our evaluation. FID measures the difference between the output distribution of original Stable Diffusion with ControlNet, and the rendering  of textured objects with textures generated by each method, similar to the metric used in \cite{cao2023texfusion}. We make use of CLIP model \cite{sanghi2022clip} to measure the similarity between the given texture prompt and the appearance of the textured model. Inspired by \cite{esser2023structure}, we also calculate a 3D-consistency score of generated textures by computing the average similarities between all possible pairs of rendered view, for each textured object. The similarity between a pair of images is defined as the cosine distance between the embeddings produced by the CLIP image encoder. Here we include the SD method in our comparison which uses images directly generated by Stable Diffusion and ControlNet (with background removed) instead of renders of textured objects, serving as a performance reference for single-view generation using the same input. 
% Ideally, such difference should be as small as possible, because textures are derived from the pretrained T2I model. 
The quantitative evaluation in Table~\ref{tab:userstudy} shows that our method achieves the best FID score, CLIP score and consistency score among the baselines. The proposed method shows noticeable improvement on image quality compared to baseline methods, due to successful suppression of fragment artifacts common in multi-view texturing.

\subsection{User study} 
We conducted a user study to evaluate the human preference of our method compared to other approaches (Table~\ref{tab:userstudy}). The user study includes 10 different mesh-text pairs from various categories, each textured with the above methods and rendered from two opposite directions. The users are required to rank the methods on the following properties: better naturalness of textured color, more of details, less artifact, better 3D-consistency, and more aligned with text. The performance of the methods is summarized using the percentage of being the most preferred choice (the higher the better).
The user study demonstrated superior performance of our method on several aspects, especially in terms of natural color, less artifact and better consistency, where the proposed method leads by a large margin. Nevertheless, Meshy and TEXTure have been ranked to be slightly more detailed. We hypothesize that certain details could be hard to develop with strict consistency constraint; on the other hand, users may sometimes regard the over-fragmentation and seams as details, as Meshy and TEXTure results also come with more artifacts (column ``Less Artifact'' in Table~\ref{tab:userstudy}) and lower consistency (column ``3D-Consistency'').

\begin{figure}[!h]
    % \centering
    \begin{subfigure}{0.9\linewidth}
        \includegraphics[width=0.48\linewidth]{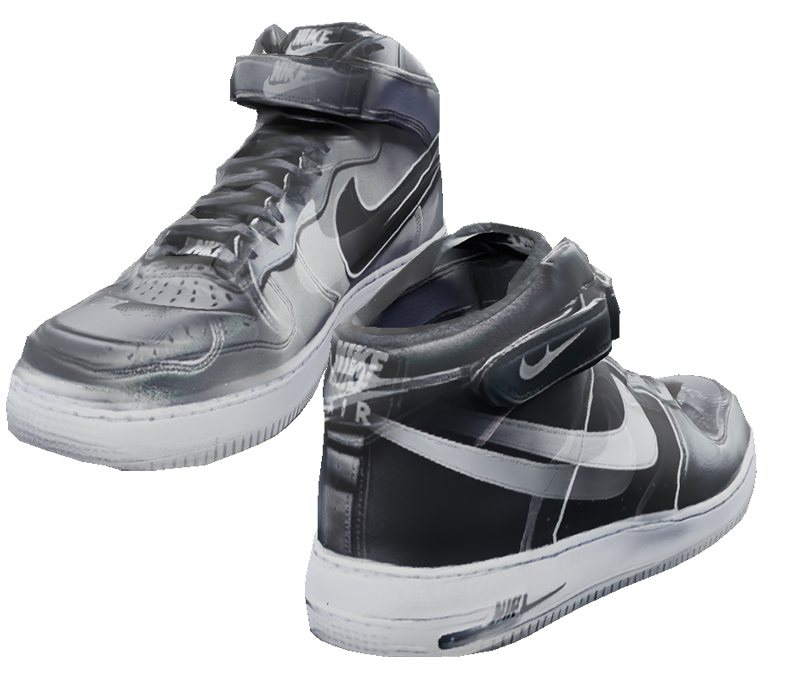}
        \includegraphics[width=0.48\linewidth]{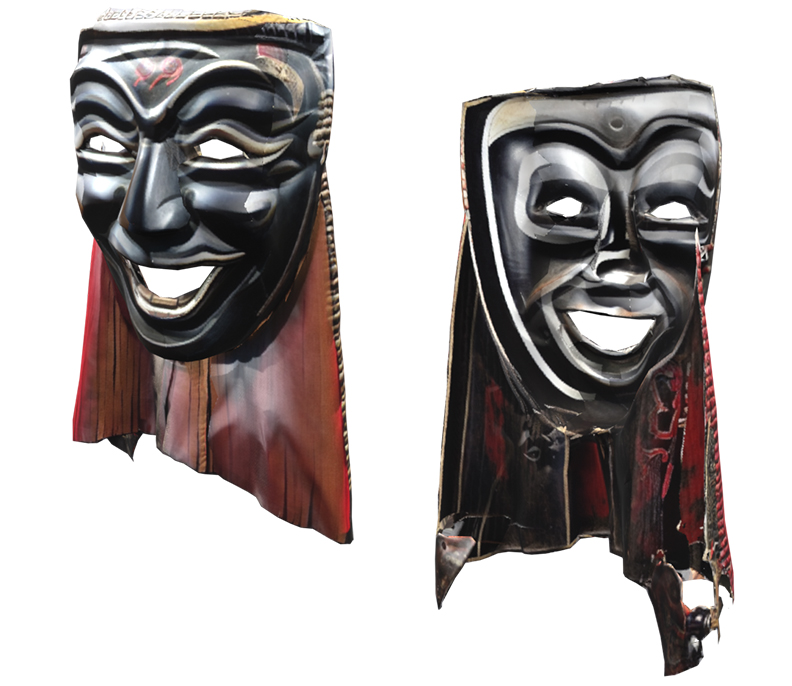}
        \caption{w/o MVD}
        % \label{fig:attnA}
    \end{subfigure}
    \begin{subfigure}{0.9\linewidth}
        \includegraphics[width=0.48\linewidth]{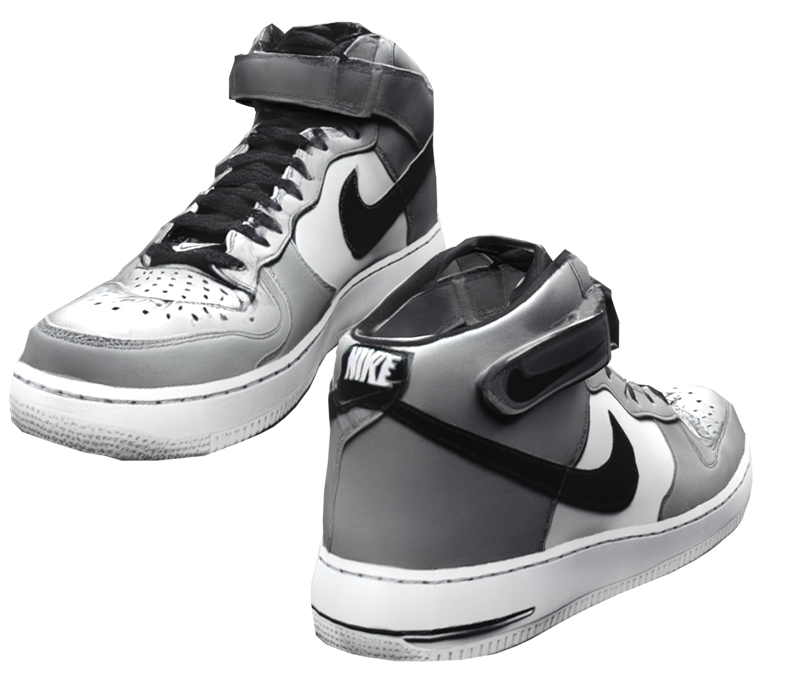}
        \includegraphics[width=0.48\linewidth]{gallery/5}
        \caption{w MVD}
        % \label{fig:attnB}
    \end{subfigure}
    % \vspace{-.05in}
    \caption{Ablation study of Multi-View Diffusion module (MVD). Ghosting artifact appears when multiple views without consensus are projected into a texture.}
    \label{fig:ablation-mvd}
\end{figure}

\begin{figure}[!h]
    % \centering
    \begin{subfigure}[t]{0.9\linewidth}
        \includegraphics[width=0.49\linewidth]{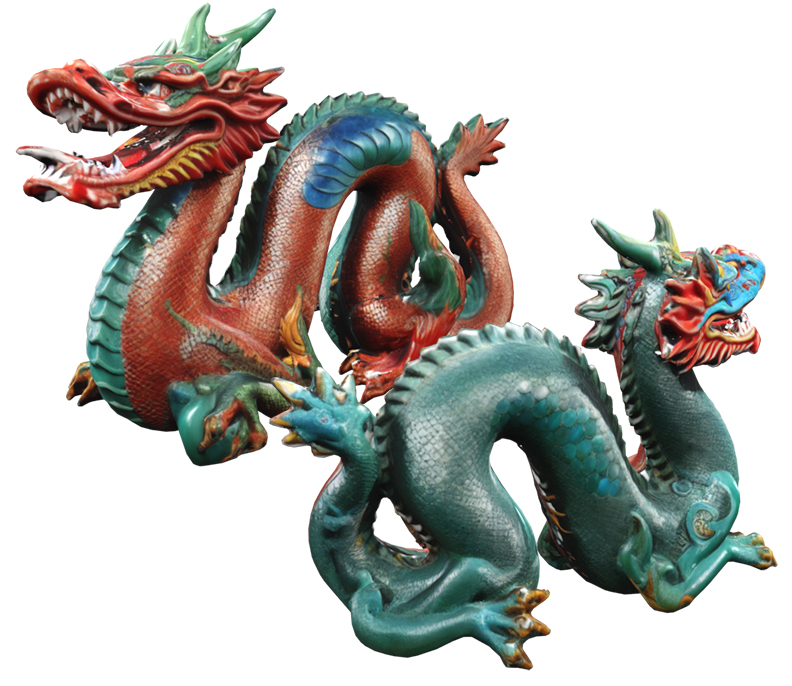}
        \includegraphics[width=0.47\linewidth]{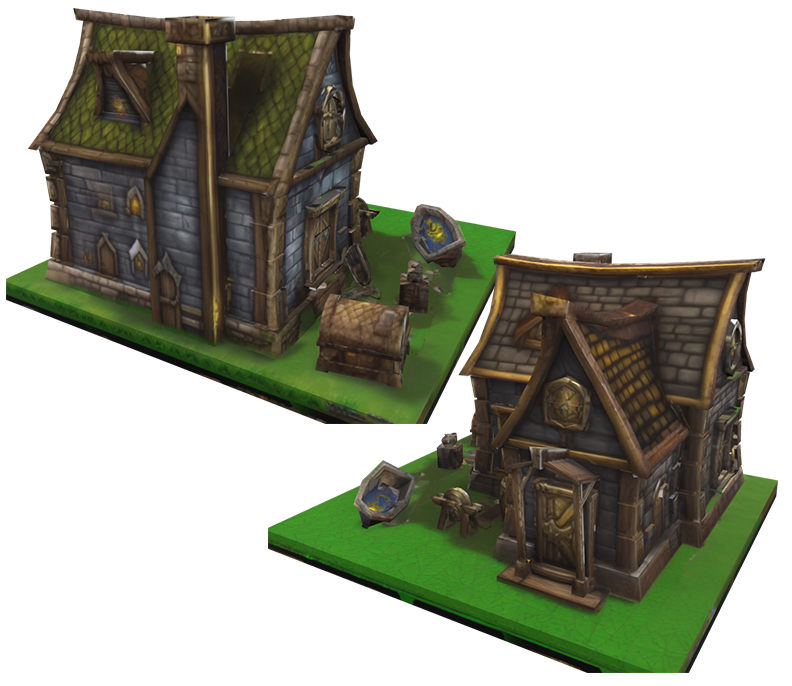}
        \caption{w/o SAR}
        % \label{fig:attnA}
    \end{subfigure}
    \begin{subfigure}[t]{0.9\linewidth}
        \includegraphics[width=0.48\linewidth]{gallery/2}
        \includegraphics[width=0.48\linewidth]{gallery/20}
        \caption{w SAR}
        % \label{fig:attnB}
    \end{subfigure}
    % \vspace{-.05in}
    \caption{Ablation study of Self-attention reuse (SAR). Distant views (e.g. front and back) tends to be more coherent in color with self-attention reuse enabled.}
    % \vspace{-.1in}
    \label{fig:ablation-sar}
\end{figure}

\subsection{Ablation Study}

% \noindent\textbf{Multi-View Diffusion Module.} 
\subsubsection{Multi-View Diffusion Module} 
Fig.~\ref{fig:ablation-mvd} compares the results with and without our Multi-View Diffusion module. In the experiment without the MVD module, we initialize 3D consistent initial latent noise and denoise each view individually using default Stable Diffusion and ControlNet pipeline.  
The final texture of the case without MVD exhibits severe ghosting artifact, because the generated appearances from different views are significantly different. Without MVD, there is no consensus in content among different views.  This proves that MVD module is essential for alignment and localization of surface details on the textured object.

% \noindent\textbf{Self-attention Reuse.}
\subsubsection{Self-attention Reuse}
We also examined the effect of self-attention reuse. When MVD is performed without attention reuse, consensus on the object appearance sometimes could be difficult to reach in the early diffusion process, leaving conflicting appearance in the end result. In certain extreme cases, this also leads to redundant or incomplete content on the object surface. For instance, the two sides of the dragon statue and the roof of the house are colored differently from different view directions.  (Fig. \ref{fig:ablation-sar}). Therefore self-attention reuse can compensate the MVD in reaching consensus for certain cases, especially when the overlap among views is insufficient. 

\begin{figure}[!hb]
    % \centering

    \begin{subfigure}{0.9\linewidth} 
        \includegraphics[width=1\linewidth]{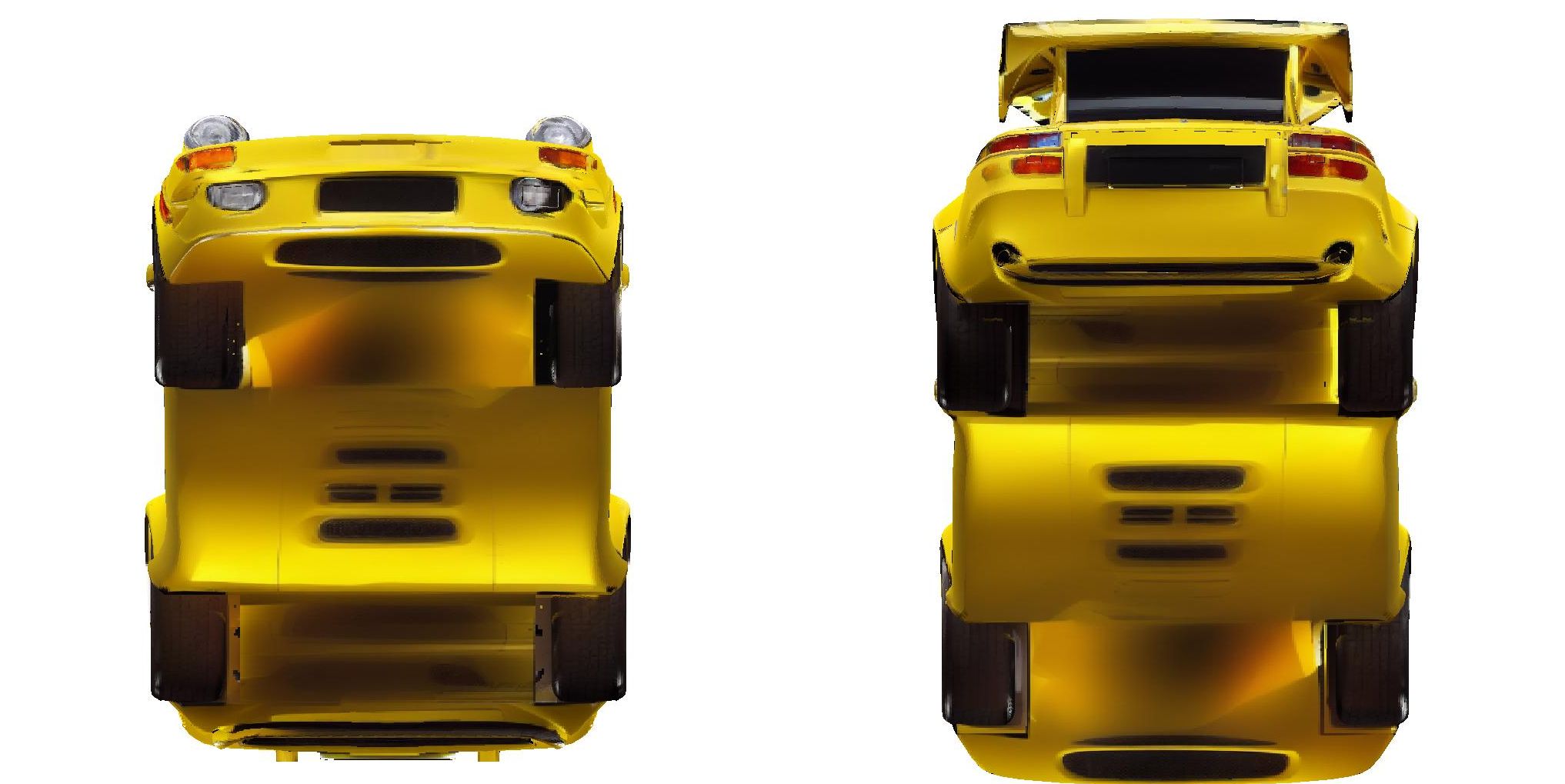}
        \caption{improper bottom view}
    \end{subfigure}
    \begin{subfigure}{0.9\linewidth}
        \includegraphics[width=1\linewidth]{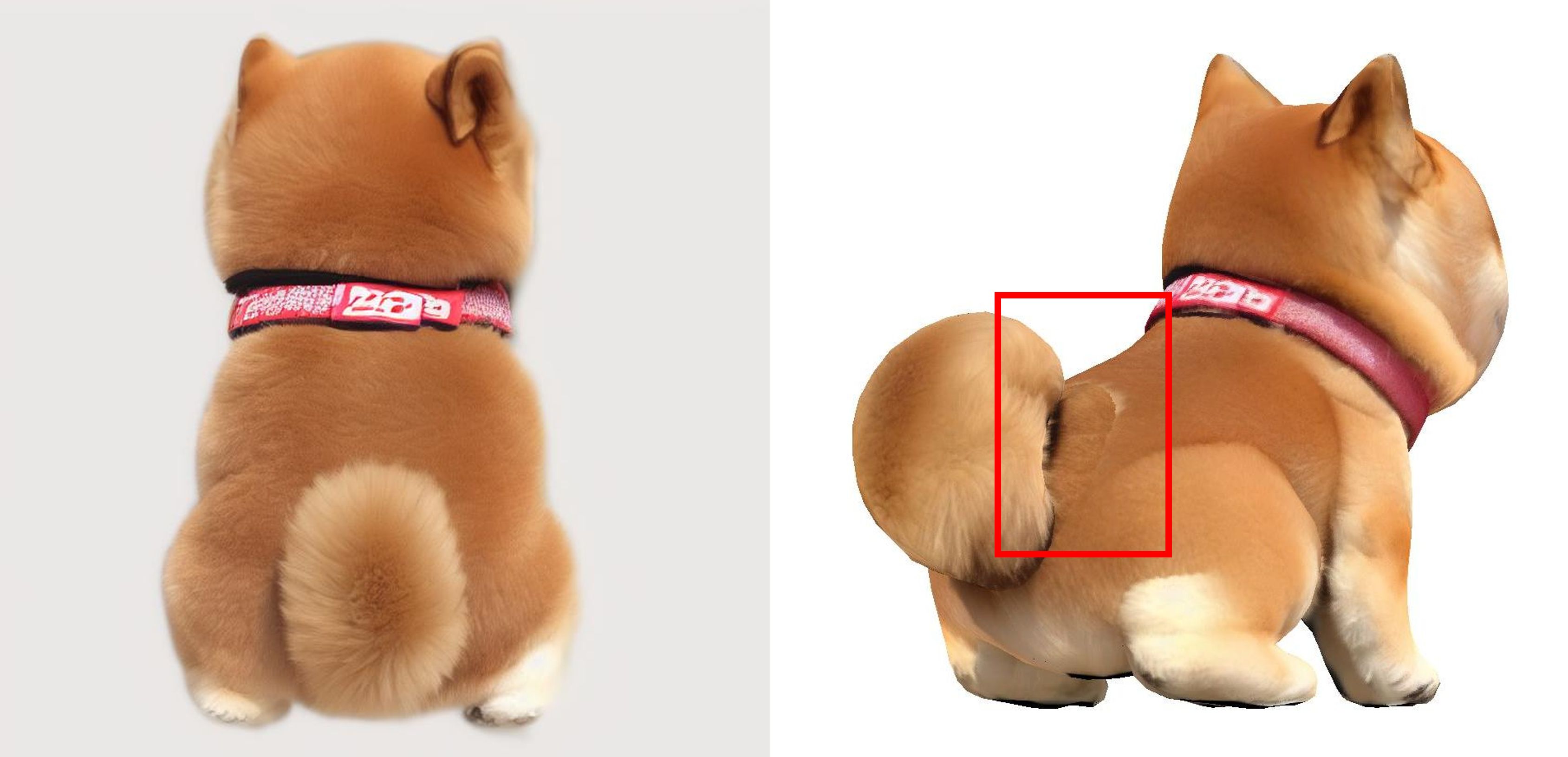}
        \caption{incorrect projection}
    \end{subfigure}
    % \vspace{-.05in}
    % \vspace{-.05in}
    \caption{Limitations.
    (a) A radiator grill which is supposed to be generated in the front view is generated on the lower surface when two additional cameras below the car mesh are place to texture the bottom.
    (b) The furry tail of the dog does not have a clear silhouette, leading to an incorrect projection of the texture onto the dog’s back.}
    % \vspace{-.05in}
    \label{fig:limitation}
\end{figure}

\subsection{Limitations}
Although our method performs well in generating consistent and high-quality textures, we observe the following limitations. Firstly, the textures generated using our method may embed (bake) the lighting effect  (e.g., highlight on batman in Fig.~\ref{fig:illustrate}). Consequently, directly rendering with our texture under a new lighting condition may not be correct. Secondly, the pre-trained model has a bias towards generating common views (e.g., frontal views) of the object specified in the text prompt, making it unlikely to generate proper bottom view of the object. For instance, it may inadvertently depict a radiator grill that should only appear in the frontal view when attempting to render the bottom view of a car (see Fig.~\ref{fig:limitation}(a)). We believe this limitation is inherited from the pre-trained model, and is hard to fully circumvent using techniques like directional prompt and depth-guided generation adopted in our method.
Lastly, our method does not guarantee perfect boundaries at discontinuous depth boundaries in the denoised views (e.g., boundaries of self-occluding geometry or between foreground and background), this may lead to problems during RGB texture extraction and cause color bleeding to the unconnected regions. A potential solution is to introduce optimization based extraction method with perceptual losses or  mask out the unreliable boundaries before projecting to the texture.

\section{Conclusion}
\label{sec:conclusion}
In this paper, we present Synchronized Multi-View Diffusion, which synthesizes consistent and high-quality textures in a zero-shot setting, by utilizing the pre-trained T2I latent diffusion models. By sharing information among views in each intermediate denoising step through overlapping regions in texture space, our method solves the over-fragmentation and seam problems of existing progressive inpainting methods. Furthermore, our method utilizes the pretrained self-attention layers as an additional assurance for consistency, further eliminating inconsistent results. Both qualitative and quantitative results demonstrate the effectiveness of the synchronized combination of intermediate diffusion results in achieving a good convergence to a 3D-consistent texturing result without significant artifacts. 

\begin{figure}[h!]

    \includegraphics[width=0.45\linewidth]{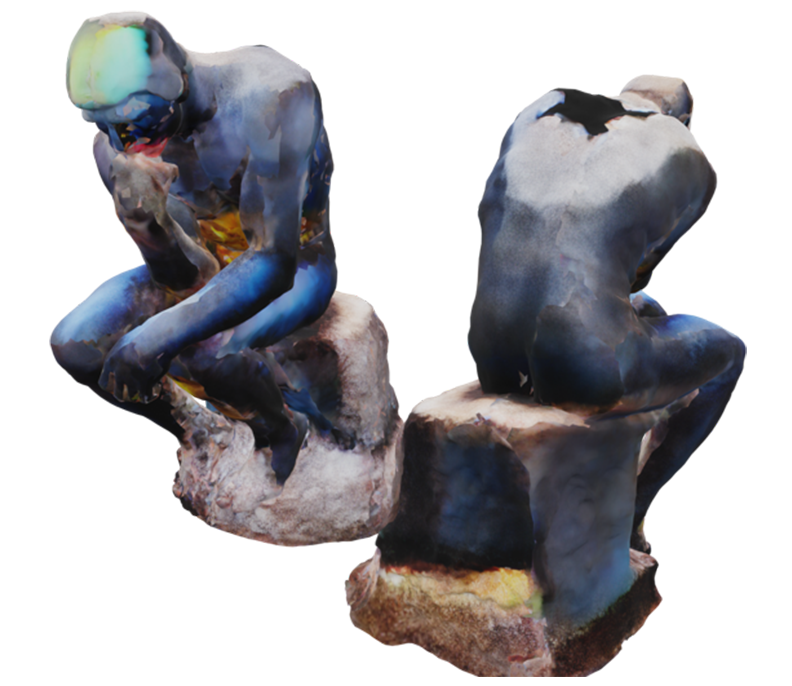}
    \includegraphics[width=0.45\linewidth]{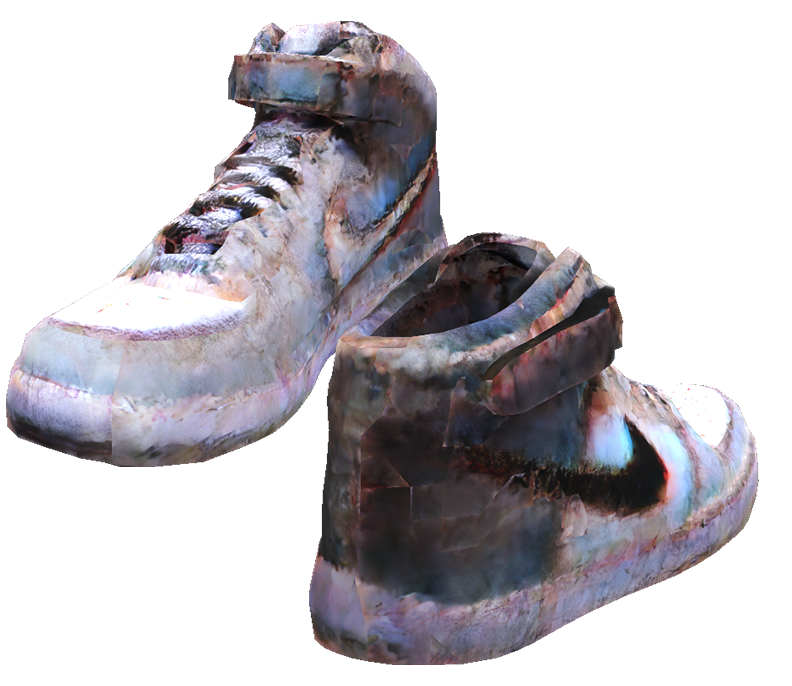}
    \includegraphics[width=0.45\linewidth]{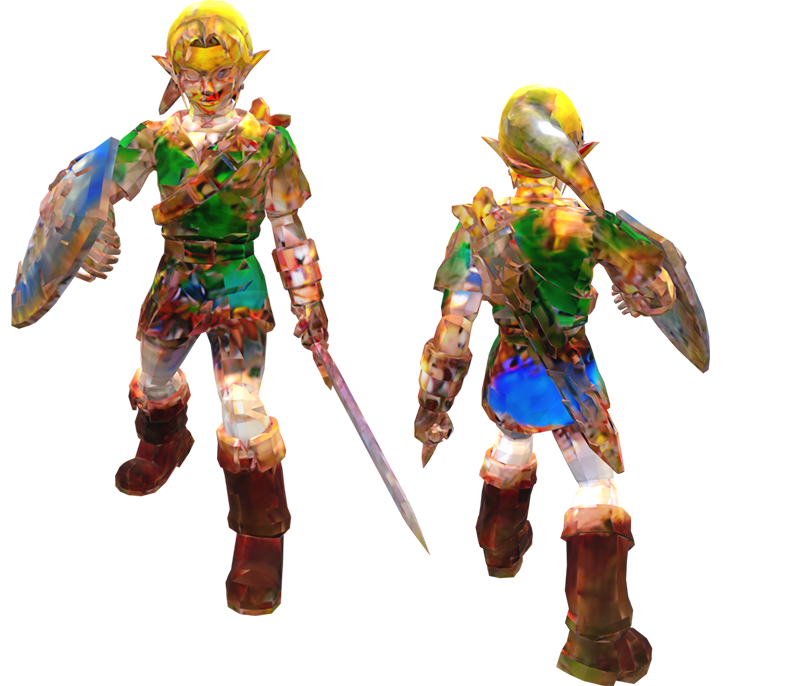}
    \includegraphics[width=0.45\linewidth]{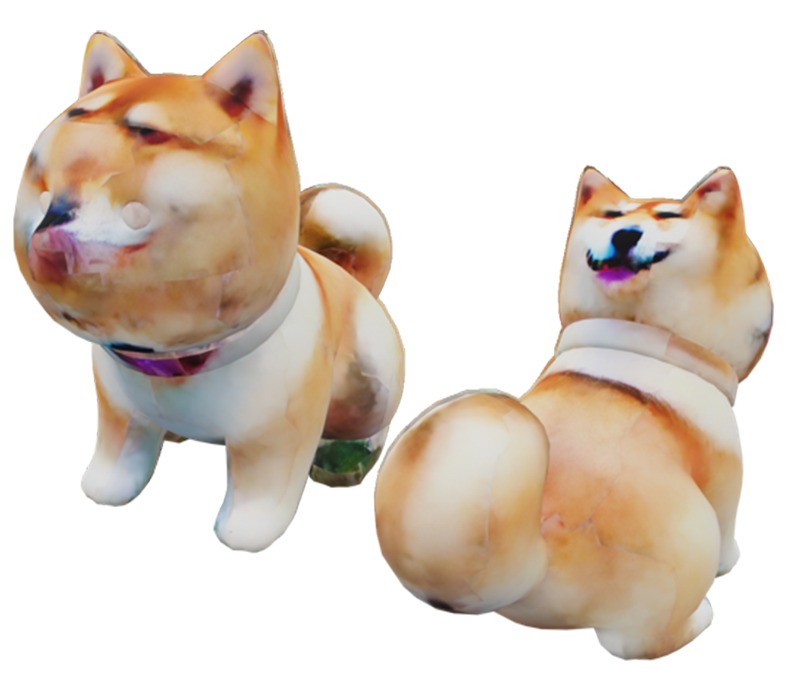}
    \includegraphics[width=0.45\linewidth]{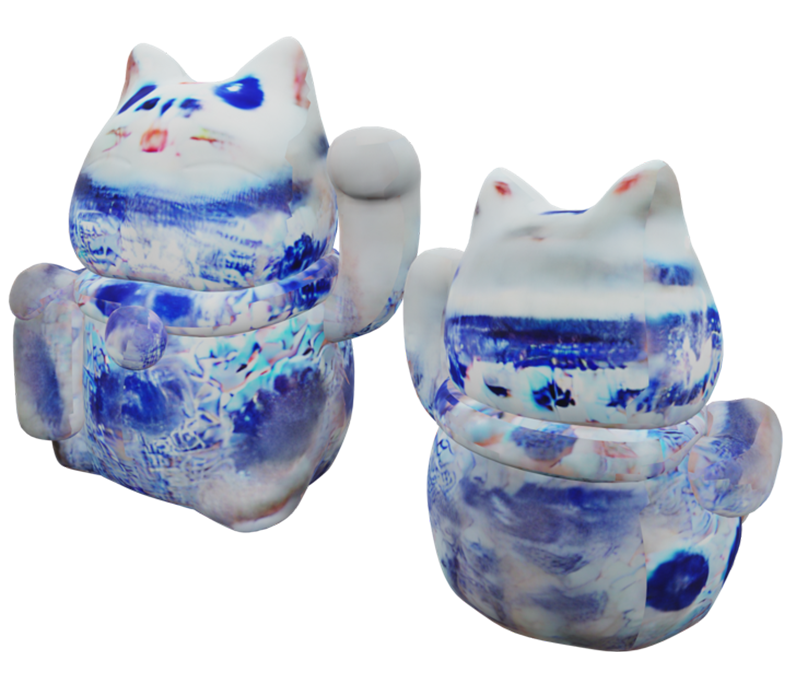}

    \caption{Object textured by LatentPaint. Refer to Fig.~\ref{fig:illustrate} for source mesh and text prompt input.}
    \label{fig:illustrate_latentpaint}
    \vspace{-0.2cm}
\end{figure}

\begin{acks}
The 3D models used to present the results in this paper are primarily sourced from the Objaverse dataset \cite{deitke2023objaverse}. Remaining 3D models are obtained from \href{https://sketchfab.com}{Sketchfab} \cite{3dmodelbearbrick, 3dmodelgloves}.
\end{acks}

\bibliographystyle{ACM-Reference-Format} %plainnat
\bibliography{main}


\begin{thebibliography}{54}

%%% the Web address.

\ifx \showCODEN    \undefined \def \showCODEN     #1{\unskip}     \fi
\ifx \showDOI      \undefined \def \showDOI       #1{#1}\fi
\ifx \showISBNx    \undefined \def \showISBNx     #1{\unskip}     \fi
\ifx \showISBNxiii \undefined \def \showISBNxiii  #1{\unskip}     \fi
\ifx \showISSN     \undefined \def \showISSN      #1{\unskip}     \fi
\ifx \showLCCN     \undefined \def \showLCCN      #1{\unskip}     \fi
\ifx \shownote     \undefined \def \shownote      #1{#1}          \fi
\ifx \showarticletitle \undefined \def \showarticletitle #1{#1}   \fi
\ifx \showURL      \undefined \def \showURL       {\relax}        \fi
% invisible to TeX
\providecommand\bibfield[2]{#2}
\providecommand\bibinfo[2]{#2}
\providecommand\natexlab[1]{#1}
\providecommand\showeprint[2][]{arXiv:#2}

\bibitem[alixor22(2023)]%
        {3dmodelgloves}
\bibfield{author}{\bibinfo{person}{alixor22}.} \bibinfo{year}{2023}\natexlab{}.
\newblock \bibinfo{booktitle}{\emph{Military Glove 3D Model}}.
\newblock
\urldef\tempurl%
\url{https://sketchfab.com/3d-models/military-glove-d1a4dee1c8594dfea92f8afe35692c13}
\showURL{%
\tempurl}
\newblock
\shownote{Accessed 15-09-2024}.


\bibitem[Ashikhmin(2001)]%
        {ashikhmin2001synthesizing}
\bibfield{author}{\bibinfo{person}{Michael Ashikhmin}.} \bibinfo{year}{2001}\natexlab{}.
\newblock \showarticletitle{Synthesizing natural textures}. In \bibinfo{booktitle}{\emph{Proceedings of the 2001 symposium on Interactive 3D graphics}}. \bibinfo{pages}{217--226}.
\newblock


\bibitem[Aurenhammer(1991)]%
        {aurenhammer1991voronoi}
\bibfield{author}{\bibinfo{person}{Franz Aurenhammer}.} \bibinfo{year}{1991}\natexlab{}.
\newblock \showarticletitle{Voronoi diagrams—a survey of a fundamental geometric data structure}.
\newblock \bibinfo{journal}{\emph{ACM Computing Surveys (CSUR)}} \bibinfo{volume}{23}, \bibinfo{number}{3} (\bibinfo{year}{1991}), \bibinfo{pages}{345--405}.
\newblock


\bibitem[Cao et~al\mbox{.}(2023)]%
        {cao2023texfusion}
\bibfield{author}{\bibinfo{person}{Tianshi Cao}, \bibinfo{person}{Karsten Kreis}, \bibinfo{person}{Sanja Fidler}, \bibinfo{person}{Nicholas Sharp}, {and} \bibinfo{person}{KangXue Yin}.} \bibinfo{year}{2023}\natexlab{}.
\newblock \showarticletitle{TexFusion: Synthesizing 3D Textures with Text-Guided Image Diffusion Models}. In \bibinfo{booktitle}{\emph{Proceedings of the IEEE/CVF International Conference on Computer Vision (ICCV)}}.
\newblock


\bibitem[Chen et~al\mbox{.}(2023b)]%
        {chen2023text2tex}
\bibfield{author}{\bibinfo{person}{Dave~Zhenyu Chen}, \bibinfo{person}{Yawar Siddiqui}, \bibinfo{person}{Hsin-Ying Lee}, \bibinfo{person}{Sergey Tulyakov}, {and} \bibinfo{person}{Matthias Nie{\ss}ner}.} \bibinfo{year}{2023}\natexlab{b}.
\newblock \showarticletitle{Text2tex: Text-driven texture synthesis via diffusion models}.
\newblock \bibinfo{journal}{\emph{arXiv preprint arXiv:2303.11396}} (\bibinfo{year}{2023}).
\newblock


\bibitem[Chen et~al\mbox{.}(2023a)]%
        {chen2023fantasia3d}
\bibfield{author}{\bibinfo{person}{Rui Chen}, \bibinfo{person}{Yongwei Chen}, \bibinfo{person}{Ningxin Jiao}, {and} \bibinfo{person}{Kui Jia}.} \bibinfo{year}{2023}\natexlab{a}.
\newblock \bibinfo{title}{Fantasia3D: Disentangling Geometry and Appearance for High-quality Text-to-3D Content Creation}.
\newblock
\newblock
\showeprint[arxiv]{2303.13873}~[cs.CV]


\bibitem[Deitke et~al\mbox{.}(2023)]%
        {deitke2023objaverse}
\bibfield{author}{\bibinfo{person}{Matt Deitke}, \bibinfo{person}{Dustin Schwenk}, \bibinfo{person}{Jordi Salvador}, \bibinfo{person}{Luca Weihs}, \bibinfo{person}{Oscar Michel}, \bibinfo{person}{Eli VanderBilt}, \bibinfo{person}{Ludwig Schmidt}, \bibinfo{person}{Kiana Ehsani}, \bibinfo{person}{Aniruddha Kembhavi}, {and} \bibinfo{person}{Ali Farhadi}.} \bibinfo{year}{2023}\natexlab{}.
\newblock \showarticletitle{Objaverse: A universe of annotated 3d objects}. In \bibinfo{booktitle}{\emph{Proceedings of the IEEE/CVF Conference on Computer Vision and Pattern Recognition}}. \bibinfo{pages}{13142--13153}.
\newblock


\bibitem[Ebert(2003)]%
        {ebert2003texturing}
\bibfield{author}{\bibinfo{person}{David~S Ebert}.} \bibinfo{year}{2003}\natexlab{}.
\newblock \bibinfo{booktitle}{\emph{Texturing \& modeling: a procedural approach}}.
\newblock \bibinfo{publisher}{Morgan Kaufmann}.
\newblock


\bibitem[Ebert et~al\mbox{.}(2003)]%
        {ebert2003}
\bibfield{author}{\bibinfo{person}{David~S. Ebert}, \bibinfo{person}{F.~Kenton Musgrave}, \bibinfo{person}{Darwyn Peachey}, \bibinfo{person}{Ken Perlin}, {and} \bibinfo{person}{Steve Worley}.} \bibinfo{year}{2003}\natexlab{}.
\newblock \bibinfo{booktitle}{\emph{Texturing and Modeling: A Procedural Approach, Third Edition}}.
\newblock \bibinfo{publisher}{Morgan Kaufmann Publishers}.
\newblock


\bibitem[Efros and Leung(1999)]%
        {efros1999texture}
\bibfield{author}{\bibinfo{person}{Alexei~A Efros} {and} \bibinfo{person}{Thomas~K Leung}.} \bibinfo{year}{1999}\natexlab{}.
\newblock \showarticletitle{Texture synthesis by non-parametric sampling}. In \bibinfo{booktitle}{\emph{Proceedings of the seventh IEEE international conference on computer vision}}, Vol.~\bibinfo{volume}{2}. IEEE, \bibinfo{pages}{1033--1038}.
\newblock


\bibitem[elisaghbini(2023)]%
        {3dmodelbearbrick}
\bibfield{author}{\bibinfo{person}{elisaghbini}.} \bibinfo{year}{2023}\natexlab{}.
\newblock \bibinfo{booktitle}{\emph{Bearbrick 3D Model}}.
\newblock
\urldef\tempurl%
\url{https://sketchfab.com/3d-models/bearbrick-78446f8e13364929a32000d5108ef17a}
\showURL{%
\tempurl}
\newblock
\shownote{Accessed 15-09-2024}.


\bibitem[Esser et~al\mbox{.}(2023)]%
        {esser2023structure}
\bibfield{author}{\bibinfo{person}{Patrick Esser}, \bibinfo{person}{Johnathan Chiu}, \bibinfo{person}{Parmida Atighehchian}, \bibinfo{person}{Jonathan Granskog}, {and} \bibinfo{person}{Anastasis Germanidis}.} \bibinfo{year}{2023}\natexlab{}.
\newblock \showarticletitle{Structure and content-guided video synthesis with diffusion models}. In \bibinfo{booktitle}{\emph{Proceedings of the IEEE/CVF International Conference on Computer Vision}}. \bibinfo{pages}{7346--7356}.
\newblock


\bibitem[Goodfellow et~al\mbox{.}(2014)]%
        {goodfellow2014generative}
\bibfield{author}{\bibinfo{person}{Ian Goodfellow}, \bibinfo{person}{Jean Pouget-Abadie}, \bibinfo{person}{Mehdi Mirza}, \bibinfo{person}{Bing Xu}, \bibinfo{person}{David Warde-Farley}, \bibinfo{person}{Sherjil Ozair}, \bibinfo{person}{Aaron Courville}, {and} \bibinfo{person}{Yoshua Bengio}.} \bibinfo{year}{2014}\natexlab{}.
\newblock \showarticletitle{Generative adversarial nets}.
\newblock \bibinfo{journal}{\emph{Advances in neural information processing systems}}  \bibinfo{volume}{27} (\bibinfo{year}{2014}).
\newblock


\bibitem[Gupta et~al\mbox{.}(2023)]%
        {gupta20233dgen}
\bibfield{author}{\bibinfo{person}{Anchit Gupta}, \bibinfo{person}{Wenhan Xiong}, \bibinfo{person}{Yixin Nie}, \bibinfo{person}{Ian Jones}, {and} \bibinfo{person}{Barlas O{\u{g}}uz}.} \bibinfo{year}{2023}\natexlab{}.
\newblock \showarticletitle{3dgen: Triplane latent diffusion for textured mesh generation}.
\newblock \bibinfo{journal}{\emph{arXiv preprint arXiv:2303.05371}} (\bibinfo{year}{2023}).
\newblock


\bibitem[Ho et~al\mbox{.}(2020)]%
        {ho2020denoising}
\bibfield{author}{\bibinfo{person}{Jonathan Ho}, \bibinfo{person}{Ajay Jain}, {and} \bibinfo{person}{Pieter Abbeel}.} \bibinfo{year}{2020}\natexlab{}.
\newblock \showarticletitle{Denoising diffusion probabilistic models}.
\newblock \bibinfo{journal}{\emph{Advances in neural information processing systems}}  \bibinfo{volume}{33} (\bibinfo{year}{2020}), \bibinfo{pages}{6840--6851}.
\newblock


\bibitem[Jain et~al\mbox{.}(2022)]%
        {jain2022zero}
\bibfield{author}{\bibinfo{person}{Ajay Jain}, \bibinfo{person}{Ben Mildenhall}, \bibinfo{person}{Jonathan~T Barron}, \bibinfo{person}{Pieter Abbeel}, {and} \bibinfo{person}{Ben Poole}.} \bibinfo{year}{2022}\natexlab{}.
\newblock \showarticletitle{Zero-shot text-guided object generation with dream fields}. In \bibinfo{booktitle}{\emph{Proceedings of the IEEE/CVF Conference on Computer Vision and Pattern Recognition}}. \bibinfo{pages}{867--876}.
\newblock


\bibitem[jpcy(2016)]%
        {xatlas2016}
\bibfield{author}{\bibinfo{person}{jpcy}.} \bibinfo{year}{2016}\natexlab{}.
\newblock \bibinfo{title}{xatlas}.
\newblock \bibinfo{howpublished}{\url{https://github.com/jpcy/xatlas}}.
\newblock


\bibitem[Karras et~al\mbox{.}(2019)]%
        {karras2019style}
\bibfield{author}{\bibinfo{person}{Tero Karras}, \bibinfo{person}{Samuli Laine}, {and} \bibinfo{person}{Timo Aila}.} \bibinfo{year}{2019}\natexlab{}.
\newblock \showarticletitle{A style-based generator architecture for generative adversarial networks}. In \bibinfo{booktitle}{\emph{Proceedings of the IEEE/CVF conference on computer vision and pattern recognition}}. \bibinfo{pages}{4401--4410}.
\newblock


\bibitem[Khalid et~al\mbox{.}(2022)]%
        {Khalid2022CLIPMeshGT}
\bibfield{author}{\bibinfo{person}{Nasir~Mohammad Khalid}, \bibinfo{person}{Tianhao Xie}, \bibinfo{person}{Eugene Belilovsky}, {and} \bibinfo{person}{Tiberiu Popa}.} \bibinfo{year}{2022}\natexlab{}.
\newblock \showarticletitle{CLIP-Mesh: Generating textured meshes from text using pretrained image-text models}.
\newblock \bibinfo{journal}{\emph{SIGGRAPH Asia 2022 Conference Papers}} (\bibinfo{year}{2022}).
\newblock
\urldef\tempurl%
\url{https://api.semanticscholar.org/CorpusID:252089441}
\showURL{%
\tempurl}


\bibitem[Kingma and Welling(2013)]%
        {kingma2013auto}
\bibfield{author}{\bibinfo{person}{Diederik~P Kingma} {and} \bibinfo{person}{Max Welling}.} \bibinfo{year}{2013}\natexlab{}.
\newblock \showarticletitle{Auto-encoding variational bayes}.
\newblock \bibinfo{journal}{\emph{arXiv preprint arXiv:1312.6114}} (\bibinfo{year}{2013}).
\newblock


\bibitem[Kopf et~al\mbox{.}(2007)]%
        {kopf2007solid}
\bibfield{author}{\bibinfo{person}{Johannes Kopf}, \bibinfo{person}{Chi-Wing Fu}, \bibinfo{person}{Daniel Cohen-Or}, \bibinfo{person}{Oliver Deussen}, \bibinfo{person}{Dani Lischinski}, {and} \bibinfo{person}{Tien-Tsin Wong}.} \bibinfo{year}{2007}\natexlab{}.
\newblock \showarticletitle{Solid Texture Synthesis from 2D Exemplars}.
\newblock \bibinfo{journal}{\emph{ACM Trans. Graph.}} \bibinfo{volume}{26}, \bibinfo{number}{3} (\bibinfo{date}{jul} \bibinfo{year}{2007}), \bibinfo{pages}{2–es}.
\newblock


\bibitem[Kwatra et~al\mbox{.}(2005)]%
        {kwatra2005texture}
\bibfield{author}{\bibinfo{person}{Vivek Kwatra}, \bibinfo{person}{Irfan Essa}, \bibinfo{person}{Aaron Bobick}, {and} \bibinfo{person}{Nipun Kwatra}.} \bibinfo{year}{2005}\natexlab{}.
\newblock \showarticletitle{Texture optimization for example-based synthesis}.
\newblock In \bibinfo{booktitle}{\emph{ACM SIGGRAPH 2005 Papers}}. \bibinfo{pages}{795--802}.
\newblock


\bibitem[Li et~al\mbox{.}(2022)]%
        {Li2022DiffusionSDFTV}
\bibfield{author}{\bibinfo{person}{Muheng Li}, \bibinfo{person}{Yueqi Duan}, \bibinfo{person}{Jie Zhou}, {and} \bibinfo{person}{Jiwen Lu}.} \bibinfo{year}{2022}\natexlab{}.
\newblock \showarticletitle{Diffusion-SDF: Text-to-Shape via Voxelized Diffusion}.
\newblock \bibinfo{journal}{\emph{2023 IEEE/CVF Conference on Computer Vision and Pattern Recognition (CVPR)}} (\bibinfo{year}{2022}), \bibinfo{pages}{12642--12651}.
\newblock
\urldef\tempurl%
\url{https://api.semanticscholar.org/CorpusID:254366593}
\showURL{%
\tempurl}


\bibitem[Lin et~al\mbox{.}(2023)]%
        {lin2023magic3d}
\bibfield{author}{\bibinfo{person}{Chen-Hsuan Lin}, \bibinfo{person}{Jun Gao}, \bibinfo{person}{Luming Tang}, \bibinfo{person}{Towaki Takikawa}, \bibinfo{person}{Xiaohui Zeng}, \bibinfo{person}{Xun Huang}, \bibinfo{person}{Karsten Kreis}, \bibinfo{person}{Sanja Fidler}, \bibinfo{person}{Ming-Yu Liu}, {and} \bibinfo{person}{Tsung-Yi Lin}.} \bibinfo{year}{2023}\natexlab{}.
\newblock \showarticletitle{Magic3d: High-resolution text-to-3d content creation}. In \bibinfo{booktitle}{\emph{Proceedings of the IEEE/CVF Conference on Computer Vision and Pattern Recognition}}. \bibinfo{pages}{300--309}.
\newblock


\bibitem[Lorraine et~al\mbox{.}(2023)]%
        {lorraine2023att3d}
\bibfield{author}{\bibinfo{person}{Jonathan Lorraine}, \bibinfo{person}{Kevin Xie}, \bibinfo{person}{Xiaohui Zeng}, \bibinfo{person}{Chen-Hsuan Lin}, \bibinfo{person}{Towaki Takikawa}, \bibinfo{person}{Nicholas Sharp}, \bibinfo{person}{Tsung-Yi Lin}, \bibinfo{person}{Ming-Yu Liu}, \bibinfo{person}{Sanja Fidler}, {and} \bibinfo{person}{James Lucas}.} \bibinfo{year}{2023}\natexlab{}.
\newblock \showarticletitle{ATT3D: Amortized Text-to-3D Object Synthesis}.
\newblock \bibinfo{journal}{\emph{arXiv preprint arXiv:2306.07349}} (\bibinfo{year}{2023}).
\newblock


\bibitem[Lu et~al\mbox{.}(2007)]%
        {lu2007context}
\bibfield{author}{\bibinfo{person}{Jianye Lu}, \bibinfo{person}{Athinodoros~S Georghiades}, \bibinfo{person}{Andreas Glaser}, \bibinfo{person}{Hongzhi Wu}, \bibinfo{person}{Li-Yi Wei}, \bibinfo{person}{Baining Guo}, \bibinfo{person}{Julie Dorsey}, {and} \bibinfo{person}{Holly Rushmeier}.} \bibinfo{year}{2007}\natexlab{}.
\newblock \showarticletitle{Context-aware textures}.
\newblock \bibinfo{journal}{\emph{ACM Transactions on Graphics (TOG)}} \bibinfo{volume}{26}, \bibinfo{number}{1} (\bibinfo{year}{2007}), \bibinfo{pages}{3--es}.
\newblock


\bibitem[Mertens et~al\mbox{.}(2006)]%
        {mertens2006texture}
\bibfield{author}{\bibinfo{person}{Tom Mertens}, \bibinfo{person}{Jan Kautz}, \bibinfo{person}{Jiawen Chen}, \bibinfo{person}{Philippe Bekaert}, {and} \bibinfo{person}{Fr{\'e}do Durand}.} \bibinfo{year}{2006}\natexlab{}.
\newblock \showarticletitle{Texture Transfer Using Geometry Correlation.}
\newblock \bibinfo{journal}{\emph{Rendering Techniques}} \bibinfo{volume}{273}, \bibinfo{number}{10.2312} (\bibinfo{year}{2006}), \bibinfo{pages}{273--284}.
\newblock


\bibitem[Meshy(2023)]%
        {meshy}
\bibfield{author}{\bibinfo{person}{Meshy}.} \bibinfo{year}{2023}\natexlab{}.
\newblock \bibinfo{title}{Meshy | 3D AI Generator}.
\newblock \bibinfo{howpublished}{\url{https://www.meshy.ai/}}.
\newblock


\bibitem[Metzer et~al\mbox{.}(2023)]%
        {metzer2023latent}
\bibfield{author}{\bibinfo{person}{Gal Metzer}, \bibinfo{person}{Elad Richardson}, \bibinfo{person}{Or Patashnik}, \bibinfo{person}{Raja Giryes}, {and} \bibinfo{person}{Daniel Cohen-Or}.} \bibinfo{year}{2023}\natexlab{}.
\newblock \showarticletitle{Latent-nerf for shape-guided generation of 3d shapes and textures}. In \bibinfo{booktitle}{\emph{Proceedings of the IEEE/CVF Conference on Computer Vision and Pattern Recognition}}. \bibinfo{pages}{12663--12673}.
\newblock


\bibitem[Mou et~al\mbox{.}(2023)]%
        {mou2023t2i}
\bibfield{author}{\bibinfo{person}{Chong Mou}, \bibinfo{person}{Xintao Wang}, \bibinfo{person}{Liangbin Xie}, \bibinfo{person}{Jian Zhang}, \bibinfo{person}{Zhongang Qi}, \bibinfo{person}{Ying Shan}, {and} \bibinfo{person}{Xiaohu Qie}.} \bibinfo{year}{2023}\natexlab{}.
\newblock \showarticletitle{T2i-adapter: Learning adapters to dig out more controllable ability for text-to-image diffusion models}.
\newblock \bibinfo{journal}{\emph{arXiv preprint arXiv:2302.08453}} (\bibinfo{year}{2023}).
\newblock


\bibitem[Nam et~al\mbox{.}(2022)]%
        {nam20223d}
\bibfield{author}{\bibinfo{person}{Gimin Nam}, \bibinfo{person}{Mariem Khlifi}, \bibinfo{person}{Andrew Rodriguez}, \bibinfo{person}{Alberto Tono}, \bibinfo{person}{Linqi Zhou}, {and} \bibinfo{person}{Paul Guerrero}.} \bibinfo{year}{2022}\natexlab{}.
\newblock \showarticletitle{3d-ldm: Neural implicit 3d shape generation with latent diffusion models}.
\newblock \bibinfo{journal}{\emph{arXiv preprint arXiv:2212.00842}} (\bibinfo{year}{2022}).
\newblock


\bibitem[Nichol et~al\mbox{.}(2022)]%
        {Nichol2022PointEAS}
\bibfield{author}{\bibinfo{person}{Alex Nichol}, \bibinfo{person}{Heewoo Jun}, \bibinfo{person}{Prafulla Dhariwal}, \bibinfo{person}{Pamela Mishkin}, {and} \bibinfo{person}{Mark Chen}.} \bibinfo{year}{2022}\natexlab{}.
\newblock \showarticletitle{Point-E: A System for Generating 3D Point Clouds from Complex Prompts}.
\newblock \bibinfo{journal}{\emph{ArXiv}}  \bibinfo{volume}{abs/2212.08751} (\bibinfo{year}{2022}).
\newblock
\urldef\tempurl%
\url{https://api.semanticscholar.org/CorpusID:254854214}
\showURL{%
\tempurl}


\bibitem[Poole et~al\mbox{.}(2022)]%
        {poole2022dreamfusion}
\bibfield{author}{\bibinfo{person}{Ben Poole}, \bibinfo{person}{Ajay Jain}, \bibinfo{person}{Jonathan~T Barron}, {and} \bibinfo{person}{Ben Mildenhall}.} \bibinfo{year}{2022}\natexlab{}.
\newblock \showarticletitle{Dreamfusion: Text-to-3d using 2d diffusion}.
\newblock \bibinfo{journal}{\emph{arXiv preprint arXiv:2209.14988}} (\bibinfo{year}{2022}).
\newblock


\bibitem[Praun et~al\mbox{.}(2000)]%
        {praun2000lapped}
\bibfield{author}{\bibinfo{person}{Emil Praun}, \bibinfo{person}{Adam Finkelstein}, {and} \bibinfo{person}{Hugues Hoppe}.} \bibinfo{year}{2000}\natexlab{}.
\newblock \showarticletitle{Lapped textures}. In \bibinfo{booktitle}{\emph{Proceedings of the 27th annual conference on Computer graphics and interactive techniques}}. \bibinfo{pages}{465--470}.
\newblock


\bibitem[Radford et~al\mbox{.}(2021)]%
        {radford2021learning}
\bibfield{author}{\bibinfo{person}{Alec Radford}, \bibinfo{person}{Jong~Wook Kim}, \bibinfo{person}{Chris Hallacy}, \bibinfo{person}{Aditya Ramesh}, \bibinfo{person}{Gabriel Goh}, \bibinfo{person}{Sandhini Agarwal}, \bibinfo{person}{Girish Sastry}, \bibinfo{person}{Amanda Askell}, \bibinfo{person}{Pamela Mishkin}, \bibinfo{person}{Jack Clark}, {et~al\mbox{.}}} \bibinfo{year}{2021}\natexlab{}.
\newblock \showarticletitle{Learning transferable visual models from natural language supervision}. In \bibinfo{booktitle}{\emph{International conference on machine learning}}. PMLR, \bibinfo{pages}{8748--8763}.
\newblock


\bibitem[Raj et~al\mbox{.}(2019)]%
        {raj2019learning}
\bibfield{author}{\bibinfo{person}{Amit Raj}, \bibinfo{person}{Cusuh Ham}, \bibinfo{person}{Connelly Barnes}, \bibinfo{person}{Vladimir Kim}, \bibinfo{person}{Jingwan Lu}, {and} \bibinfo{person}{James Hays}.} \bibinfo{year}{2019}\natexlab{}.
\newblock \showarticletitle{Learning to generate textures on 3d meshes}. In \bibinfo{booktitle}{\emph{Proceedings of the IEEE/CVF Conference on Computer Vision and Pattern Recognition Workshops}}. \bibinfo{pages}{32--38}.
\newblock


\bibitem[Richardson et~al\mbox{.}(2023)]%
        {richardson2023texture}
\bibfield{author}{\bibinfo{person}{Elad Richardson}, \bibinfo{person}{Gal Metzer}, \bibinfo{person}{Yuval Alaluf}, \bibinfo{person}{Raja Giryes}, {and} \bibinfo{person}{Daniel Cohen-Or}.} \bibinfo{year}{2023}\natexlab{}.
\newblock \showarticletitle{TEXTure: Text-guided texturing of 3d shapes}.
\newblock \bibinfo{journal}{\emph{arXiv preprint arXiv:2302.01721}} (\bibinfo{year}{2023}).
\newblock


\bibitem[Rombach et~al\mbox{.}(2022)]%
        {rombach2022high}
\bibfield{author}{\bibinfo{person}{Robin Rombach}, \bibinfo{person}{Andreas Blattmann}, \bibinfo{person}{Dominik Lorenz}, \bibinfo{person}{Patrick Esser}, {and} \bibinfo{person}{Bj{\"o}rn Ommer}.} \bibinfo{year}{2022}\natexlab{}.
\newblock \showarticletitle{High-resolution image synthesis with latent diffusion models}. In \bibinfo{booktitle}{\emph{Proceedings of the IEEE/CVF conference on computer vision and pattern recognition}}. \bibinfo{pages}{10684--10695}.
\newblock


\bibitem[Sanghi et~al\mbox{.}(2022a)]%
        {sanghi2022clip}
\bibfield{author}{\bibinfo{person}{Aditya Sanghi}, \bibinfo{person}{Hang Chu}, \bibinfo{person}{Joseph~G Lambourne}, \bibinfo{person}{Ye Wang}, \bibinfo{person}{Chin-Yi Cheng}, \bibinfo{person}{Marco Fumero}, {and} \bibinfo{person}{Kamal~Rahimi Malekshan}.} \bibinfo{year}{2022}\natexlab{a}.
\newblock \showarticletitle{Clip-forge: Towards zero-shot text-to-shape generation}. In \bibinfo{booktitle}{\emph{Proceedings of the IEEE/CVF Conference on Computer Vision and Pattern Recognition}}. \bibinfo{pages}{18603--18613}.
\newblock


\bibitem[Sanghi et~al\mbox{.}(2022b)]%
        {9878592}
\bibfield{author}{\bibinfo{person}{Aditya Sanghi}, \bibinfo{person}{Hang Chu}, \bibinfo{person}{Joseph~G. Lambourne}, \bibinfo{person}{Ye Wang}, \bibinfo{person}{Chin-Yi Cheng}, \bibinfo{person}{Marco Fumero}, {and} \bibinfo{person}{Kamal~Rahimi Malekshan}.} \bibinfo{year}{2022}\natexlab{b}.
\newblock \showarticletitle{CLIP-Forge: Towards Zero-Shot Text-to-Shape Generation}. In \bibinfo{booktitle}{\emph{2022 IEEE/CVF Conference on Computer Vision and Pattern Recognition (CVPR)}}. \bibinfo{pages}{18582--18592}.
\newblock
\urldef\tempurl%
\url{https://doi.org/10.1109/CVPR52688.2022.01805}
\showDOI{\tempurl}


\bibitem[Smith and Meger(2017)]%
        {smith2017improved}
\bibfield{author}{\bibinfo{person}{Edward~J Smith} {and} \bibinfo{person}{David Meger}.} \bibinfo{year}{2017}\natexlab{}.
\newblock \showarticletitle{Improved adversarial systems for 3d object generation and reconstruction}. In \bibinfo{booktitle}{\emph{Conference on Robot Learning}}. PMLR, \bibinfo{pages}{87--96}.
\newblock


\bibitem[Sohl-Dickstein et~al\mbox{.}(2015)]%
        {sohl2015deep}
\bibfield{author}{\bibinfo{person}{Jascha Sohl-Dickstein}, \bibinfo{person}{Eric Weiss}, \bibinfo{person}{Niru Maheswaranathan}, {and} \bibinfo{person}{Surya Ganguli}.} \bibinfo{year}{2015}\natexlab{}.
\newblock \showarticletitle{Deep unsupervised learning using nonequilibrium thermodynamics}. In \bibinfo{booktitle}{\emph{International conference on machine learning}}. PMLR, \bibinfo{pages}{2256--2265}.
\newblock


\bibitem[Tsalicoglou et~al\mbox{.}(2023)]%
        {tsalicoglou2023textmesh}
\bibfield{author}{\bibinfo{person}{Christina Tsalicoglou}, \bibinfo{person}{Fabian Manhardt}, \bibinfo{person}{Alessio Tonioni}, \bibinfo{person}{Michael Niemeyer}, {and} \bibinfo{person}{Federico Tombari}.} \bibinfo{year}{2023}\natexlab{}.
\newblock \showarticletitle{TextMesh: Generation of Realistic 3D Meshes From Text Prompts}.
\newblock \bibinfo{journal}{\emph{arXiv preprint arXiv:2304.12439}} (\bibinfo{year}{2023}).
\newblock


\bibitem[Van Den~Oord et~al\mbox{.}(2017)]%
        {van2017neural}
\bibfield{author}{\bibinfo{person}{Aaron Van Den~Oord}, \bibinfo{person}{Oriol Vinyals}, {et~al\mbox{.}}} \bibinfo{year}{2017}\natexlab{}.
\newblock \showarticletitle{Neural discrete representation learning}.
\newblock \bibinfo{journal}{\emph{Advances in neural information processing systems}}  \bibinfo{volume}{30} (\bibinfo{year}{2017}).
\newblock


\bibitem[Vaswani et~al\mbox{.}(2017)]%
        {vaswani2017attention}
\bibfield{author}{\bibinfo{person}{Ashish Vaswani}, \bibinfo{person}{Noam Shazeer}, \bibinfo{person}{Niki Parmar}, \bibinfo{person}{Jakob Uszkoreit}, \bibinfo{person}{Llion Jones}, \bibinfo{person}{Aidan~N Gomez}, \bibinfo{person}{{\L}ukasz Kaiser}, {and} \bibinfo{person}{Illia Polosukhin}.} \bibinfo{year}{2017}\natexlab{}.
\newblock \showarticletitle{Attention is all you need}.
\newblock \bibinfo{journal}{\emph{Advances in neural information processing systems}}  \bibinfo{volume}{30} (\bibinfo{year}{2017}).
\newblock


\bibitem[Wang et~al\mbox{.}(2024)]%
        {wang2024prolificdreamer}
\bibfield{author}{\bibinfo{person}{Zhengyi Wang}, \bibinfo{person}{Cheng Lu}, \bibinfo{person}{Yikai Wang}, \bibinfo{person}{Fan Bao}, \bibinfo{person}{Chongxuan Li}, \bibinfo{person}{Hang Su}, {and} \bibinfo{person}{Jun Zhu}.} \bibinfo{year}{2024}\natexlab{}.
\newblock \showarticletitle{Prolificdreamer: High-fidelity and diverse text-to-3d generation with variational score distillation}.
\newblock \bibinfo{journal}{\emph{Advances in Neural Information Processing Systems}}  \bibinfo{volume}{36} (\bibinfo{year}{2024}).
\newblock


\bibitem[Wong et~al\mbox{.}(1997)]%
        {wong1997geometry}
\bibfield{author}{\bibinfo{person}{Tien-Tsin Wong}, \bibinfo{person}{Wai-Yin Ng}, {and} \bibinfo{person}{Pheng-Ann Heng}.} \bibinfo{year}{1997}\natexlab{}.
\newblock \showarticletitle{A geometry dependent texture generation framework for simulating surface imperfections}. In \bibinfo{booktitle}{\emph{Rendering Techniques' 97: Proceedings of the Eurographics Workshop in St. Etienne, France, June 16--18, 1997 8}}. Springer, \bibinfo{pages}{139--150}.
\newblock


\bibitem[Xian et~al\mbox{.}(2018)]%
        {xian2018texturegan}
\bibfield{author}{\bibinfo{person}{Wenqi Xian}, \bibinfo{person}{Patsorn Sangkloy}, \bibinfo{person}{Varun Agrawal}, \bibinfo{person}{Amit Raj}, \bibinfo{person}{Jingwan Lu}, \bibinfo{person}{Chen Fang}, \bibinfo{person}{Fisher Yu}, {and} \bibinfo{person}{James Hays}.} \bibinfo{year}{2018}\natexlab{}.
\newblock \showarticletitle{Texturegan: Controlling deep image synthesis with texture patches}. In \bibinfo{booktitle}{\emph{Proceedings of the IEEE conference on computer vision and pattern recognition}}. \bibinfo{pages}{8456--8465}.
\newblock


\bibitem[Youwang et~al\mbox{.}(2024)]%
        {youwang2024paint}
\bibfield{author}{\bibinfo{person}{Kim Youwang}, \bibinfo{person}{Tae-Hyun Oh}, {and} \bibinfo{person}{Gerard Pons-Moll}.} \bibinfo{year}{2024}\natexlab{}.
\newblock \showarticletitle{Paint-it: Text-to-texture synthesis via deep convolutional texture map optimization and physically-based rendering}. In \bibinfo{booktitle}{\emph{Proceedings of the IEEE/CVF Conference on Computer Vision and Pattern Recognition}}. \bibinfo{pages}{4347--4356}.
\newblock


\bibitem[Yu et~al\mbox{.}(2023)]%
        {yu2023pointsto3d}
\bibfield{author}{\bibinfo{person}{Chaohui Yu}, \bibinfo{person}{Qiang Zhou}, \bibinfo{person}{Jingliang Li}, \bibinfo{person}{Zhe Zhang}, \bibinfo{person}{Zhibin Wang}, {and} \bibinfo{person}{Fan Wang}.} \bibinfo{year}{2023}\natexlab{}.
\newblock \bibinfo{title}{Points-to-3D: Bridging the Gap between Sparse Points and Shape-Controllable Text-to-3D Generation}.
\newblock
\newblock
\showeprint[arxiv]{2307.13908}~[cs.CV]


\bibitem[Zeng et~al\mbox{.}(2024)]%
        {zeng2024paint3d}
\bibfield{author}{\bibinfo{person}{Xianfang Zeng}, \bibinfo{person}{Xin Chen}, \bibinfo{person}{Zhongqi Qi}, \bibinfo{person}{Wen Liu}, \bibinfo{person}{Zibo Zhao}, \bibinfo{person}{Zhibin Wang}, \bibinfo{person}{Bin Fu}, \bibinfo{person}{Yong Liu}, {and} \bibinfo{person}{Gang Yu}.} \bibinfo{year}{2024}\natexlab{}.
\newblock \showarticletitle{Paint3d: Paint anything 3d with lighting-less texture diffusion models}. In \bibinfo{booktitle}{\emph{Proceedings of the IEEE/CVF Conference on Computer Vision and Pattern Recognition}}. \bibinfo{pages}{4252--4262}.
\newblock


\bibitem[Zeng et~al\mbox{.}(2022)]%
        {Zeng2022LIONLP}
\bibfield{author}{\bibinfo{person}{Xiaohui Zeng}, \bibinfo{person}{Arash Vahdat}, \bibinfo{person}{Francis Williams}, \bibinfo{person}{Zan Gojcic}, \bibinfo{person}{Or Litany}, \bibinfo{person}{Sanja Fidler}, {and} \bibinfo{person}{Karsten Kreis}.} \bibinfo{year}{2022}\natexlab{}.
\newblock \showarticletitle{LION: Latent Point Diffusion Models for 3D Shape Generation}.
\newblock \bibinfo{journal}{\emph{ArXiv}}  \bibinfo{volume}{abs/2210.06978} (\bibinfo{year}{2022}).
\newblock
\urldef\tempurl%
\url{https://api.semanticscholar.org/CorpusID:252872881}
\showURL{%
\tempurl}


\bibitem[Zhang et~al\mbox{.}(2023)]%
        {zhang2023adding}
\bibfield{author}{\bibinfo{person}{Lvmin Zhang}, \bibinfo{person}{Anyi Rao}, {and} \bibinfo{person}{Maneesh Agrawala}.} \bibinfo{year}{2023}\natexlab{}.
\newblock \showarticletitle{Adding conditional control to text-to-image diffusion models}. In \bibinfo{booktitle}{\emph{Proceedings of the IEEE/CVF International Conference on Computer Vision}}. \bibinfo{pages}{3836--3847}.
\newblock


\bibitem[Zhou et~al\mbox{.}(2021)]%
        {zhou20213d}
\bibfield{author}{\bibinfo{person}{Linqi Zhou}, \bibinfo{person}{Yilun Du}, {and} \bibinfo{person}{Jiajun Wu}.} \bibinfo{year}{2021}\natexlab{}.
\newblock \showarticletitle{3d shape generation and completion through point-voxel diffusion}. In \bibinfo{booktitle}{\emph{Proceedings of the IEEE/CVF International Conference on Computer Vision}}. \bibinfo{pages}{5826--5835}.
\newblock


\end{thebibliography}

% Appendix
\appendix
\clearpage

\end{document}